%% file: arxiv.tex
\documentclass[]{fairmeta}

\usepackage{lmodern}       
\usepackage[T1]{fontenc}
\usepackage{hyperref}
\usepackage{url}
\usepackage{booktabs}
\usepackage{verbatim}
\usepackage{url}
\usepackage{algorithm}
\usepackage{algorithmicx}
\usepackage[noend]{algpseudocode}
\usepackage{amsmath, amsfonts, amssymb, amsthm, amsbsy, amscd, bm, bbm,mathrsfs} \usepackage{amssymb}
\usepackage{graphicx}
\usepackage{setspace}
\usepackage{hyperref}
\usepackage{cancel}
\usepackage{cleveref}
\usepackage[hyperpageref]{backref}
\usepackage{appendix}
\usepackage{xspace}
\usepackage{epigraph}
\usepackage[thinc]{esdiff}
\graphicspath{{figs/}}
\usepackage[most]{tcolorbox}
\usepackage{wrapfig}
\usepackage{multirow}
\newtcolorbox{untitledbox}{
    enhanced, breakable, rounded corners, center title,
    colframe = cyan,         
    colback = cyan!10,       
    colbacktitle = cyan!80!black, 
  left=0mm,   
  right=0mm,  
  top=1mm,bottom=1mm,
  }
\usepackage[labelfont=bf]{caption}
\title{
Scaling depth capacity via zero/one-layer model expansion}

\author{Zhiqi Bu}
\affiliation{Meta FAIR}
\correspondence{\email{zhiqibu@meta.com}}
\makeatletter
\newcommand{\be}{\begin{equation}}
\newcommand{\ee}{\end{equation}}

\newcommand{\W}{\mathbf{W}}

\newcommand{\A}{\mathbf{A}}

\allowdisplaybreaks \numberwithin{equation}{section}
\makeatother

\def\x{\mathbf{x}}

\newcommand{\w}{\mathbf{w}}

\newcommand{\R}{\mathbb{R}}

\abstract{
Model depth is a double-edged sword in deep learning: deeper models achieve higher accuracy but require higher computational cost. To efficiently train models at scale, progressive training (also known as model expansion) scales up model capacity during training and significantly reduces computation with little performance degradation.
In this work, we study the depth expansion of large-scale models through the lens of optimization theory and feature learning, offering insights on the initialization of new layers, hyperparameter transfer, learning rate schedule, and timing of model expansion. Specifically, we propose zero/one-layer progressive training to achieve an optimal tradeoff between computation and loss, with a comprehensive ablations on our expansion strategy. For example, zero/one-layer progressive training on GPT2 can save $\approx 80\%$ compute, or equivalently achieve an $\approx 5\times$ acceleration, while attaining a loss comparable to that of a fully trained 60-layer model with 7B parameters, thus demonstrating a mixing behavior in terms of loss. Furthermore, scaling laws on LLAMA3 and DeepSeekV3 models show a $3\sim 5\times$ improvement in compute efficiency, with an increasing advantage at larger scales.
}

\begin{document}

\maketitle

\input{main_arxiv}

\clearpage
\bibliography{references}
\bibliographystyle{assets/plainnat}

\appendix
\input{appendix}

\end{document}

%% file: main_arxiv.tex
\section{Introduction}
Strong performance of deep learning models is highly correlated to model sizes, with larger model having higher accuracy but also incurring higher computation cost to train, e.g. LLAMA-4 training costs over 7M GPU hours and an estimated 2,000 tons of carbon emissions. This phenomenon leads to a tradeoff between model utility (measured by loss or accuracy) and computational cost (measured by floating point operation, or FLOP), and has motivated scaling laws to train compute-optimal large language models \citep{hoffmann2022training,kaplan2020scaling}.

\begin{figure*}[!htb]
	\centering
	\includegraphics[width=0.49\linewidth]{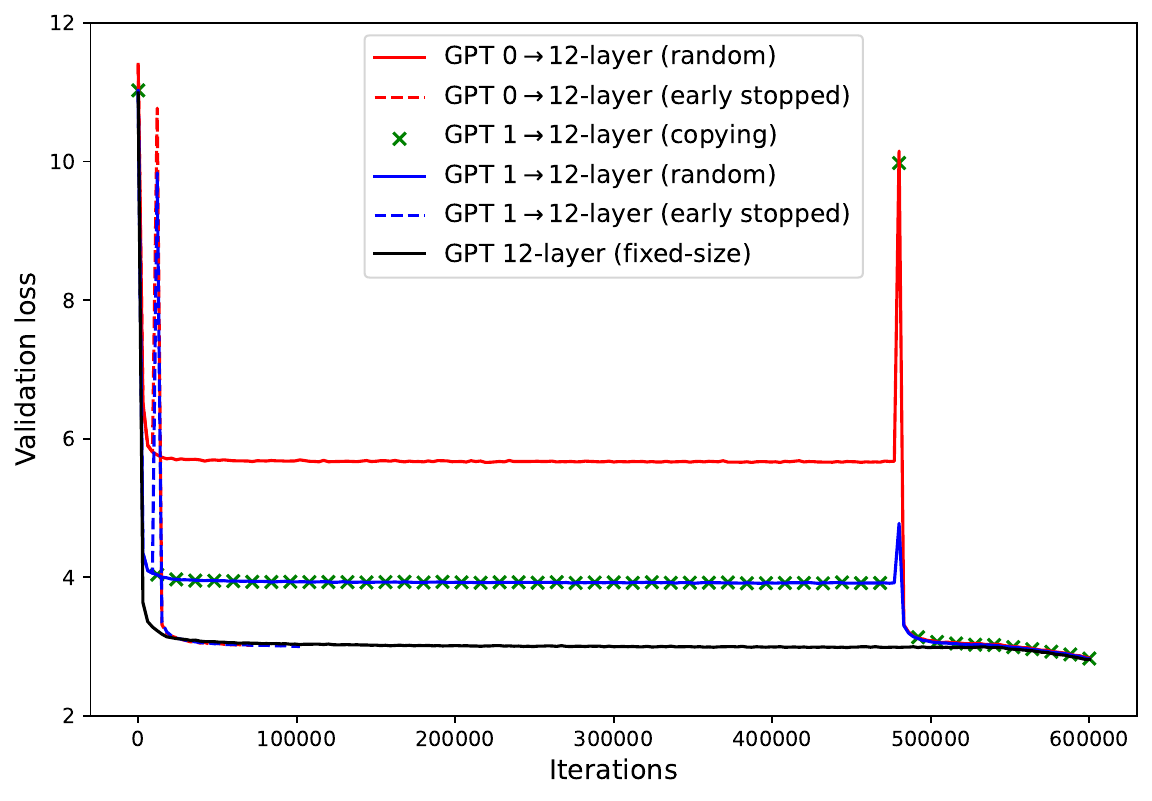}
	\includegraphics[width=0.49\linewidth]{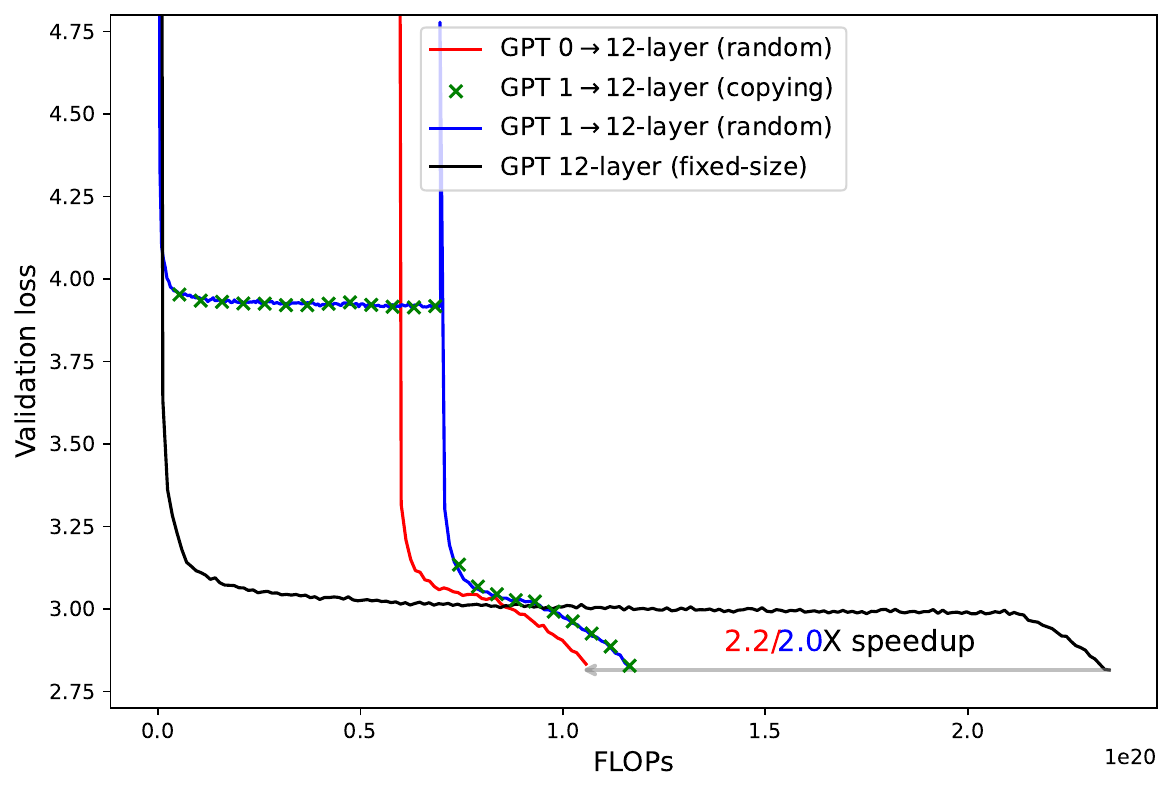}
	\\
	\includegraphics[width=0.49\linewidth]{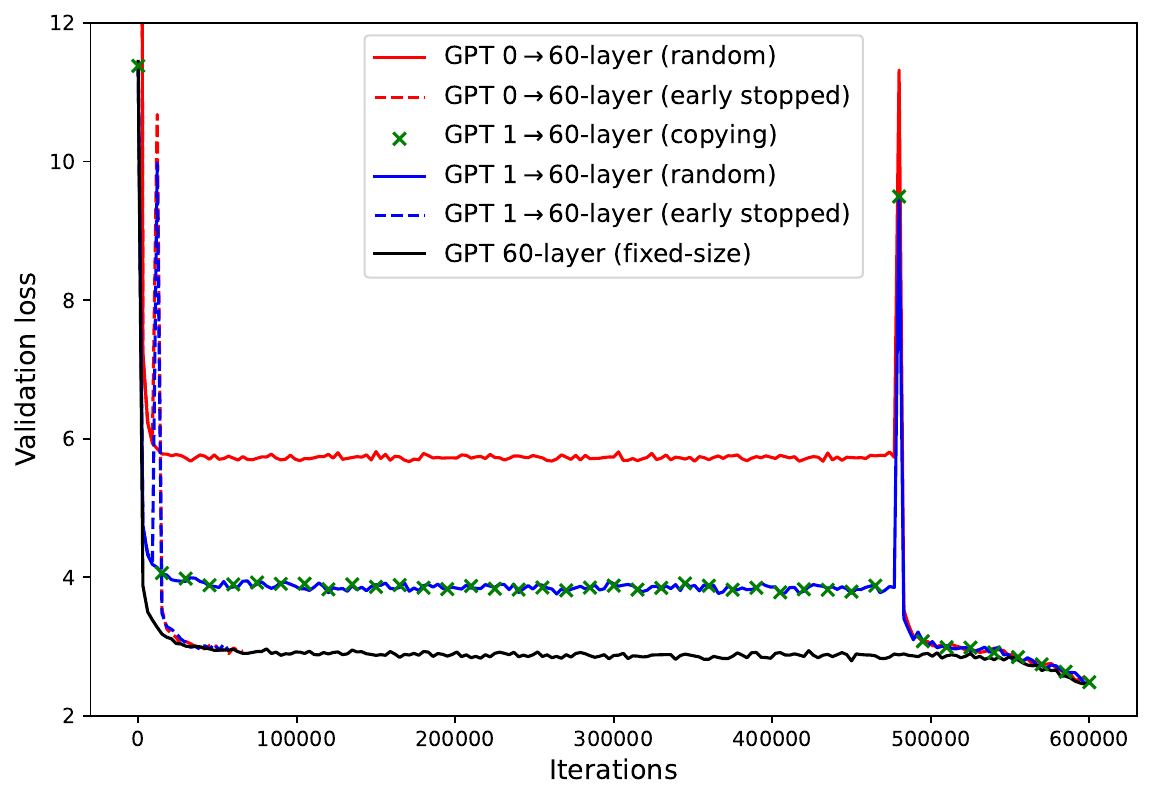}
	\includegraphics[width=0.49\linewidth]{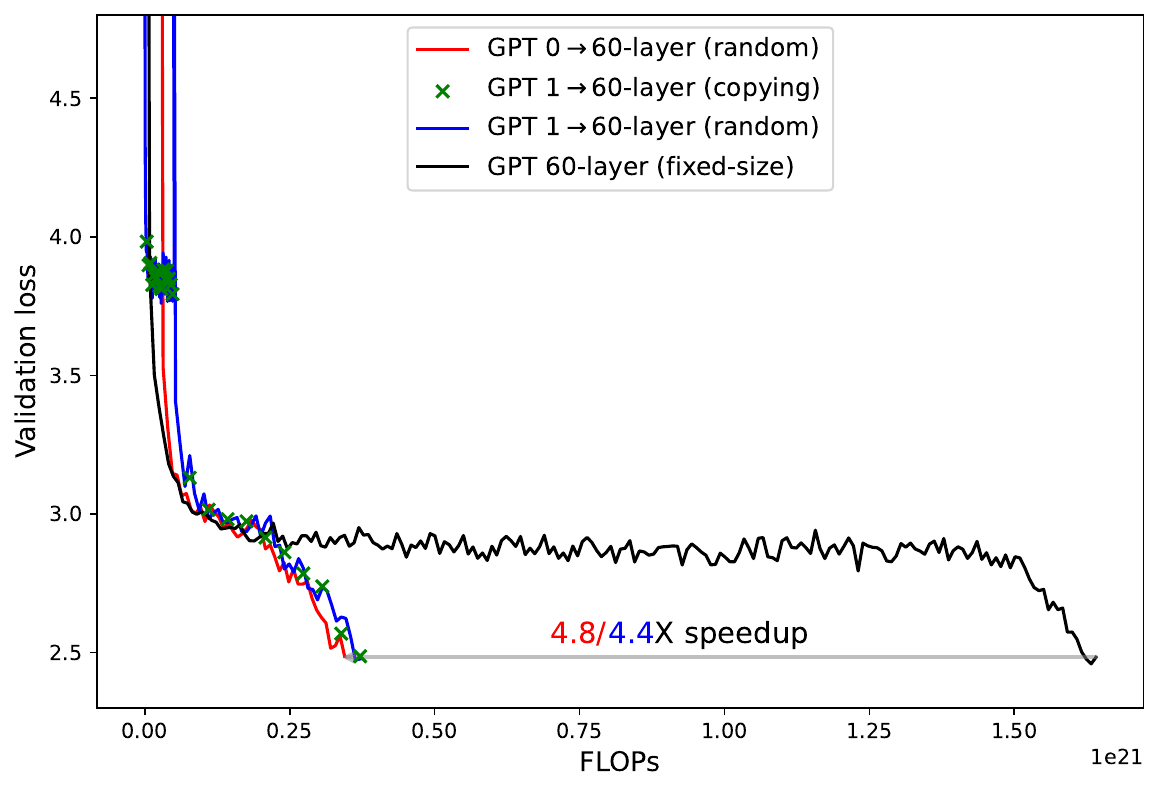}
	\caption{Zero-layer (\textcolor{red}{red}, 39M or 0.15B) and one-layer (\textcolor{blue}{blue}, 46M or 0.27B) progressive training can achieve significant speedup over fixed-sized training (black, 12-layer 124M or 60-layer 7B) in GPT2 pre-training on OpenWebText under WSD schedule. The difference in final validation loss is $<0.5\%$ for 124M runs and $<0.2\%$ for 7B runs. For full runs, the depth expansion happens at 80\% of iterations; for early stopped runs, the depth expansion happens at 2\% of iterations where the warmup ends.}
	\label{fig:long}
\end{figure*}

To accelerate the training of large models, one direction is known as progressive training, or model expansion, or model growth, which initially trains a small model (a.k.a. teacher or source model) and then scales up to large models (a.k.a. student or grown model) during training. In contrast to the fixed-size training, the progressive training formulates the model size as a time-dependent variable, and it is clearly more efficient because the compute is $6BTN$, proportional to the model size $N$. For example, consider a progressive training that scales up the model size at iteration $\tau$:
\begin{align}
	N(t)=
	\begin{cases}
		N_\text{small} & \textup{ if } t\leq\tau
		\\
		N_\text{large} & \textup{ if } t>\tau
	\end{cases}
	\label{eq:1 stage}
\end{align}
The fixed-size training requires $6BTN_\text{large}$ FLOPs, whereas the progressive training requires $6B(\tau N_\text{small}+(T-\tau)N_\text{large})$, which is significantly less if (I) $\tau$ is close to $T$ and (II) $N_\text{small}\ll N_\text{large}$. As a brief preview, we will develop techniques to push $\tau\approx 0.8T$ and to train zero/one-layer small models, hence accelerating by $\approx 5\times$ in \Cref{fig:long}.

A long list of research has contributed to the development of progressive training, especially on initialization of large models, multi-stage training, training regime, and theory.
\paragraph{Initialization from precedented small models.}
\citep{chen2015net2net, wanglemon,yaomasked} study the function-preserving initialization, such that the large model has the same loss and function as the small model at the moment of depth expansion. These works scale up the depth of convolution networks and BERT by $2\times$ and reduce the computation to $\approx 70\%$ computation. However, while function-preserving guarantees nice behavior during depth expansion, it does not guarantee fast convergence after the expansion. Alternatively, without function-preserving, \citep{chen2021bert2bert} linearly combines two layers to initialize a new layer; \citep{gong2019efficient,yang2020progressively,du2024stacking} stacks the old layers multiple times; \citep{qin2021knowledge,wang2023learning} propose learning-based methods that require extra training. These methods empirically scale up the depth by $2\sim 4\times$ and reduce the ``grown v.s. target'' computation to $\approx 55\sim 70\%$ (e.g. \citep{wang2023learning}, Figure 1 in \citep{chen2021bert2bert}, Figure 6 in \citep{pan2024preparing}, and Figure 1 in \citep{du2024stacking}). In contrast, we scale up the depth to $60\times$ and reduce the computation to 20\%.

\begin{figure*}[!htb]
	\centering
	\includegraphics[width=0.49\linewidth]{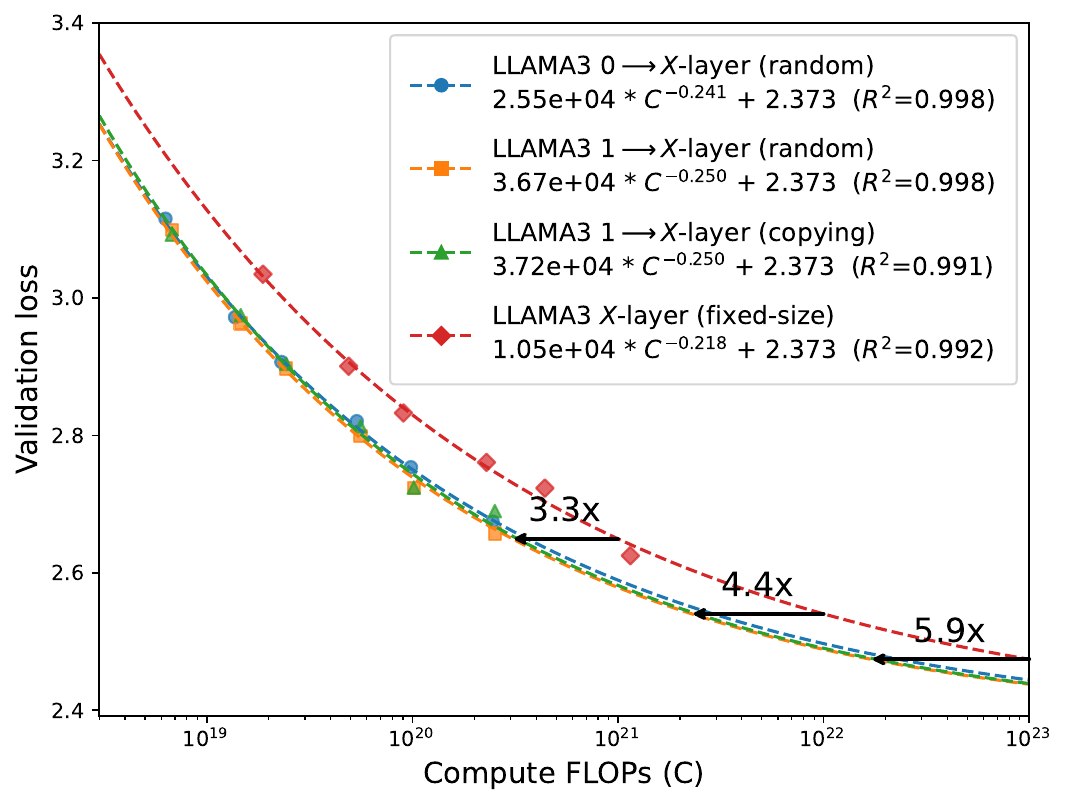}
	\includegraphics[width=0.49\linewidth]{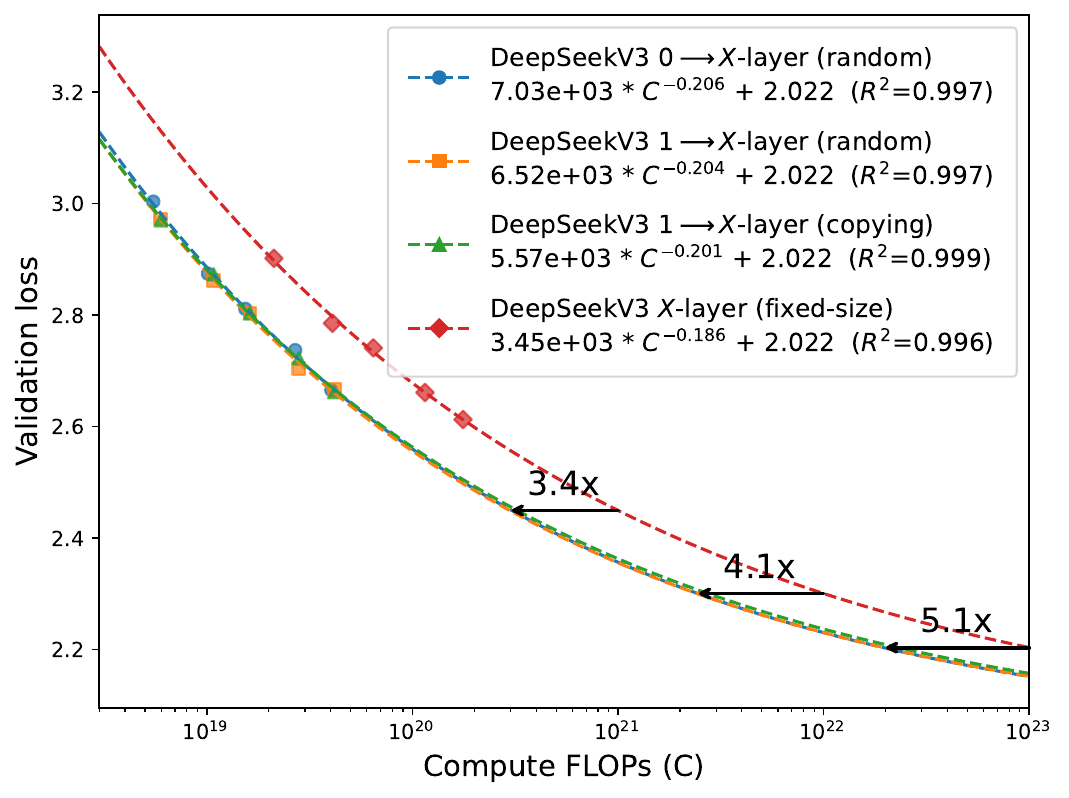}
	\caption{Scaling laws of zero-layer and one-layer progressive training show significant compute efficiency gain over fixed-sized training in LLAMA3 (dense, token-per-param=50) and DeepSeekV3 (MoE, token-per-param=100) pre-training on OpenWebText under WSD schedule. The depth expansion happens at 80\% of iterations. Model sizes range from 0.25B to 2B for LLAMA3, and from 0.2B to 0.5B (active) for DeepSeekV3. We observe that our progressive training consistently has better exponent in the scaling laws.}
	\label{fig:scaling}
\end{figure*}
\paragraph{Multi-stage training.}
Most works in progressive training expand the small models once like \eqref{eq:1 stage}.
However, many works study multi-stage training and gradual stacking \citep{reddi2023efficient}. For instance, \citep{gong2019efficient,shen2022staged,qin2021knowledge,pan2024preparing,yaomasked} scale up the sizes of BERT for $3\sim 4 \times$ during training, optionally freezing some of the layers at some stages \citep{agarwal2024stacking,yang2020progressively}. We note that none of these multi-stage methods demonstrate the mixing behavior as shown in this work.

\paragraph{Training regime.}
While most progressive training methods are tested on classification models like BERT \citep{devlin2019bert} and ViT \citep{dosovitskiy2020image}, some recent papers have evaluated on generative language models like GPT2 and reported $1.4\sim 2\times$ speedup. As shown in \Cref{fig:long}, our method enjoys $5\times$ speedup across different model sizes. We also note some papers that scale up MoE (not in terms of depth though) but the speedup seems transient as discussed in \Cref{sec:recipe}.

\paragraph{Theory.}
Theoretical analysis in progressive training is largely lacking, except \citep{agarwal2024stacking} on strongly-convex and smooth loss. In contrast, we give a convergence theory of convex and Lipschitz continuous (non-smooth) loss and empirically validate its insights. Besides a convergence theory, we also study feature learning and hyperparameter transfer for progressive training.

\subsection{Related work}
In addition to previous works on progressive training, this work is closely related to convex optimization in \Cref{sec:convex theory}, feature learning theory in \Cref{sec:muP}, and learning rate schedules (especially warmup-stable-decay; WSD \citep{xing2018walk,hagele2024scaling}).

\subsection{Contributions}
To our best knowledge, we are the first to advocate zero/one-layer depth expansion and to explore WSD schedule in progressive training. We summarize our main contributions here and provide extra insights on optimizer choice, optimizer states, model size, batch size, multi-stage expansion, and other topics in \Cref{app:insights}. 
\begin{enumerate}
	\item We analyze the depth expansion as an initialization problem and ensure feature learning. This approach allows \textbf{hyperparameter transfer} (e.g. learning rate) throughout the progressive training, in contrast to extra hyperparameter tuning \citep{gu2020transformer,yano2025efficient}. 
	\item We reveal the important role of learning rate schedule, especially \textbf{WSD schedule} which theoretically and empirically improves the convergence.
	\item We discover the \textbf{mixing behaviors} of progressive training, which supports the mixing time transfer and single-stage expansion, in contrast to multi-stage expansion by \citep{gong2019efficient,yang2020progressively,qin2021knowledge,shen2022staged,pan2024preparing}.
	\item We show that \textbf{zero/one-layer progressive training} has the best tradeoff between computational cost and loss, compared to other progressive or fixed-size training. Specifically, we validate the scaling laws in \Cref{fig:scaling} across a range of models and observe substantial gains in compute efficiency, with the performance advantage increasing at larger scales.
	\item We analyze the \textbf{convergence theory} of progressive training under convex optimization to give insights on initialization, learning rate schedule, and projected gradient descent.
\end{enumerate}

\section{Experiment settings}
We use the LLM (large language model) and ResNet \citep{he2016deep} as testbeds\footnote{For LLM, the models are configured as [Embedding, Hidden$\times N$, LM\_head (with LayerNorm)], where `Hidden' stands for the transformer layer and $N$ is number of layers, e.g. zero/one-layer GPT2 is $N=0$ or $1$. For ResNet, the models are configured by 4 stages, e.g. ResNet50 with [3,4,6,3] and ResNet101 with [3,4,23,3]. In each stage, the first layer has one shape, and each of other layers has the same shape which is different to the first layer.
	Hence the zero-layer analogy corresponds to ResNet14 with [1,1,1,1] and the one-layer analogy corresponds to ResNet26 with [2,2,2,2].}
. For language models, we train on OpenWebText dataset \citep{Gokaslan2019OpenWeb} with 1024 sequence length by adapting the \href{https://github.com/karpathy/nanoGPT}{nanoGPT codebase}. We experiment with GPT2 \citep{radford2019language}, LLAMA3 \citep{dubey2024llama}, Qwen3 \citep{yang2025qwen3}, Mixtral \citep{jiang2024mixtral}, and DeepSeekV3 \citep{liu2024deepseek}. For ResNet, we train on ImageNet dataset with $224\times 224$ resolutions for 100 epochs.

Our main optimizer is Muon-NSGD, with 0.01 weight decay and without gradient clipping. The Muon-NSGD is adapted from the original Muon \citep{jordan2024muon} by (1) optimizing all 2D tensors with Muon and other tensors with normalized SGD (NSGD), and (2) using a single learning rate for Muon and NSGD. We use the cosine learning rate schedule and WSD schedule, that decay to 0 with 2\% warm-up. We also experiment with AdamW and SGD optimizers. Additional details are in \Cref{app:settings}.

Notably, our experiments show consistent patterns while covering different designs of LLM:
\begin{itemize}
	\item weight tying or not;
	\item sparsity (dense and MoE);
	\item attention mechanism (MHA, MLA, and GQA);
	\item position embedding (absolute and rotary);
	\item normalization (layernorm and RMSNorm);
	\item activation function (GeLU and SwiGLU).
\end{itemize}

\section{How to expand depth?}
\subsection{Depth expansion approaches}
We introduce multiple approaches to expand the depth of a residual neural network.
\begin{itemize}
	\item \textbf{[copying]}: New layers are copied from the small model \citep{chang2018multi,gong2019efficient,li2020shallow}.
	\item \textbf{[random]}: New layers are randomly initialized \citep{wang2017growing,chen2021bert2bert}. 
	\item \textbf{[zero]}: New layers are initialized as zeros. This approach kills the gradient flow and makes the new layers untrainable, hence invalidating the progressive training
	.
	\item \textbf{[copying\_zero]}: New layers are copied from the small model, except some sub-layers are zero \citep{shen2022staged,wanglemon,tan2024dlo,wu2024llama,du2024stacking}.
\end{itemize}

\begin{figure}[!htb]
	\centering
	\includegraphics[width=0.478\linewidth]{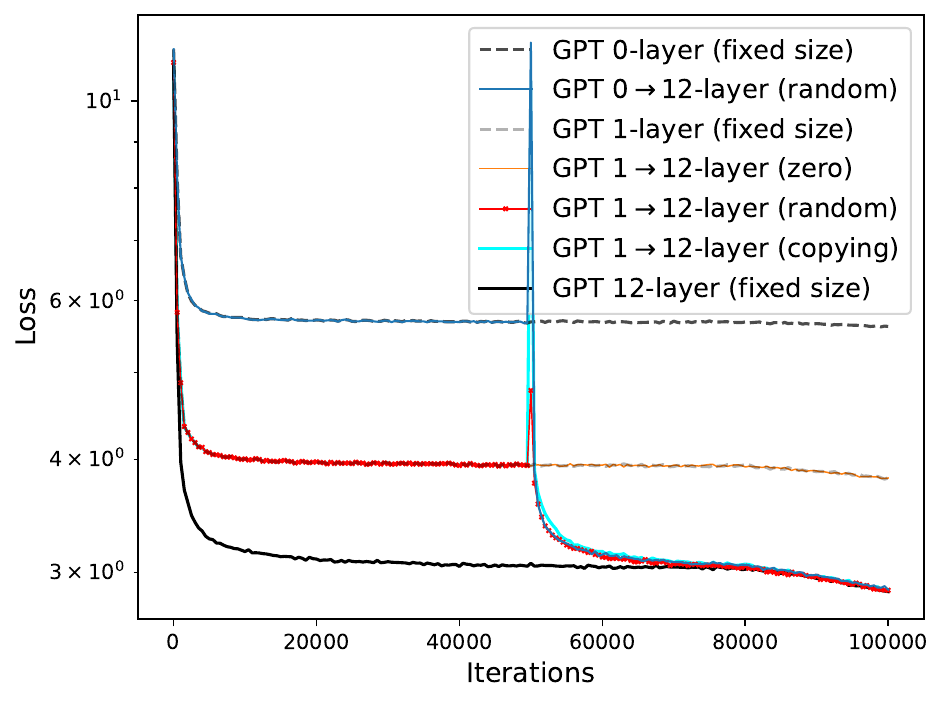}
	\\
	\includegraphics[width=0.464945\linewidth]{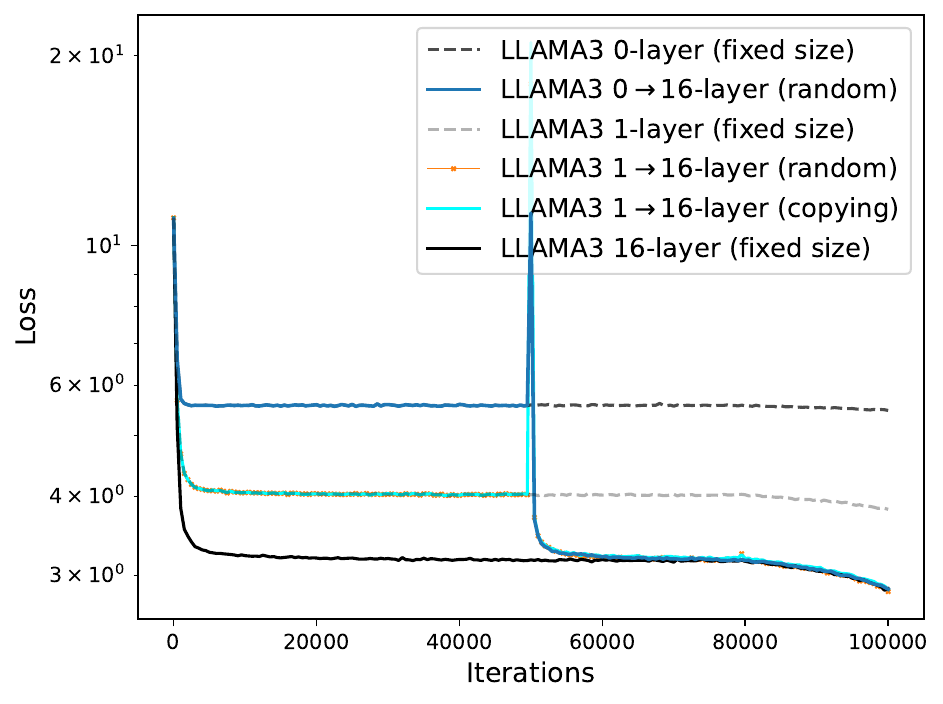}
	\includegraphics[width=0.464945\linewidth]{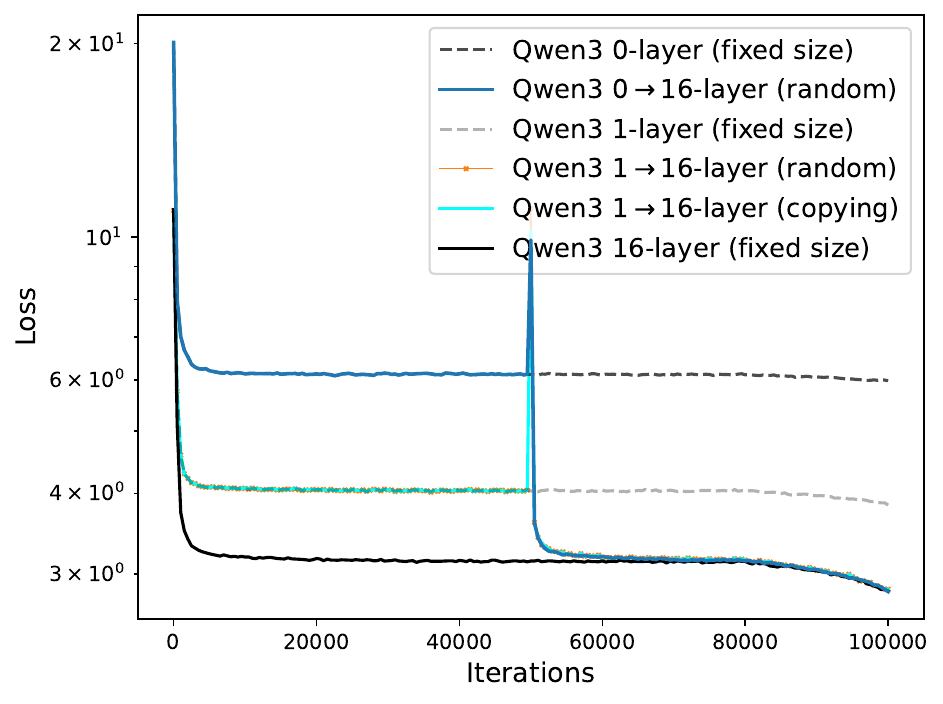}
	\includegraphics[width=0.464945\linewidth]{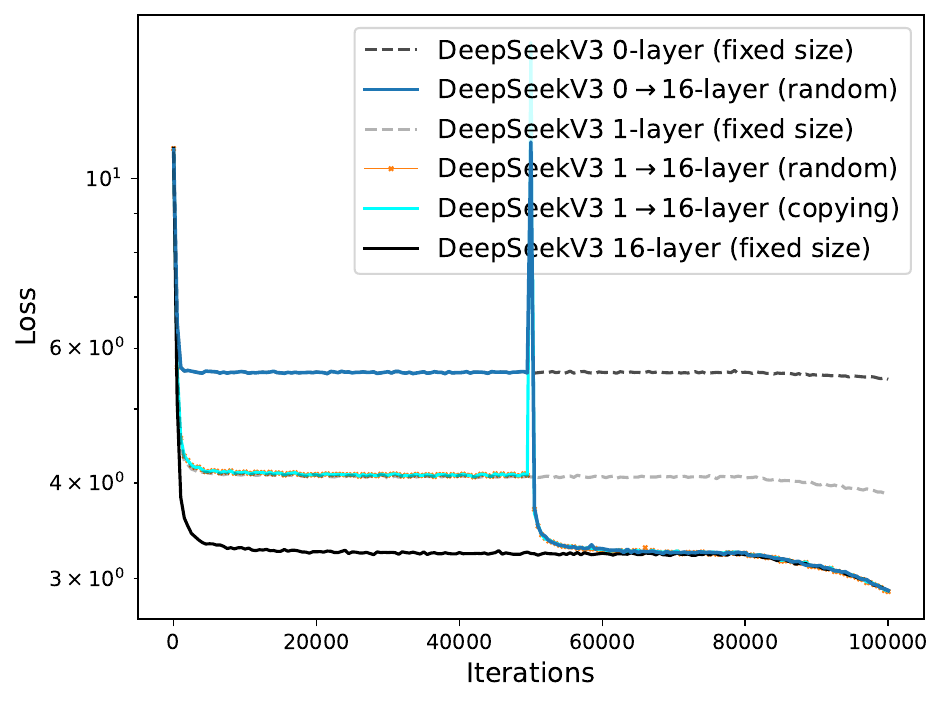}
	\includegraphics[width=0.464945\linewidth]{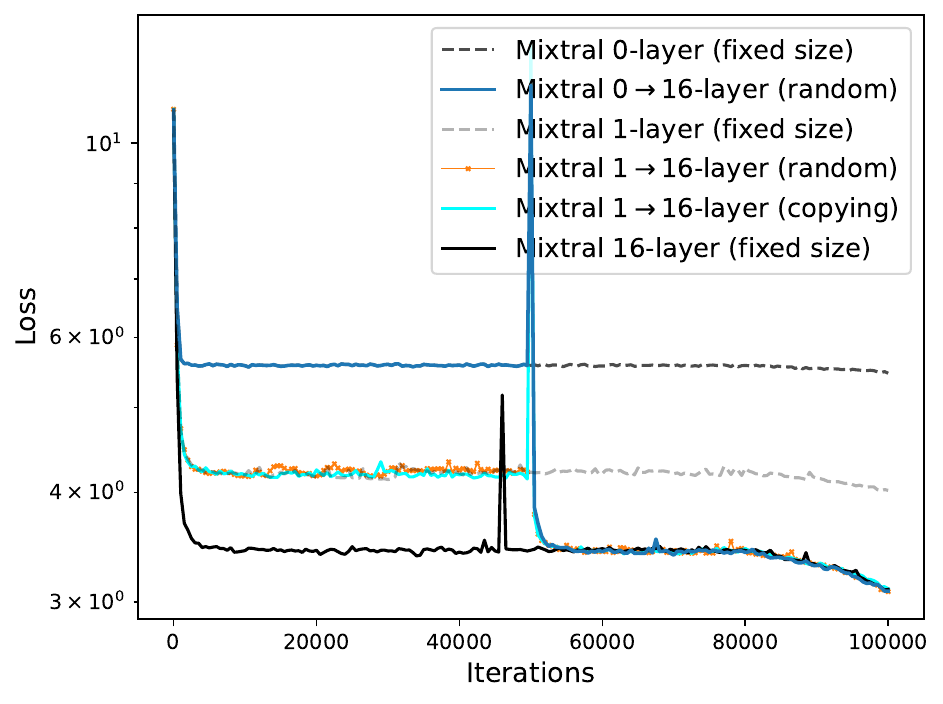}
	\vspace{-0.3cm}
	\caption{Convergence of zero/one-layer progressive training and fixed-size training, with depth expansion at 50k iterations. Top to bottom: GPT2 (dense, MHA), LLAMA3 (dense, GQA), Qwen3 (dense, GQA), DeepSeekV3 (MoE, MLA), Mixtral (MoE, GQA).}
	\label{fig:resnetGPT_cosine_all}
\end{figure}

To test these approaches in a minimalist manner, we expand zero/one-layer models to multiple layers in \Cref{fig:resnetGPT_cosine_all}. The experiments of copying\_zero approach can be found in \Cref{app:expansion methods}, and more experiments on Mixture-of-Experts (MoE, \citep{fedus2022switch}) in \Cref{sec:recipe}.

\begin{untitledbox}\textbf{Takeaway 1: }
	For zero/one-layer progressive training, \textit{random} and \textit{copying} are empirically the best initializations of new layers.
\end{untitledbox}

\subsection{Feature learning and Hyperparameter transfer}
\label{sec:muP}
\vspace{-0.1cm}
To ensure feature learning and keep the representations non-trivial and stable, each layer's activation needs to have consistent element sizes: denoting $l$-th layer's activation as  $\A_l\in\R^{n_l}$, then ${\|\A_l\|_2}/{\sqrt{n_l}}\sim {\|\A_{l+1}\|_2}/{\sqrt{n_{l+1}}}$ \citep{mei2019mean,yang2020feature,chizat2018global,yang2022tensor,yang2023spectral}. For linear layers $\A_{l+1}=\A_l\W_{l}$, this translates to the spectral scaling condition by muP theory, i.e. the spectral norm $\|\W_{l}\|_*\sim \sqrt{n_{l+1}/n_l}$ for all layers.

\begin{figure}[!htb]
	\centering
	\includegraphics[width=0.443\linewidth]{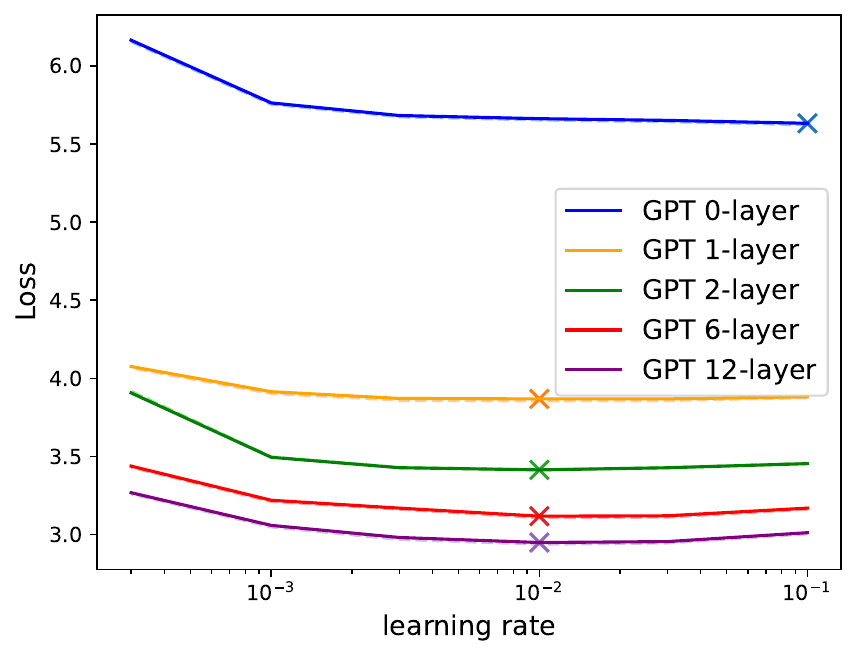}
	\includegraphics[width=0.443\linewidth]{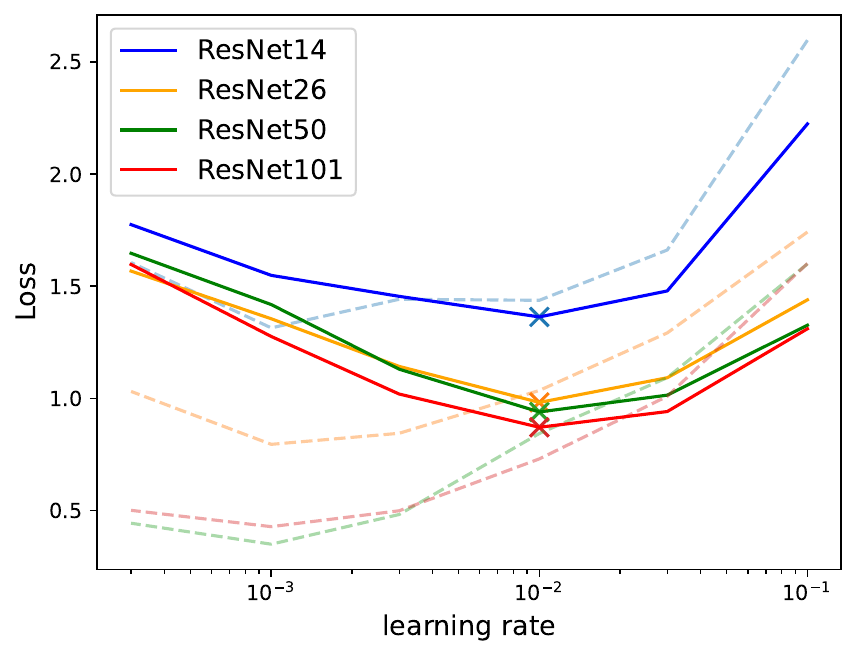}
	\vspace{-0.3cm}
	\caption{Validation (solid) and training loss (dashed) at different learning rates of Muon-NSGD.}
	\label{fig:mup table}
\end{figure}

Importantly, muP allows zero-shot hyperparameter transfer across model sizes, so that the optimal hyperparameters (e.g. learning rate) are the same for small and large models. In \Cref{fig:mup table}, we illustrate this on the Muon-NSGD optimizer with muP-scaling learning rate. We highlight that hyperparameter transfer is particularly desirable in progressive training, where model sizes change significantly before and after the model expansion.

In fact, \textit{copying} and \textit{random} satisfy muP condition, but \textit{zero} and \textit{copying\_zero} with a zero sub-layer will block the feature learning. Consequently, there is a conflict between feature learning and function-preserving. Here function-preserving means the large model has exactly the same loss as the small model \citep{chen2015net2net,shen2022staged,wanglemon}, hence no loss spikes. 

\begin{table}[!htb]
	\centering
	\caption{Summary of initialization approaches in progressive training.}
	\vspace{-0.1cm}
	\begin{tabular}{c|c|c|c}
		&function-preserving&trainability&feature learning \\\hline
		copying &no&high&yes \\
		random&no&high&yes \\
		zero&yes&low&no \\
	\end{tabular}
	\label{tab:ways}
\end{table}

\begin{untitledbox}\textbf{Takeaway 2: }
	Zero initialization enables function-preserving expansion, but can block feature learning and slow convergence.
\end{untitledbox}

\subsection{Where to expand?}
\label{sec:where}
For zero-layer expansion, only \textit{random} initialization works; for one-layer expansion, \textit{random} and \textit{copying} both work. However, for multi-layer expansion, we must consider the ordering in depth expansion. We consider three variants of \textit{copying} initialization: suppose we expand 3 to 6 layers,
\begin{itemize}
	\item \textbf{[copying\_last]}, copying only the last layer, e.g. $[1,2,3]\to [1,2,3,3,3,3]$. 
	\item \textbf{[copying\_stack]}, copying and stacking all layers, e.g. $[1,2,3]\to [1,2,3,1,2,3]$. 
	\item \textbf{[copying\_inter]}, copying and interpolating all layers, e.g. $[1,2,3]\to [1,1,2,2,3,3]$.
\end{itemize}
We note that copying\_inter is adopted by \citep{chang2018multi,pan2024preparing,dong2020towards,qin2022elle}, as well as \citep{wanglemon} if some sub-layers are zeros; copying\_stack is adopted by \citep{gong2019efficient,li2020shallow,fu2023triple,kim2024solar}, as well as \citep{shen2022staged,du2024stacking} if some sub-layers are zeros.

\begin{figure}[!htb]
	\centering
	\includegraphics[width=0.4423\linewidth]{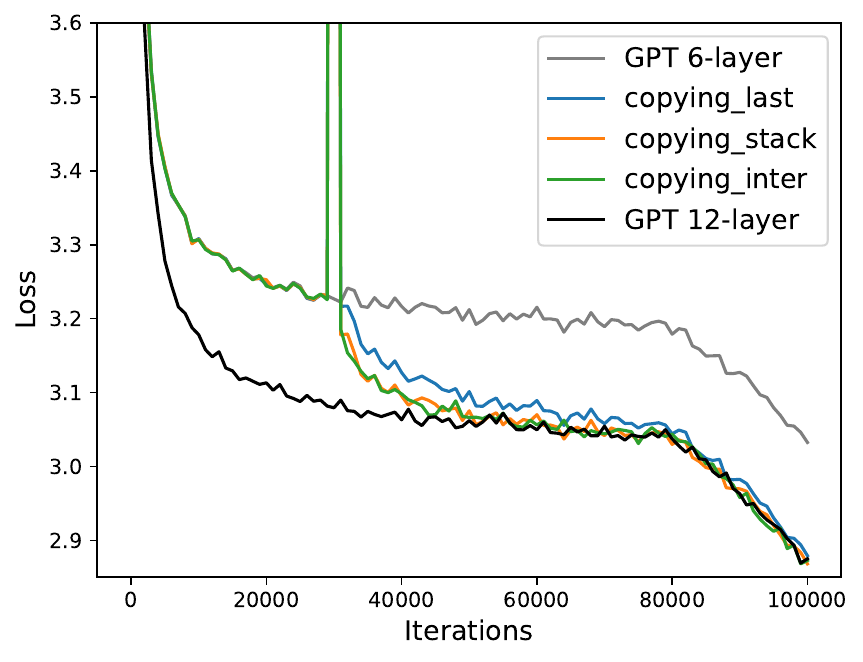}
	\includegraphics[width=0.4423\linewidth]{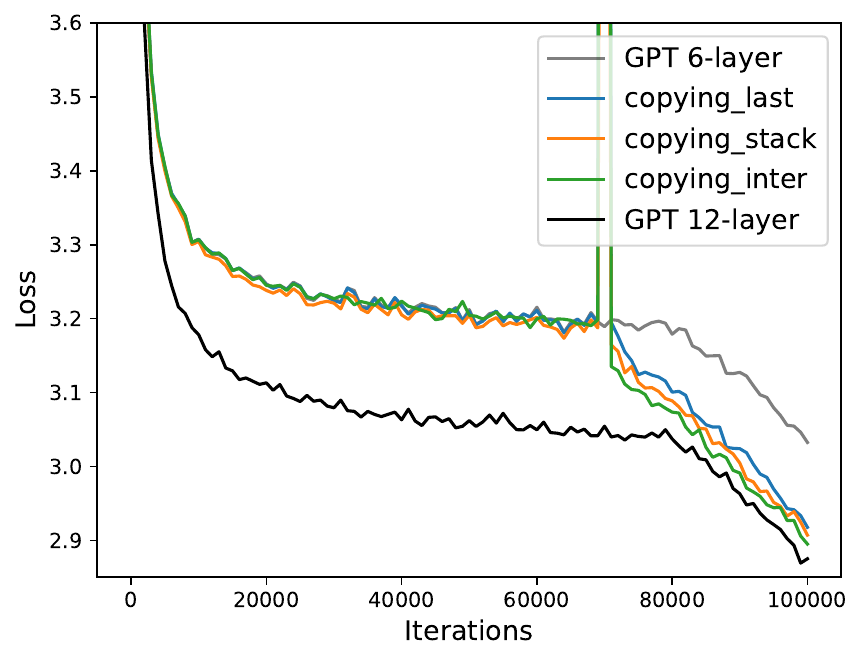}
	\caption{Convergence of multi-layer progressive training and fixed-size training. GPT2 with depth expansion at 30/70k iterations. }
	\label{fig:copy variants}
\end{figure}

To test these variants, we experiment with deeper models such as ResNet50 and GPT 6-layer in \Cref{fig:copy variants}. We observe that copying all layers is consistently better than only copying one layer (copying\_last), whereas copying\_inter and copying\_stack are almost indistinguishable.

\begin{untitledbox}\textbf{Takeaway 3: }
	\textit{Copying\_inter} and \textit{copying\_stack} are similarly performing for multi-layer depth expansion, but they are invalid for zero-layer depth expansion and equivalent for one-layer depth expansion (e.g. from $[1]\to [1,1,1,1,1,1]$).
\end{untitledbox}




\subsection{Effectiveness of progressive training}

\begin{figure}[!htb]
	\centering
	\includegraphics[width=0.45\linewidth]{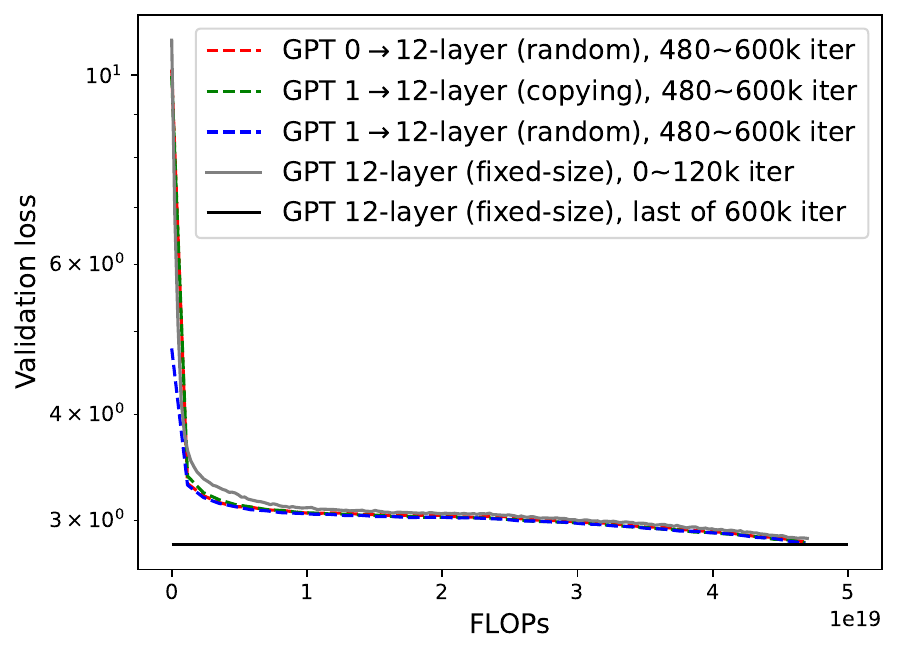}
	\includegraphics[width=0.517\linewidth]{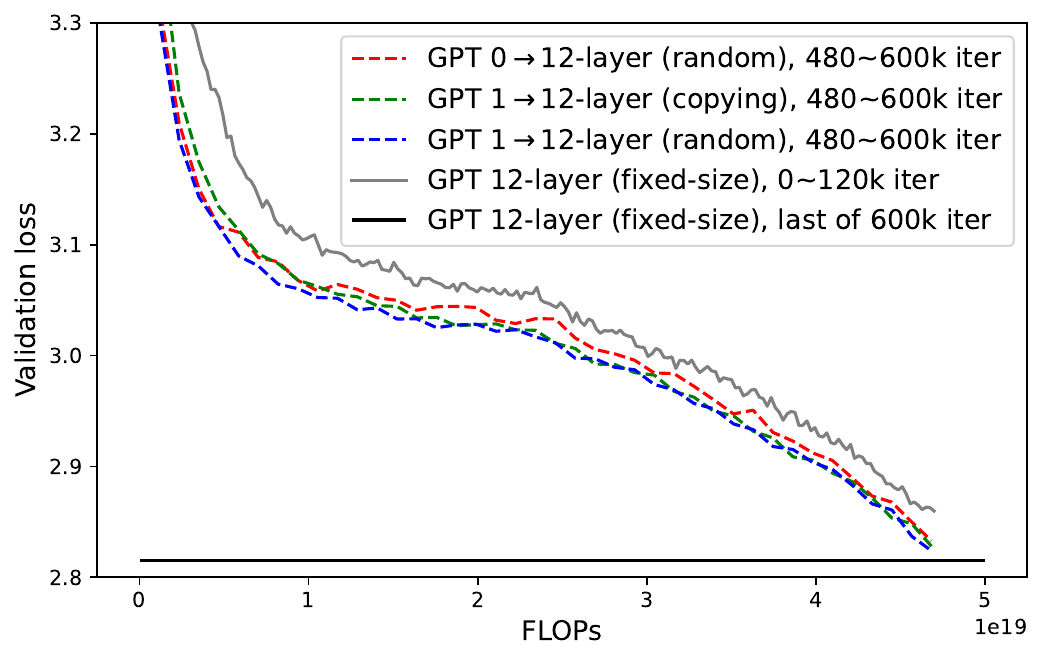}
	\label{fig:shorter}
	\caption{Comparing progressive training
		and fixed-size training via the grown model and
		target model (right plot is zoom-in of the left). Under the same compute budget, progressive training converges much faster than fixed-size training.}
\end{figure}

In \Cref{fig:long}, we have observed that progressive training can achieve similar (though sometimes slightly higher) validation loss than fixed-size training under the same number of iterations. To make sure that progressive training is actually effective, instead of just moving along the loss-compute tradeoff, we launch another fixed-size training with shorter training horizon in \Cref{fig:shorter}.

To be concrete, the depth expansion happens at $\tau=0.8T$, meaning the grown model from progressive training is trained for 120k iterations. Our second fixed-size training runs for the same 120k iterations using the same learning rate schedule.

Comparing the colored and grey curves, it is clear that the progressive training does inherit the progress from small model training and achieves significantly better loss, despite that the loss spike in \Cref{fig:long} seems to suggest all progress before model expansion is lost.

\section{A convergence theory for progressive training}
\label{sec:convex theory}

\subsection{Loss convergence of progressive training}
We analyze the convergence of  progressive training under convex and $G$-Lipschitz loss $L$.
In fact, although deep learning is non-convex, its training dynamics is similar to convex optimization \citep{anonymous2026convex,schaipp2025surprising,defazio2023learning,lee2019wide,bu2021dynamical,jacot2018neural,allen2019convergence,leclerc2020two,bugradient} and our analysis offers useful insights for the training recipe in \Cref{sec:recipe}.

We denote the small model before depth expansion as $\w_t$, the large model after depth expansion as $\W_t$, and the corresponding minima as $\w^*$ and $\W^*$.
For SGD, $\w_{t+1}=\w_t-\eta_{t+1}\frac{\partial L}{\partial \w_t}$, the classical analysis gives
\begin{align}
	\begin{split}
		\|\w_{t+1}-\w^*\|^2&= \|\w_{t}-\w^*\|^2
		-2\eta_{t+1} \langle \frac{\partial L}{\partial \w_t},\w_t-\w^*\rangle+\eta_{t+1}^2 \|\frac{\partial L}{\partial \w_t}\|^2
		\\
		&\leq \|\w_{t}-\w^*\|^2
		-2\eta_{t+1} (L(\w_t)-L(\w^*))+\eta_{t+1}^2 G^2
	\end{split}
	\label{eq:convexLipschitz_w}
\end{align}
and equivalently, for the large model training with the same learning rate schedule,
\begin{align}
	\begin{split}
		\|\W_{t+1}-\W^*\|^2&\leq \|\W_{t}-\W^*\|^2
		-2\eta_{t+1} (L(\W_t)-L(\W^*))+\eta_{t+1}^2 G^2 
	\end{split}
	\label{eq:convexLipschitz_W}
\end{align}

Now for the progressive training with depth expansion at $t=\tau$, we use telescoping sum \eqref{eq:convexLipschitz_w} from $t=0\to \tau-1$ and \eqref{eq:convexLipschitz_W} from $t=\tau\to T-1$ to obtain
\begin{align*}
	&\|\w_{\tau}-\w^*\|^2+\|\W_{T}-\W^*\|^2
	\\
	\leq &\|\w_{0}-\w^*\|^2+\|\W_{\tau}-\W^*\|^2+\sum_{t=0}^{T-1}\eta_{t+1}^2 G^2
	+2\sum_{t=0}^{\tau-1}\eta_{t+1}(L(\w^*)-L_{t})+2\sum_{t=\tau}^{T-1}\eta_{t+1}(L(\W^*)-L_{t})
\end{align*}
where $L_t=L(\w_t)$ for $t<\tau$, and $L_t=L(\W_t)$ otherwise.

Dividing by $2\sum_{t=0}^{T-1}\eta_{t+1}$ and re-arranging give
\begin{align*}
	\frac{\sum_{t=0}^{T-1}\eta_{t+1} L_{t}}{\sum_{t=0}^{T-1} \eta_{t+1}}
	\leq& \frac{\sum_{t=0}^{\tau-1}\eta_{t+1} L(\w^*)+\sum_{t=\tau}^{T-1}\eta_{t+1} L(\W^*)}{\sum_{t=0}^{T-1}\eta_{t+1}}
	+ \frac{G^2\sum_{t=0}^{T-1}\eta_{t+1}^2}{2\sum_{t=0}^{T-1} \eta_{t+1}}
	\\
	&+\frac{\|\w_0-\w^*\|^2-\|\w_{\tau}-\w^*\|^2}{2\sum_{t=0}^{T-1} \eta_{t+1}}
	+\frac{\|\W_{\tau}-\W^*\|^2-\|\W_T-\W^*\|^2}{2\sum_{t=0}^{T-1} \eta_{t+1}}
\end{align*}
On the left hand side, we apply an inequality from Corollary 11 in \cite{defazio2023optimal} to bound the averaged loss by last-iteration loss,
\begin{align*}
	\begin{split}
		L(\W_T^\text{progressive})&\leq\frac{\sum_{t=0}^{T-1}\eta_{t+1} L_{t}}{\sum_{t=0}^{T-1} \eta_{t+1}}
		+
		\frac{1}{2}
		\sum_{k=1}^{T-1}
		\frac{\eta_k}{\sum_{t=k+1}^{T} \eta_t}
		\left(
		\frac{1}{\sum_{t=k}^{T} \eta_t}
		\sum_{t=k}^{T} \eta_t^2 G^2
		\right)
	\end{split}
\end{align*}
where $\W_T^\text{progressive}$ is the last iterate of progressive training. Consequently,
\begin{align*}
	\begin{split}
		L(\W_T^\text{progressive})\leq&\frac{\sum_{t=0}^{\tau-1}\eta_{t+1} L(\w^*)+\sum_{t=\tau}^{T-1}\eta_{t+1} L(\W^*)}{\sum_{t=0}^{T-1}\eta_{t+1}}
		+ \frac{G^2\sum_{t=0}^{T-1}\eta_{t+1}^2}{2\sum_{t=0}^{T-1} \eta_{t+1}}
		\\
		&+\frac{\|\w_0-\w^*\|^2-\|\w_{\tau}-\w^*\|^2}{2\sum_{t=0}^{T-1} \eta_{t+1}}
		+\frac{\|\W_{\tau}-\W^*\|^2-\|\W_T-\W^*\|^2}{2\sum_{t=0}^{T-1} \eta_{t+1}}
		\\
		&+
		\frac{1}{2}
		\sum_{k=1}^{T-1}
		\frac{\eta_k}{\sum_{t=k+1}^{T} \eta_t}
		\left(
		\frac{1}{\sum_{t=k}^{T} \eta_t}
		\sum_{t=k}^{T} \eta_t^2 G^2
		\right)
	\end{split}
\end{align*}


We can easily recover the fixed-size large model training by setting $\tau=0$, and greatly simplify the bound:
\begin{align}
	\begin{split}
		L(\W_T^\text{fixed-size})\leq& L(\W^*)+ \frac{G^2\sum_{t=0}^{T-1}\eta_{t+1}^2}{2\sum_{t=0}^{T-1} \eta_{t+1}}
		+\frac{\|\W_0-\W^*\|^2-\|\W_T-\W^*\|^2}{2\sum_{t=0}^{T-1} \eta_{t+1}}
		\\
		&+
		\frac{1}{2}
		\sum_{k=1}^{T-1}
		\frac{\eta_k}{\sum_{t=k+1}^{T} \eta_t}
		\left(
		\frac{1}{\sum_{t=k}^{T} \eta_t}
		\sum_{t=k}^{T} \eta_t^2 G^2
		\right)
	\end{split}
	\label{eq:large from scratch}
\end{align}

Comparing the two bounds of fixed-size training and progressive training, we see that the last term cancels out and that there are two differences:
\begin{itemize}
	\item the minimum loss becomes an average of small model's minimum loss and large one's, since different models converge to different minima;
	\item the middle term (i.e. the gap between initial distance and final distance) becomes two phase, from $t=0\to \tau$ and from $t=\tau\to T$.
\end{itemize}

\subsection{Gap between progressive and fixed-size training}
Empirically, \eqref{eq:large from scratch} is very accurate as evidenced in Figure 2-5 of \cite{anonymous2026convex}, which indicates that the gap between the actual losses of progressive training and fixed-size training is properly characterized by the gap between upper bounds. Therefore, we subtract the progressive training bound from the fixed-size bound \eqref{eq:large from scratch} and would like the difference to be close to 0. 

We view the large model as the concatenation of a small model and extra parameters $\W_t=[\w_t,\x_t]$ for $t=0,\tau$, and simplify the analysis by assuming $\W^*=[\w^*,\x^*]$. We now obtain
\begin{align}
	\begin{split}
		&\frac{\sum_{t=1}^{\tau}\eta_t}{\sum_{t=1}^{T}\eta_t}(L(\w^*)-L(\W^*))+\frac{\|\x_{\tau}-\x^*\|^2-\|\x_0-\x^*\|^2}{2\sum_{t=1}^{T} \eta_{t}}.
	\end{split}
	\label{eq:gap}
\end{align}

From the viewpoint of large model, we can mathematically view the progressive training as projected gradient descent (PGD) that masks deeper layers to zero, followed by an instant teleportation of $\x_\tau$ from zero to good initialization, then continued with SGD. In words, the effectiveness of progressive training comes from both optimizers (PGD and SGD) and teleportation of deeper layers.

\begin{figure}[!htb]
	\vspace{-0.3cm}
	\centering
	\includegraphics[width=0.42493\linewidth]{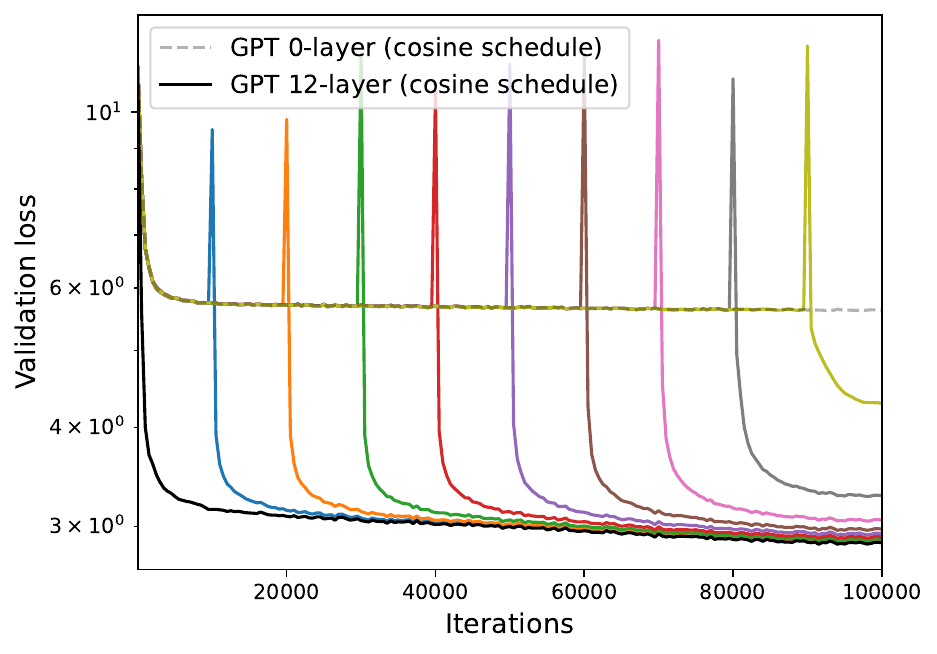}
	\includegraphics[width=0.41492\linewidth]{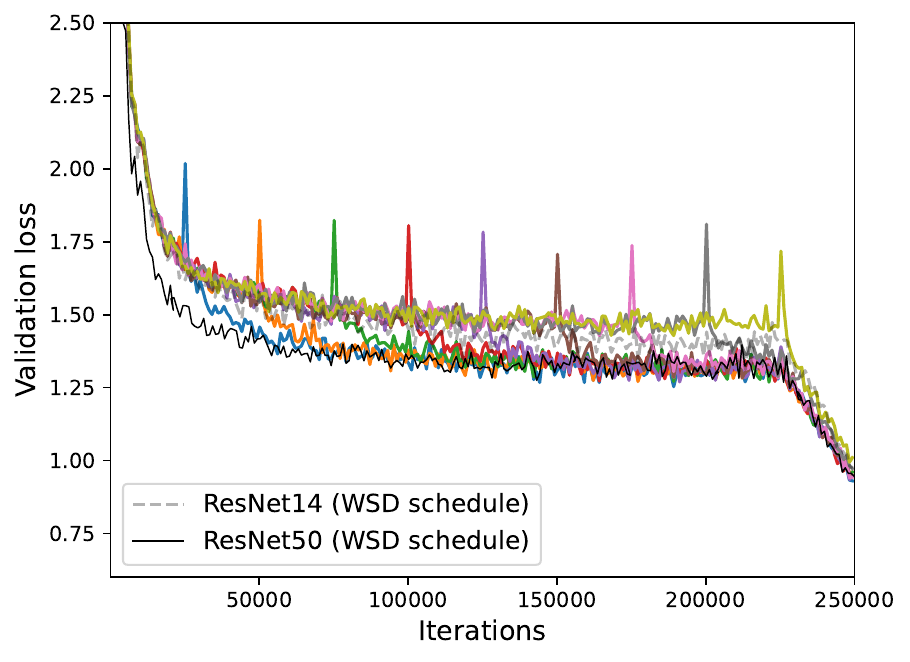}
	\includegraphics[width=0.42493\linewidth]{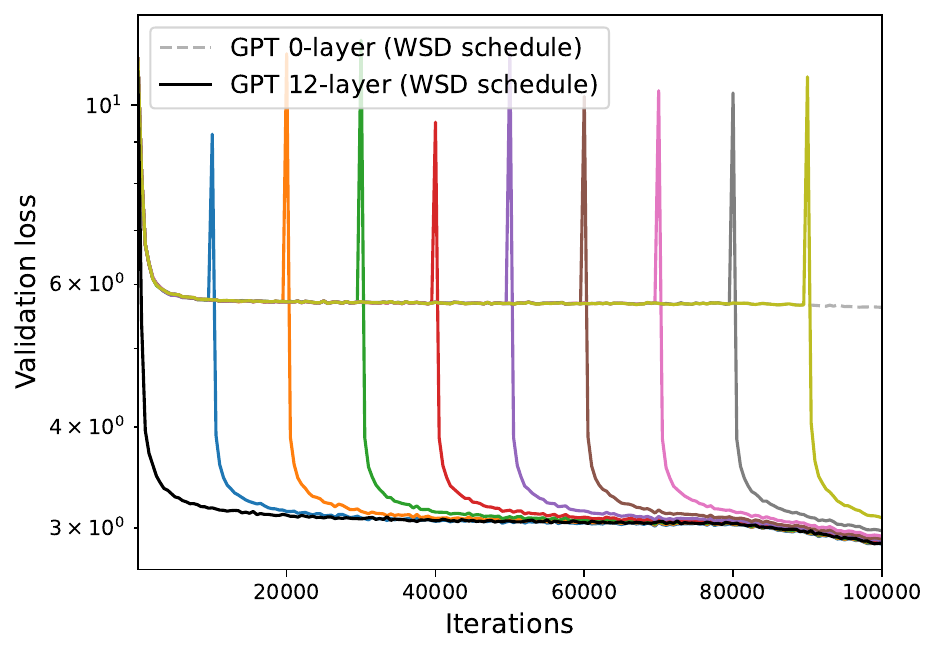}
	\includegraphics[width=0.41492\linewidth]{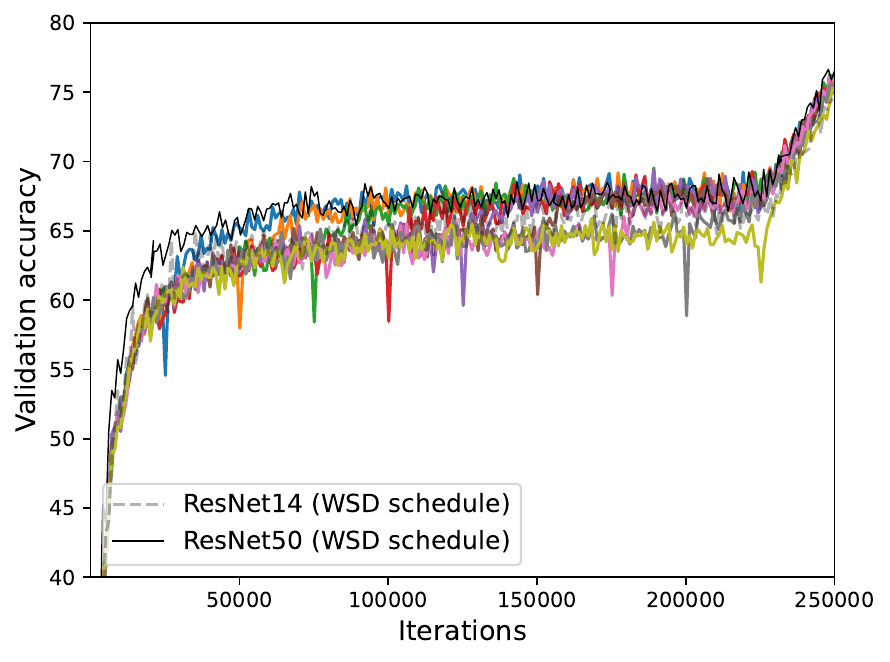}
	\includegraphics[width=0.40475\linewidth]{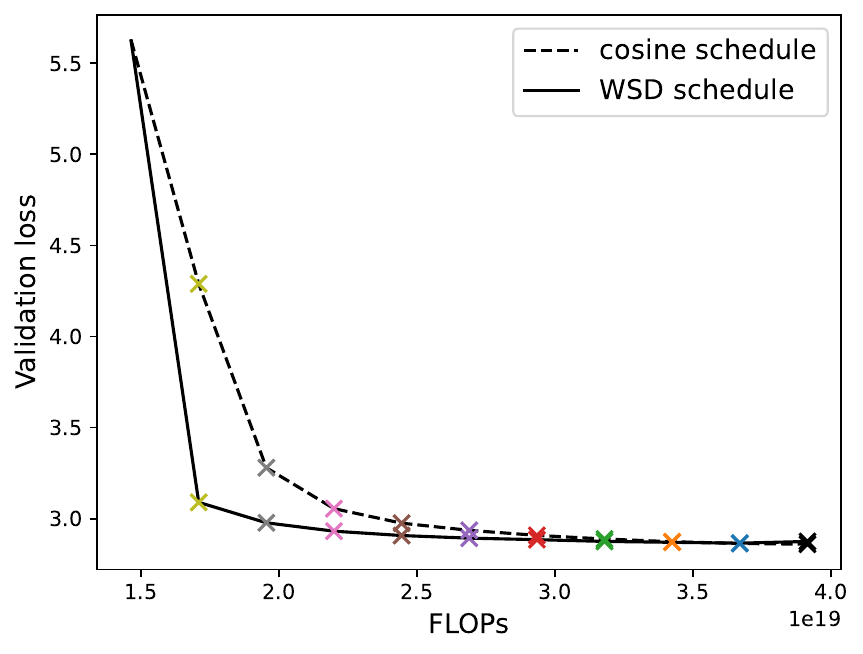}
	\includegraphics[width=0.4414\linewidth]{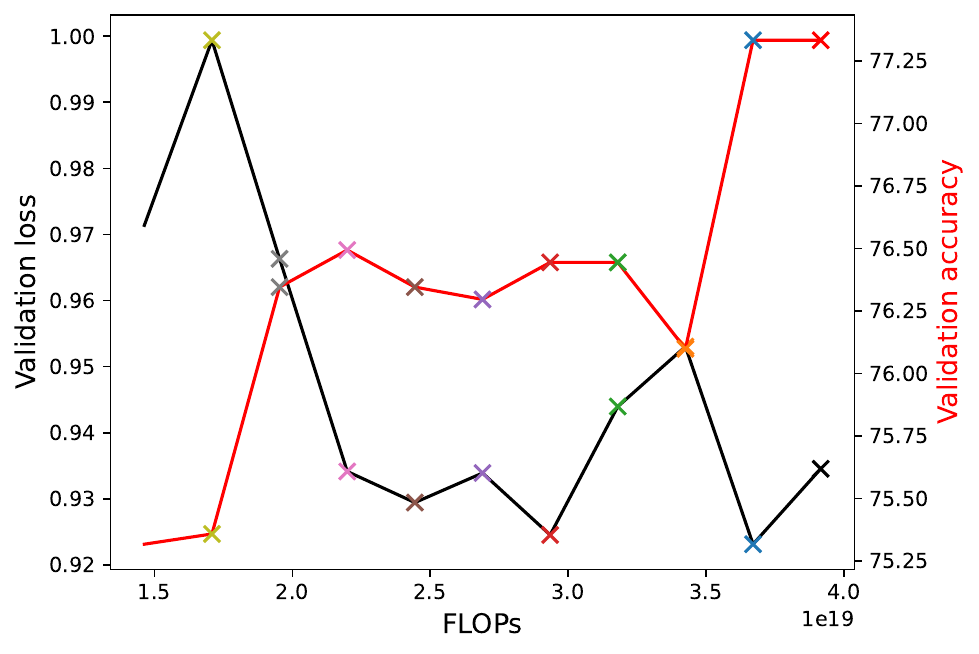}
	\caption{Performance of zero-layer progressive training and fixed-size training, where WSD schedule significantly enhances the progressive training. See one-layer results in \Cref{app:insights}.}
	\label{fig:lr schedule}
\end{figure}

Taking a closer look at \eqref{eq:gap}, we can optimize this difference via the following factors.
\begin{itemize}
	\vspace{-0.1cm}
	\item Initialization strategy of $\x_\tau$: given that $\x_0$ is randomly initialized, (1) if we randomly initialize new layers, then the second term is zero; (2) if we initialize better than random (e.g. copying), then the second term is negative and the difference is improved. This analysis is  visualized in \Cref{fig:resnetGPT_cosine_all}.
	\vspace{-0.1cm}
	\item Learning rate schedule $\eta_t$: to minimize $\frac{\sum_{t=1}^{\tau}\eta_t}{\sum_{t=1}^{T}\eta_t}$, we prefer smaller $\eta_t$ for $t\leq \tau$ than for $t>\tau$, contrary to learning rate decay but consistent with WSD schedule, where $\eta_t$ remains constant during most iterations (see \Cref{fig:lr schedule}).
\end{itemize}

To validate our insights on learning rate schedules, we experiment cosine and WSD schedules each with optimally tuned learning rate in \Cref{fig:lr schedule}. We expand small models to large models at every 10\% of total training horizon. For ResNet, the small model can still catch up with large model when $\tau\approx 0.8T$ under WSD schedule, but it fails to catch up around $\tau\geq 0.7T$ under cosine schedule; for GPT, the small model can catch up until $\tau\approx 0.8T$ under WSD schedule, but it fails around $\tau\geq 0.5T$ under cosine schedule.

\begin{untitledbox}\textbf{Takeaway 4: }
	Progressive training is essentially ``PGD + initialization of new layers + SGD'', whose effectiveness relies on good initialization (e.g. random or copying) and learning rate schedule (e.g. WSD).
\end{untitledbox}

\section{When to expand depth?}
\vspace{-0.2cm}
To determine the timing of depth expansion, we need to understand the \textit{mixing time}, which is the time until the loss of progressive training is close to the fixed-size model training. To be specific, we define $t_\text{mix}$ such that $L(\W_{\tau+t_\text{mix}}^\text{fixed-size})\approx L(\W_{\tau+t_\text{mix}}^\text{progressive})$. Clearly, if the mixing time is short, then we can expand the models at later stage and save more compute.

\subsection{Perspectives matter to mixing behaviors}
We highlight that the mixing behaviors of progressive training (e.g. Figures \ref{fig:lr schedule},
\ref{fig:lr schedule MOE},\ref{fig:Xto12/24}) have not been clearly observed in the literature, possibly due to the difference in perspectives of comparison.

\begin{figure}[!htb]
	\vspace{-0.2cm}
	\includegraphics[width=0.48\linewidth]{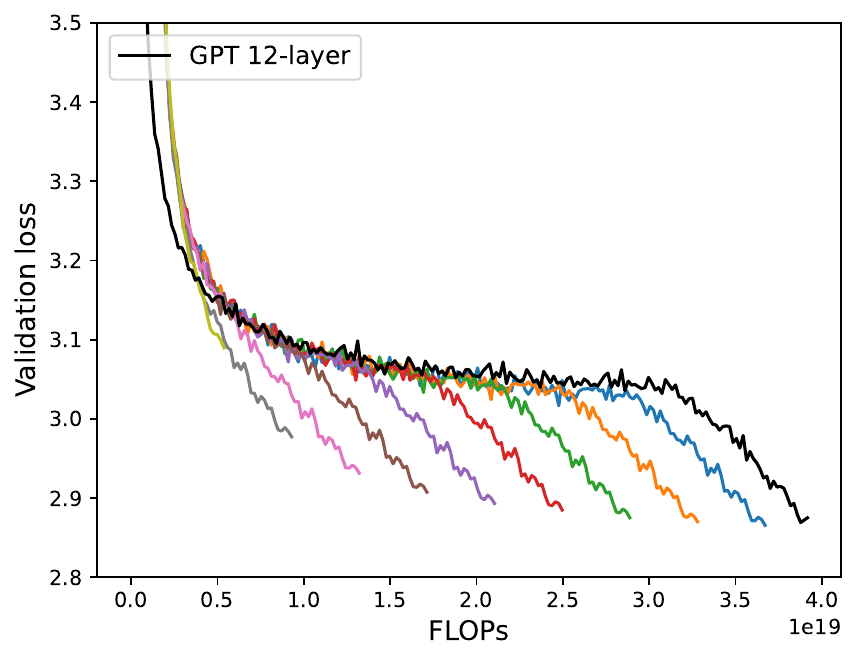}
	\includegraphics[width=0.5\linewidth]{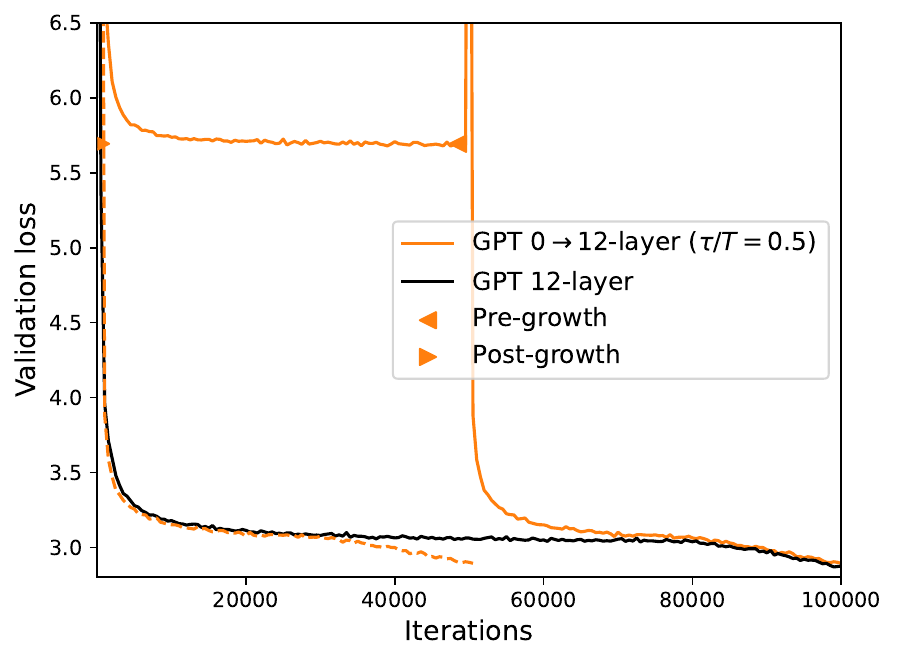}
	\vspace{-0.3cm}
	\caption{Different perspectives of \Cref{fig:lr schedule} to compare progressive training and fixed-size training. Left: only comparing the grown model and target model. Right: matching the pre-growth loss of source model to target model.}
	\label{fig:gpt_WSD_10runs_onlygrown}
\end{figure}

In figures of \citep{wang2023learning,chen2021bert2bert,pan2024preparing,du2024stacking}, the comparison is between the grown model and the target model, while our comparison is based on the entire training (source and grown models). Such a perspective omits the computational cost of small models and the stated speedup must be discounted in our context. We re-plot \Cref{fig:lr schedule} (LLM under WSD schedule) from their perspective and no longer observe the mixing behaviors in \Cref{fig:gpt_WSD_10runs_onlygrown}.

\begin{figure}[!htb]
	\vspace{-0.2cm}
	\includegraphics[width=0.49\linewidth]{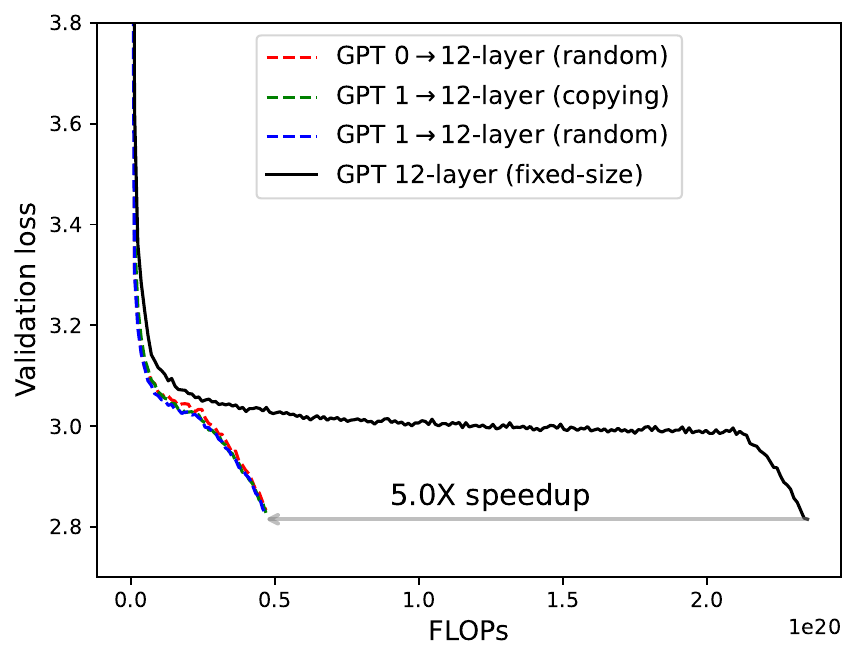}
	\includegraphics[width=0.49\linewidth]{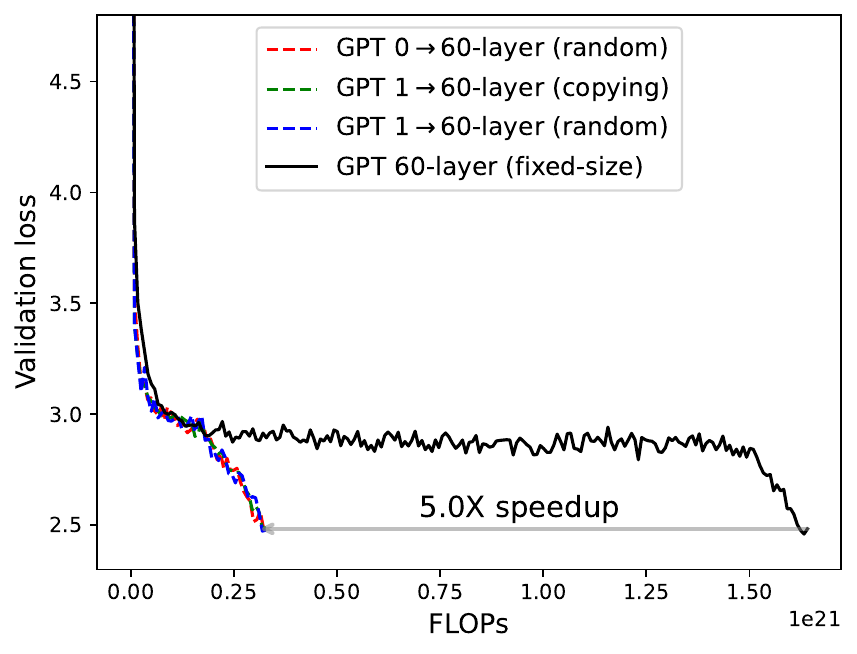}
	\vspace{-0.4cm}
	\caption{Comparing progressive training and fixed-size training via the grown model and target model of \Cref{fig:long}. Left: 124M model. Right: 7B model.}
\end{figure}

Another perspective is to ``overlay the loss curve for the grown model over the target model". \citep{shen2022staged} suggest that the convergence rate of grown model is ``the same as the target model trained from scratch", and that the training dynamics is preserved. However, we claim that our method significantly improves the training dynamics instead of preserving it, as shown in \Cref{fig:gpt_WSD_10runs_onlygrown} by comparing the dashed orange curve and the solid black curve.

\begin{untitledbox}\textbf{Takeaway 5: }
	The mixing behaviors of loss and training dynamics are the highlight of our depth expansion, and only observable via the comparison between the entire progressive training and the fixed-size training from scratch per-iteration.
\end{untitledbox}
\subsection{Sensitivity to $\tau$ under different schedules}
\vspace{-0.1cm}
Interestingly, in \Cref{fig:lr schedule}, the mixing time $t_\text{mix}(\tau)$ is highly sensitive to the timing of depth expansion $\tau$ for cosine schedule, but robust to $\tau$ for WSD schedule. For example, expanding GPT 1-layer at 10\% horizon (blue curve) and expanding at 60\% horizon (brown curve) both need $\approx 16B$ tokens or $30k$ iterations to mix with 12-layer training. However, expanding at 80\% horizon (grey curve) cannot mix well as the learning rate has decayed. The same patterns hold for ResNet as well.

As a consequence, we determine the timing of depth expansion as total duration of constant learning rates minus mixing time in \Cref{fig:long}. To be more precise, our WSD uses 2\% warmup, 10\% decay, and 528k iterations with constant learning rate. We subtract $\approx 40$k iterations of mixing time from it (derived from \Cref{fig:lr schedule} or the early stopped run), and set the timing of depth expansion at $t=$480k.

\begin{untitledbox}\textbf{Takeaway 6: }
	During the stable phase of WSD schedule, the mixing time is almost unaffected by the timing of depth expansion. Hence we can transfer the mixing time at early iterations until the decaying phase (see \Cref{fig:long}).
\end{untitledbox}

\section{Which to expand depth?}
\begin{figure}[!htb]
	\centering
	\includegraphics[width=0.481\linewidth]{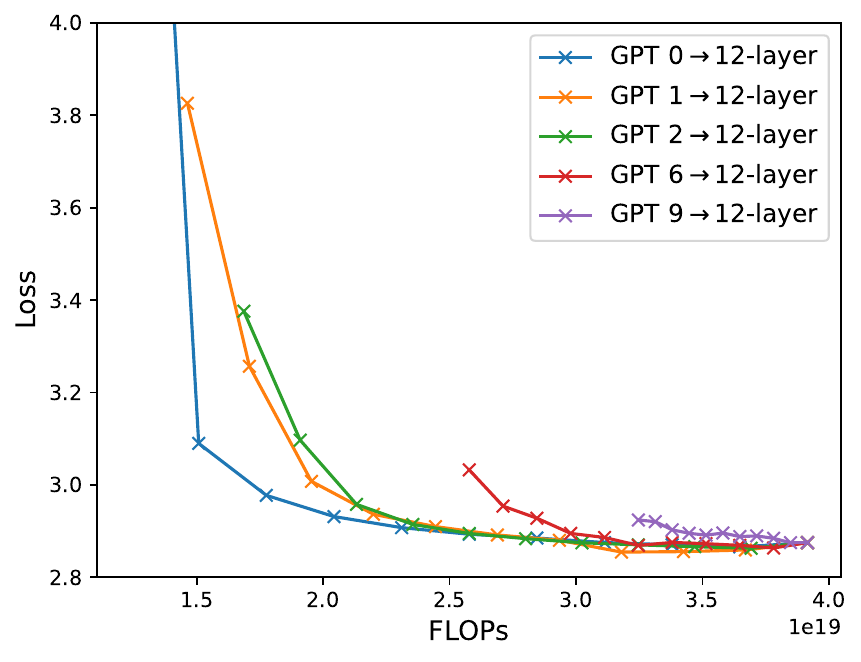}
	\includegraphics[width=0.481\linewidth]{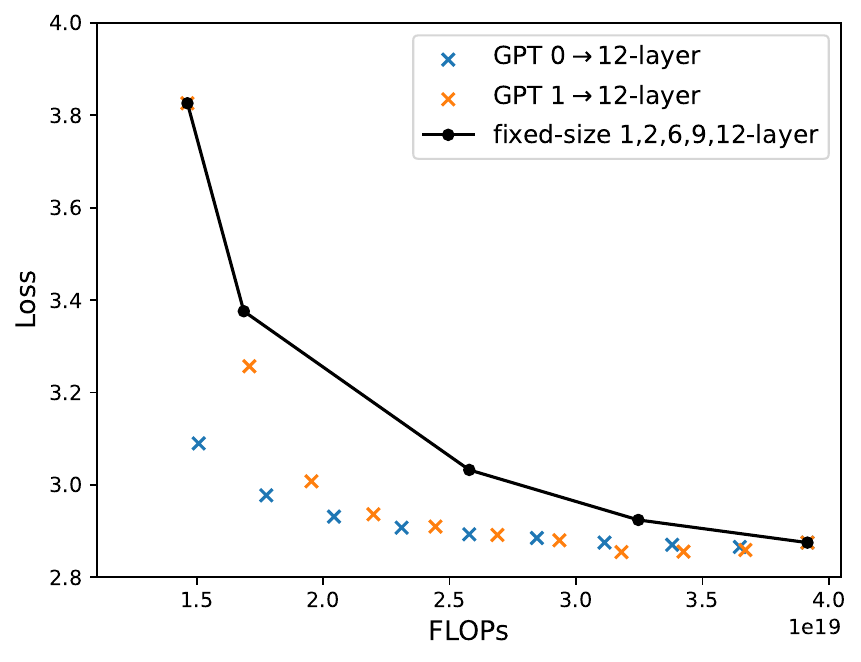}
	\includegraphics[width=0.481\linewidth]{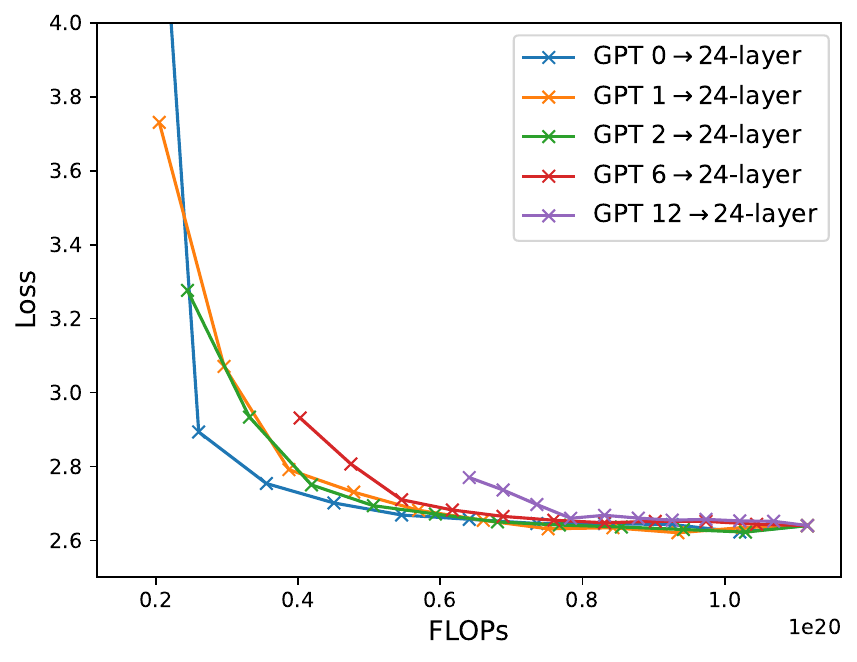}
	\includegraphics[width=0.481\linewidth]{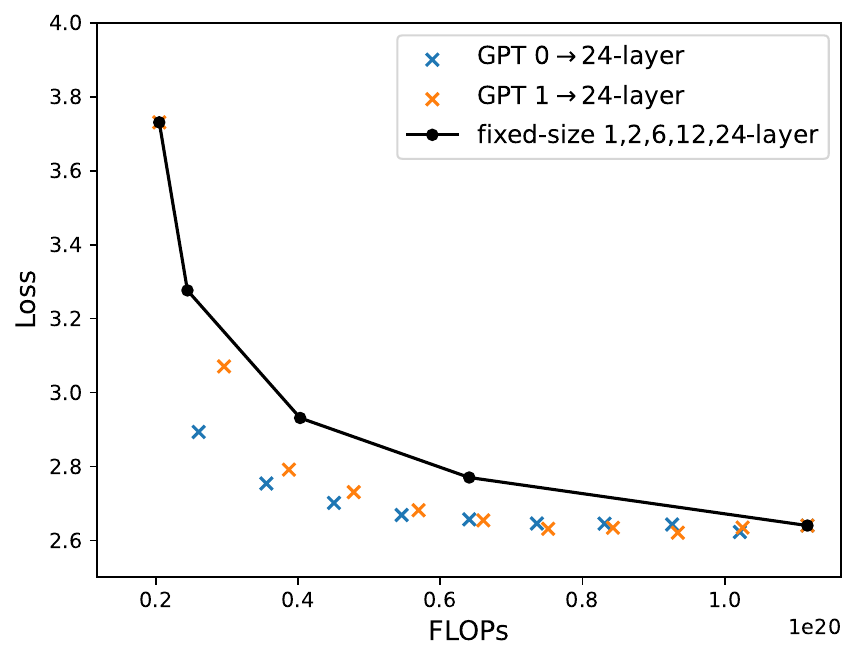}
	\includegraphics[width=0.481\linewidth]{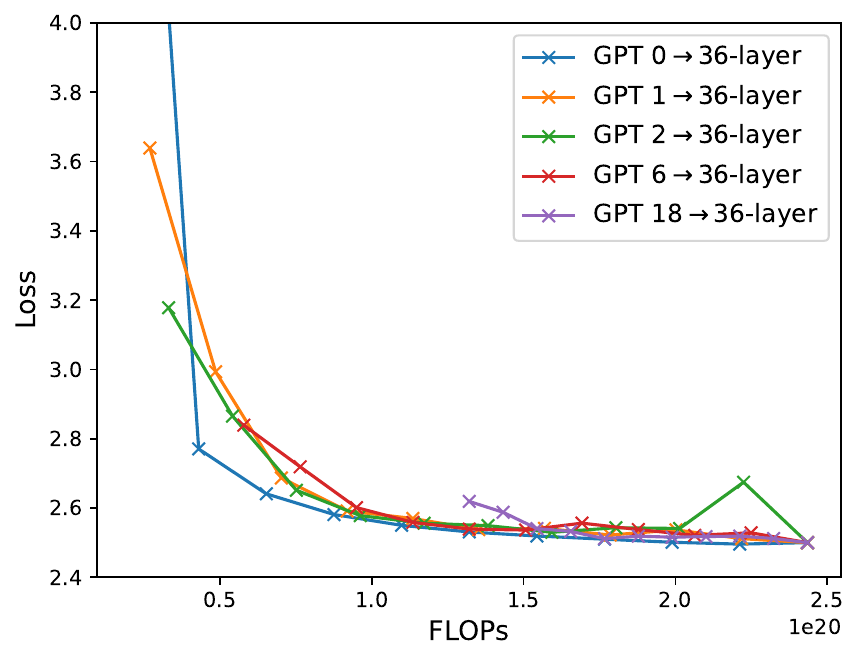}    
	\includegraphics[width=0.481\linewidth]{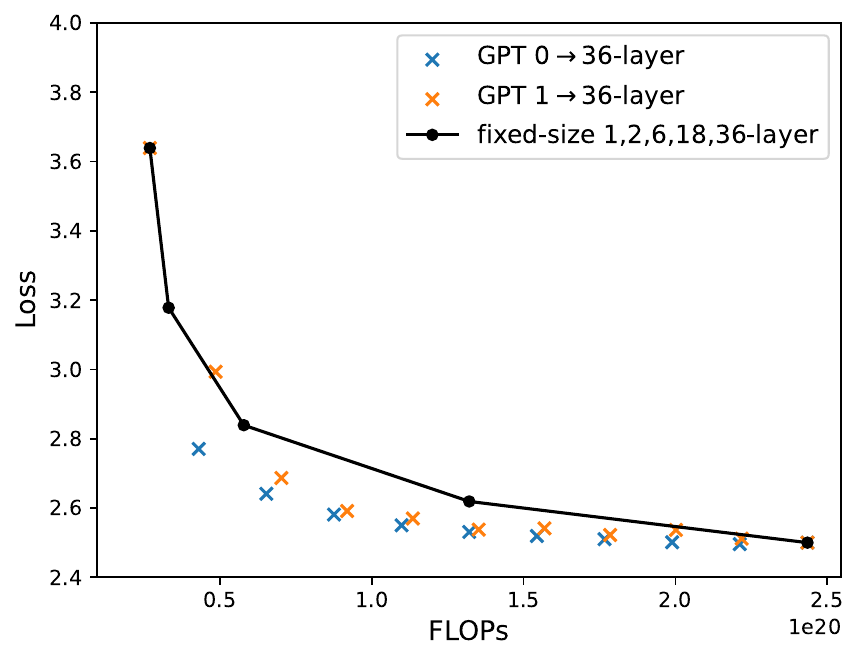}    
	\vspace{-0.3cm}
	\caption{Loss-compute tradeoff (validation loss v.s. FLOPs) of depth expansion from small models to $\{12,24,36\}$-layer GPT2 with $\{124M, 400M,1B\}$ parameters.}
	\label{fig:which to expand}
\end{figure}

\vspace{-0.2cm}
While we can expand the depth of any small model, we show the following through 150 runs (3 large model sizes, 5 small model sizes, 10 expansion times) in \Cref{fig:which to expand}.

As we see in \Cref{fig:which to expand}, the zero/one-layer progressive training almost captures the loss-compute tradeoff from a Pareto-optimal viewpoint, especially in contrast with the progressive training from more than 2 layers. Additionally, the latest timing of expansion that still allows the progressive training to mix with the fixed-size training is not sensitive to small model sizes. In other words, expanding from 1-layer or from 6-layer at $\tau/T\approx 0.6$ is similarly effective, but the latter is much more computationally expensive.

\begin{figure}[!htb]
	\centering
	\includegraphics[width=0.49\linewidth]{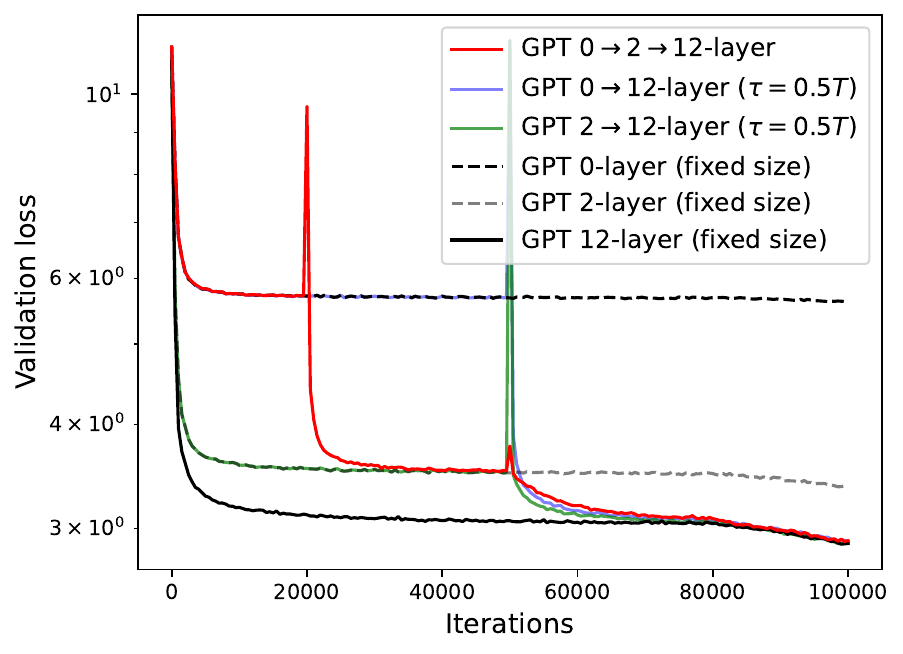}
	\includegraphics[width=0.46\linewidth]{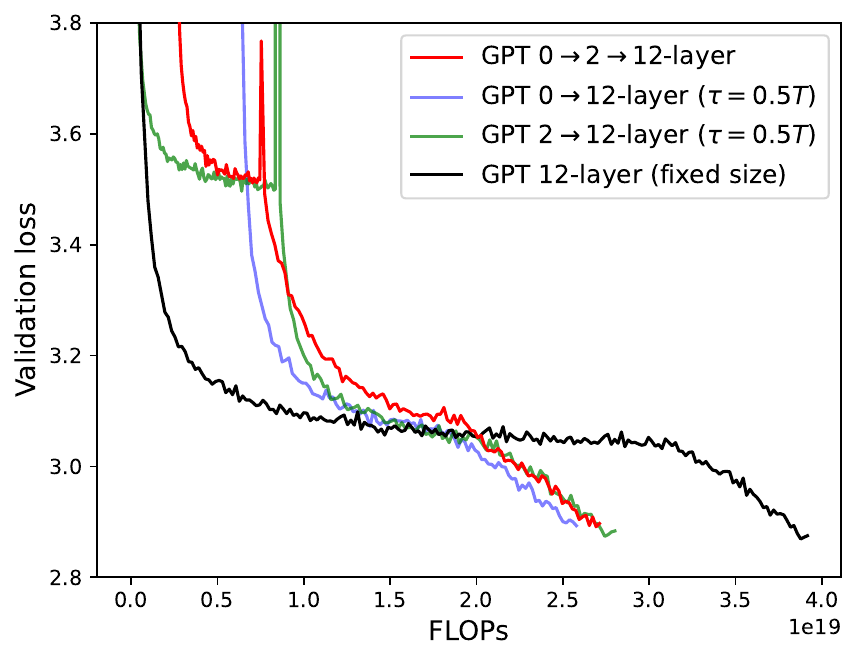}
	\caption{Multi-stage progressive training does not show better efficiency or loss, due to the mixing behaviors.}
	\label{fig:2 stages}
\end{figure}

\begin{figure}[!htb]
	\centering
	\includegraphics[width=0.4635\linewidth]{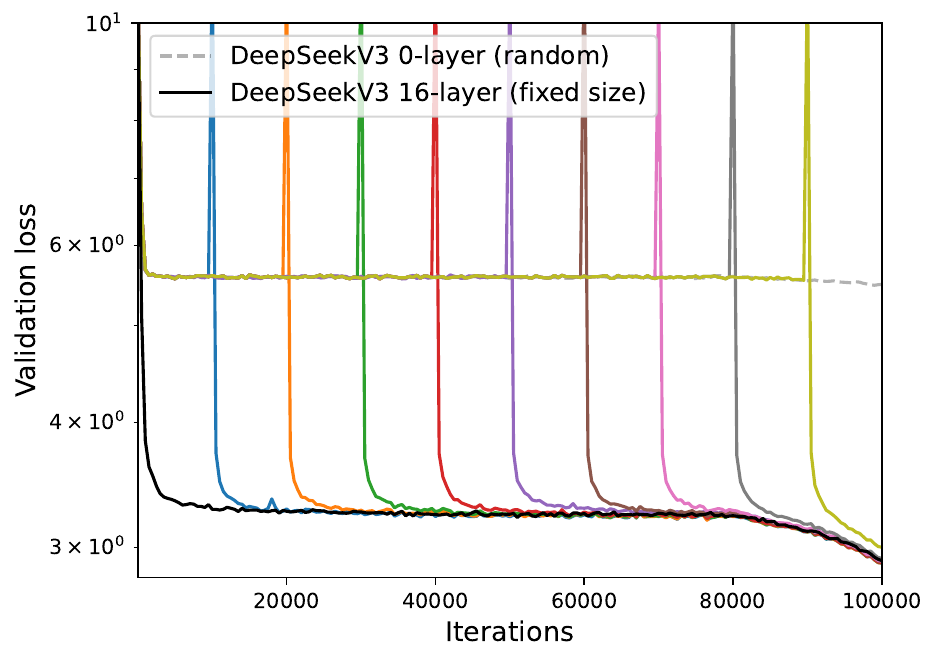}
	\includegraphics[width=0.4635\linewidth]{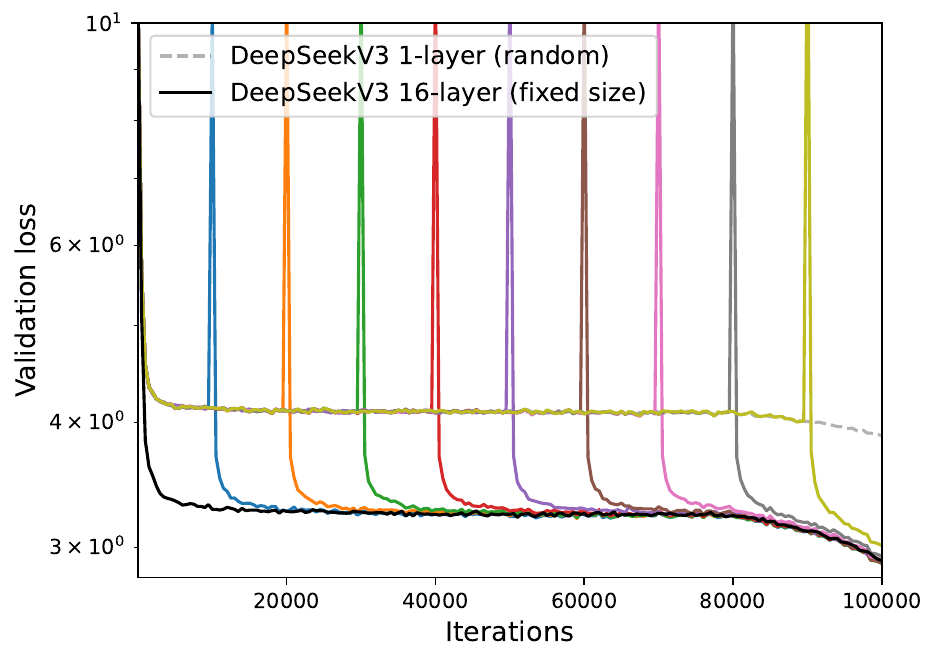}
	\includegraphics[width=0.47551\linewidth]{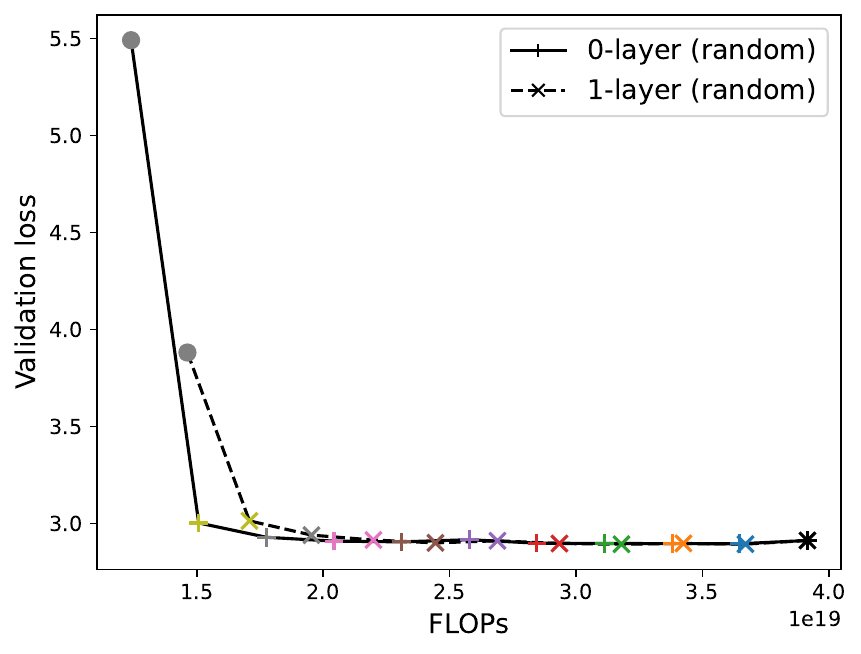}
	\vspace{-0.3cm}
	\caption{Convergence of zero/one-layer progressive training and fixed-size training for DeepSeekV3 (MoE), with random initialization of new layers. 
	}
	\label{fig:lr schedule MOE}
\end{figure}

Another insight from the loss-compute tradeoff is that, it suffices to use single-stage expansion, i.e. we do not need multi-stage expansion such as $0\to 2\to 12$ (if our target model is 12-layer). This can be explained by the mixing behaviors as $0\to 2\to 12$ can be decomposed to two single-stage expansions $0\to 2$ and $2\to 12$. Therefore, the loss curve of $0\to 2\to 12$ will mix with those of $0\to 2$ and $2\to 12$, in the first stage and the second stage, respectively. As a result in \Cref{fig:2 stages}, all runs lead to similar final losses, and multi-stage progressive training does not show improved efficiency: the multi-stage run has almost the same FLOPs as the 2-layer progressive training, but worse than the 0-layer one.

\begin{untitledbox}\textbf{Takeaway 7: }
	It is the most computationally efficient to 
	(I) scale up from the zero/one-layer models and (II) scale up only once, i.e. use single-stage progressive training.
\end{untitledbox}

\section{Deep progressive training recipe}
\label{sec:recipe}
We summarize our progressive training recipe, leveraging the theoretical insights and empirical evidences in previous sections.
\begin{enumerate}
	\item Train zero/one-layer model and then expand depth by random initialization\footnote{Alternatively, train one-layer model and expand by copying , e.g. $\w\to [\w,\w,\w]$.}. 
	\item Train models with Muon-NSGD (or other muP-scaled optimizers) and employ the same hyperparameters before and after depth expansion.
	\item Train models with WSD learning rate schedule and expand depth during the stable phase. 
	\item The timing of depth expansion $\tau$ (or equivalently the mixing time $t_\text{mix}$) can be determined by two small-scale runs: one fixed-size training and one progressive training ($\tau$ at the end of warmup), both early stopped when their losses mix.
\end{enumerate}
We further validate our recipe with MoE in addition to \Cref{fig:resnetGPT_cosine_all}, and we observe the same patterns as dense models such as the mixing behaviors. We emphasize that our approach is different and orthogonal to existing works that upcycle MoE \citep{he2024upcycling}, which scale up a small dense model to a large MoE without increasing the depth, rather than a shallow MoE to a deep MoE. This upcycling approach has reported some negative results \citep{muennighoffolmoe,komatsuzakisparse,nakamuradrop,liewscaling,wei2024skywork}, because the grown MoE becomes worse than the MoE trained from scratch after a few hundred billion tokens.

\begin{untitledbox}\textbf{Takeaway 8: }
	Zero/one-layer progressive training is effective on various vision and language (dense and MoE) model architectures.
\end{untitledbox}

\newpage
\section{Conclusion}
We show that zero/one-layer progressive training can significantly accelerate large-scale training and retain almost all the performance due to the mixing behaviors, if it is equipped with good initialization method and learning rate schedule. This work demonstrates the power of theoretical insights into progressive training, drawing tools from feature learning and optimization theory. We expect future works to continue pushing the efficiency frontier, e.g. by scaling up both width and depth.

%% file: appendix.tex
\clearpage
\section{Depth expansion approaches}
\label{app:expansion methods}

\subsection{Applicability of random, copying, and zero initialization}
We summarize the depth expansion approaches with respect to the depth of source models.

\begin{table}[!htb]
    \centering
    \caption{Applicability of depth expansion approaches. Merged cells indicate that multiple approaches are equivalent for a source model.}
    \begin{tabular}{c|c|c|c}
         &zero-layer&one-layer&multi-layer  \\\hline
         random&Yes&Yes&Yes\\ \hline
         copying\_inter&\multirow{3}{*}{No}&\multirow{3}{*}{Yes}&Yes\\ \cline{1-1}\cline{4-4}
         copying\_stack&&&Yes\\ \cline{1-1}\cline{4-4}
         copying\_last&&&Yes\\ \hline
         zero&Yes&Yes&Yes\\ \hline
    \end{tabular}
    \label{tab:placeholder}
\end{table}

We note that zero-layer or one-layer depth expansion significantly simplifies the copying approaches, as there is no ordering such as stacking or interpolation. In Appendix H of \cite{du2024stacking}, the search space of ordering is enormous but necessary, since a proper ordering indeed improves the progressive training. This aligns with our results that copying all layers is better than copying only the last layer. However, it is debatable which of copying\_inter and copying\_stack is more advantageous, e.g. \cite{pan2024preparing} claims that copying\_inter is more stable but \cite{du2024stacking} demonstrates that copying\_stack converges better. 

We highlight that such debate is completely avoided for zero-layer or one-layer progressive training.

\subsection{Copying\_zero initialization}
Completely zero initialization renders new layers not trainable, despite the depth expansion is function-preserving. It has been shown in the literature that copying with partially zero initialization has better trainability and is still function-preserving.

There are two known methods that copy all sub-layers except (1) the normalization sub-layers are initialized as zeros  \cite{shen2022staged}, or (2) the last linear sub-layer is initialized as zero (\cite{wanglemon,tan2024dlo,wu2024llama} and \cite{du2024stacking} $G_\text{zero}$), or masked with zeros \cite{yaomasked}. 
These approaches are termed as copying\_zeroN and copying\_zeroL, which enforce zero output of a new layer and are function-preserving.

We experiment in the same setting as in \Cref{fig:resnetGPT_cosine_all}. Empirically, copying\_zeroN has weak trainability, but copying\_zeroL converges as fast as copying (without any zero sub-layers) and avoids any loss spike unlike copying.
\begin{figure}[!htb]
    \centering
    \includegraphics[width=0.493\linewidth]{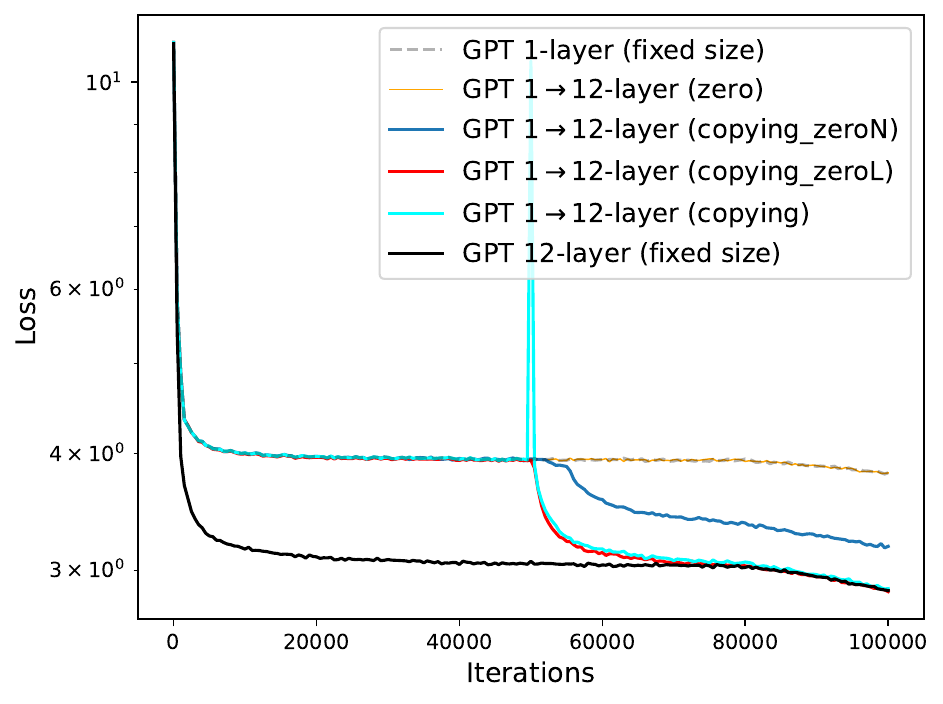}
    \caption{Convergence of one-layer progressive training and fixed-size training, with 2 different approaches of copying\_zero initialization.}
\end{figure}

\subsection{On the performance and ordering of random initialization}
\label{app:where}
We observe that random initialization of new layers works well on GPT2 and MoE, but slightly less so on ResNet. We think the reason is the location of insertion: for GPT2 and MoE, the insertion is at the bottom, e.g. [1,2,3] to [1,2,3,R,R,R]; however, for ResNet with 4 stages, the insertion is intermittent since the model architecture is inhomogeneous, e.g. ResNet26 to ResNet50 is like [[1],[2],[3],[4]] to [[1,R],[2,R], [3,R,R,R],[4,R]].

On a related note, we analyze the ordering of random initialization. We train 6-layer or 12-layer GPT2 and expand the depth at $\tau=0.1T$. We insert randomly initialized layers on top or bottom of old layers, i.e. $[1,2,3,4,5,6]\to [R,...,R,1,...,6]$ or $[1,2,3,4,5,6]\to [1,...,6, R,...,R]$.
\begin{figure}[!htb]
    \centering
    \includegraphics[width=0.45\linewidth]{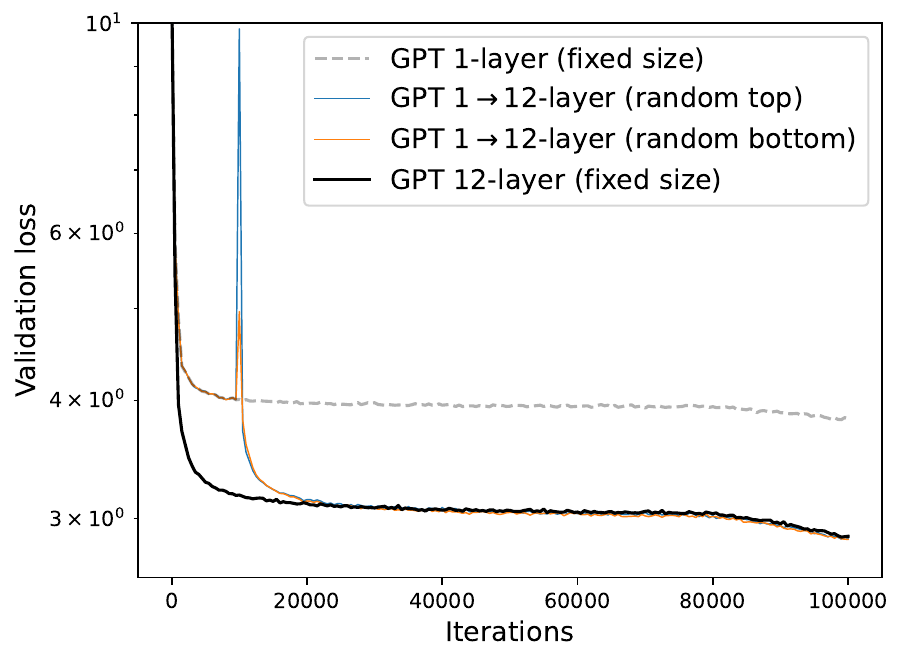}
    \includegraphics[width=0.43\linewidth]{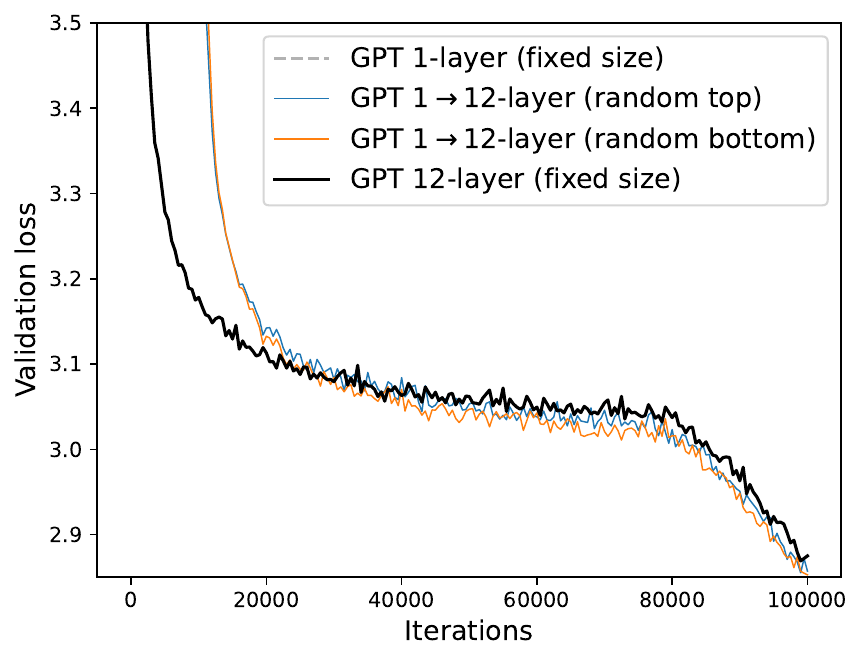}
\\
    \includegraphics[width=0.45\linewidth]{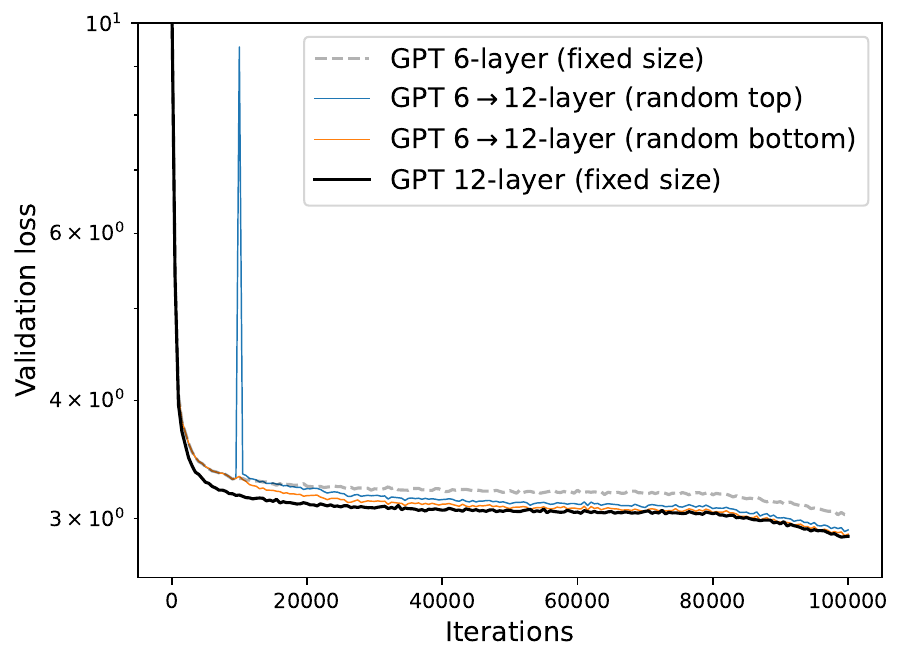}
    \includegraphics[width=0.43\linewidth]{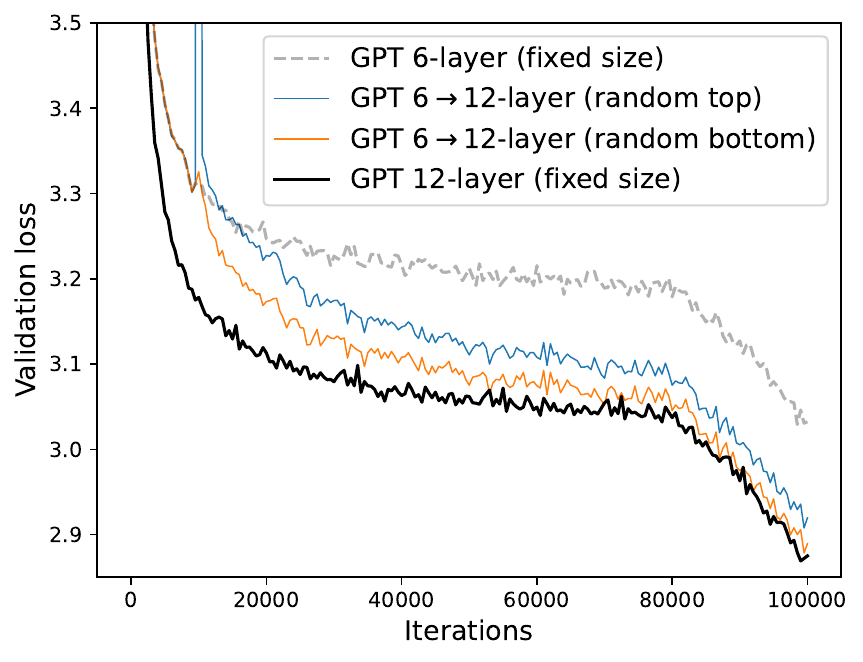}
    \caption{Convergence of progressive training and fixed-size training with different insertion of random initialization (right plot is zoom-in of the left). Adding new layers at the bottom of old layers works best, with much smaller loss spikes.}
\end{figure}

We highlight that appending new layers at the bottom is empirically the best approach, and has much smaller loss spike than inserting at the top. However, such choice is completely avoided for zero-layer progressive training.

\section{Experiment settings}
\label{app:settings}
Default batch size is 512 and decaying is 20\% for WSD schedule, except for the long runs in \Cref{fig:long} where decaying is 10\% and batch size=512 for 1B models and 64 for 7B models. For WSD schedule, the learning rate is 0.01 as shown to be optimal in \Cref{fig:mup table} (only here GPT2 are trained for 25k iterations); for cosine schedule, the learning rate is 0.05.

Regarding the optimizer, Muon-NSGD uses Muon \cite{jordan2024muon} and NSGD as in \cite{boreiko2025towards} in order to orthogonalize all tensors: denoting $\mathbf{W}$ as a layer's parameter, we apply
\begin{align*}
\textup{Muon: } & \mathbf{W}_{t+1}=(1-\eta\lambda)\mathbf{W}_t-\eta\cdot\textup{NS}(\mathbf{m}_t)
\\
\textup{NSGD: } & \mathbf{W}_{t+1}=(1-\eta\lambda)\mathbf{W}_t-\eta\cdot \mathbf{m}_t/\|\mathbf{m}_t\|_2
\end{align*}
where NS is the Newton-Schulz matrix iteration, $\mathbf{m}_t$ is the momentum, and $\lambda$ is the weight decay.

For GPT2 models, we always keep n\_embd/n\_head=64. Different depth uses different n\_head: full 12-layer uses 12 heads; full 24-layer uses 16 heads; full 36-layer uses 20 heads; full 60-layer uses 48 heads.

For LLMs other than GPT2, we train 0.3B variants. These LLMs have different designs. For example,
\begin{itemize}
    \item weight tying: enabled for GPT2 and Qwen3, and disabled for others.
    \item sparsity: GPT2, LLAMA3, Qwen3 are dense; DeepSeek V3 and Mixtral are sparse MoE.
    \item attention mechanism: GPT2 uses MHA (Multi-Head Attention \cite{vaswani2017attention}); DeepSeek V3 uses MLA (Multi-Head Latent Attention \cite{liu2024deepseek2}); others use GQA (Grouped-Query Attention \cite{ainslie2023gqa}).
    \item position embedding: GPT2 uses absolute embedding and the others use rotary embedding \cite{su2024roformer}).
    \item normalization: GPT2 uses layer normalization and the others use RMSNorm \cite{zhang2019root}.
    \item activation function: GPT2 uses GeLU \cite{hendrycks2016bridging} and the others use SwiGLU \cite{shazeer2020glu}.
\end{itemize}


For LLAMA3, hidden size=1024, intermediate size=2048, num attention heads=16, num key value heads=8, no weight tying; for Qwen3, hidden size=1024, intermediate size=2048, num attention heads=16, num key value heads=8, weight tying is used; for DeepseekV3, hidden size=512, intermediate size=1024, num attention heads=8, num key value heads=4, MLA and sparse attention is used; Mixtral, hidden size=512, intermediate size=1024, num attention heads=8, num key value heads=4.

\section{Additional experiments}
\label{app:insights}

\subsection{Mixing behaviors across model sizes}
\begin{figure}[!htb]
    \centering

    \includegraphics[width=0.32\linewidth]{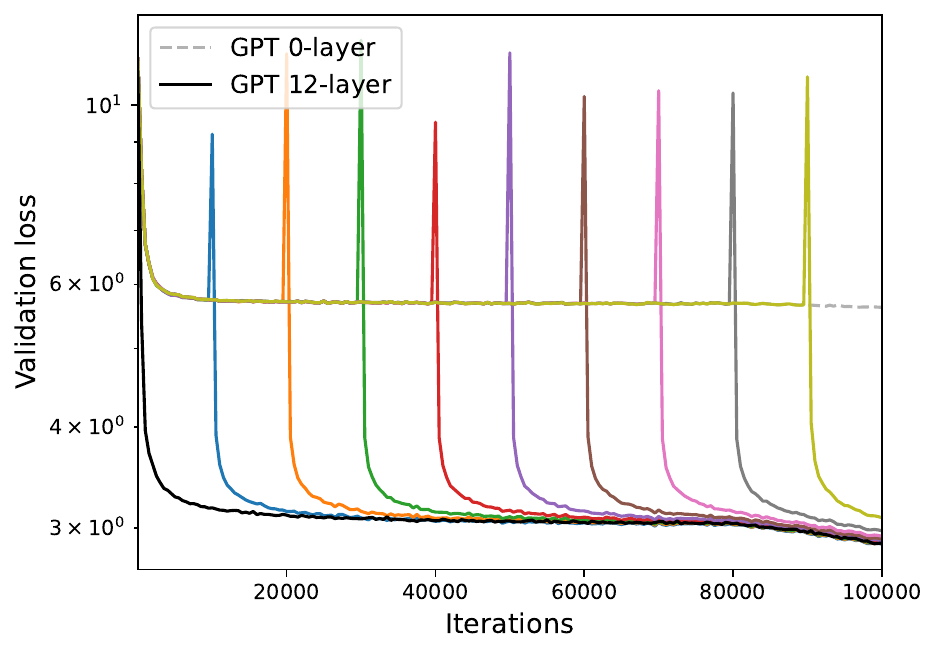}
    \includegraphics[width=0.32\linewidth]{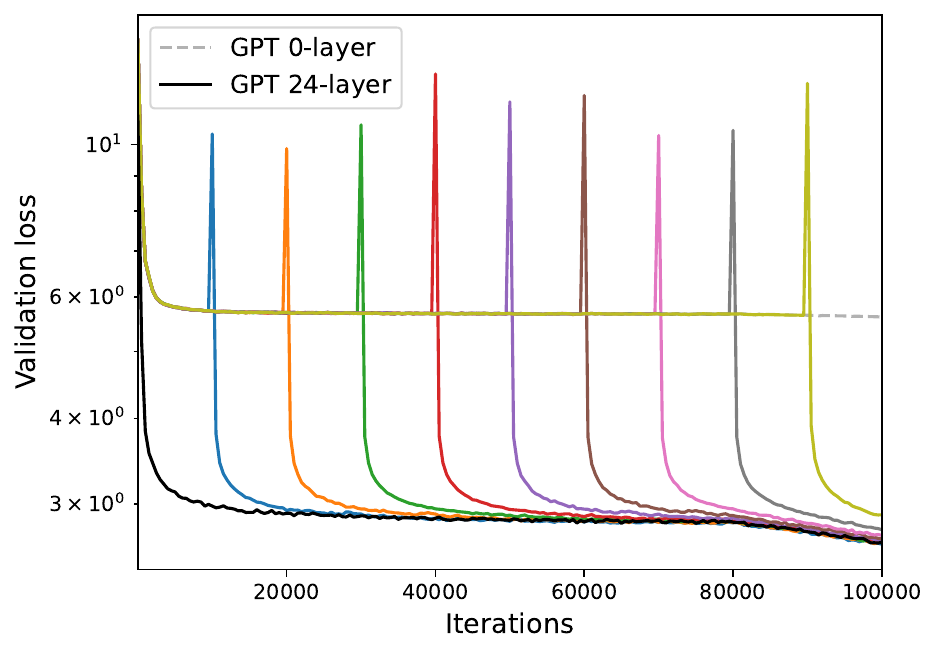}
    \includegraphics[width=0.32\linewidth]{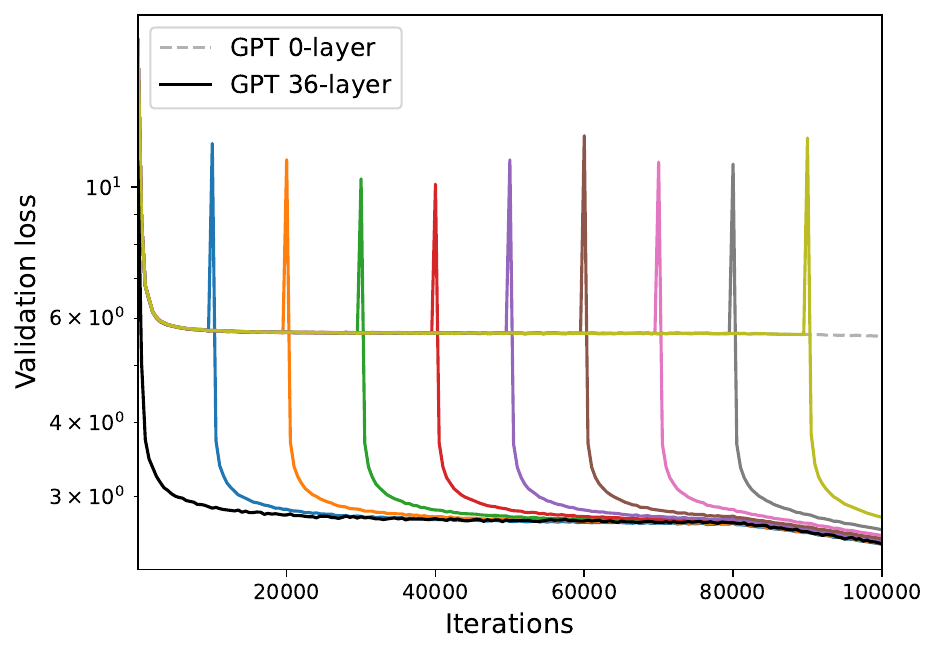}
    
    \includegraphics[width=0.32\linewidth]{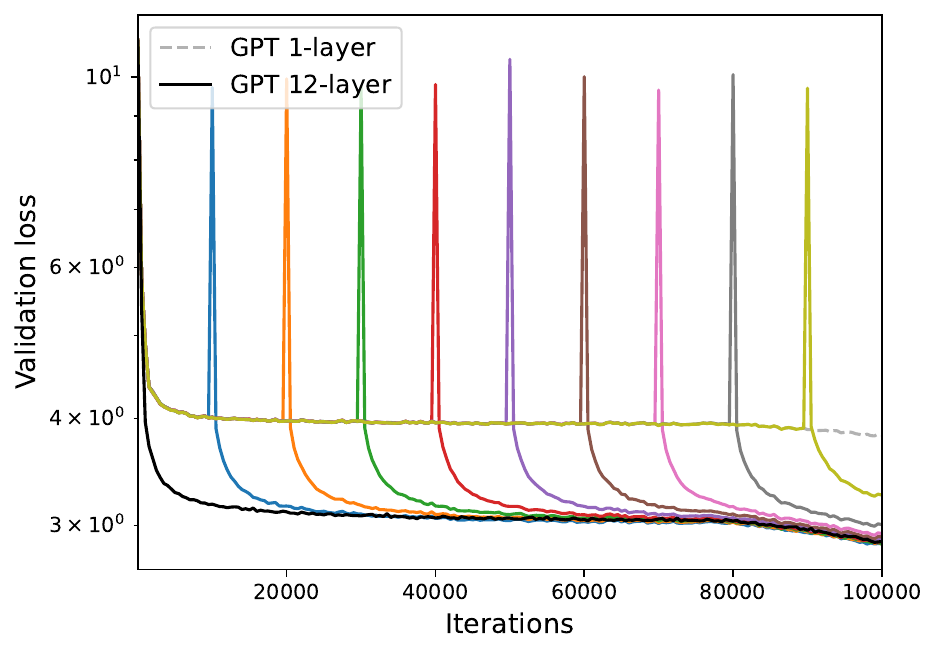}
    \includegraphics[width=0.32\linewidth]{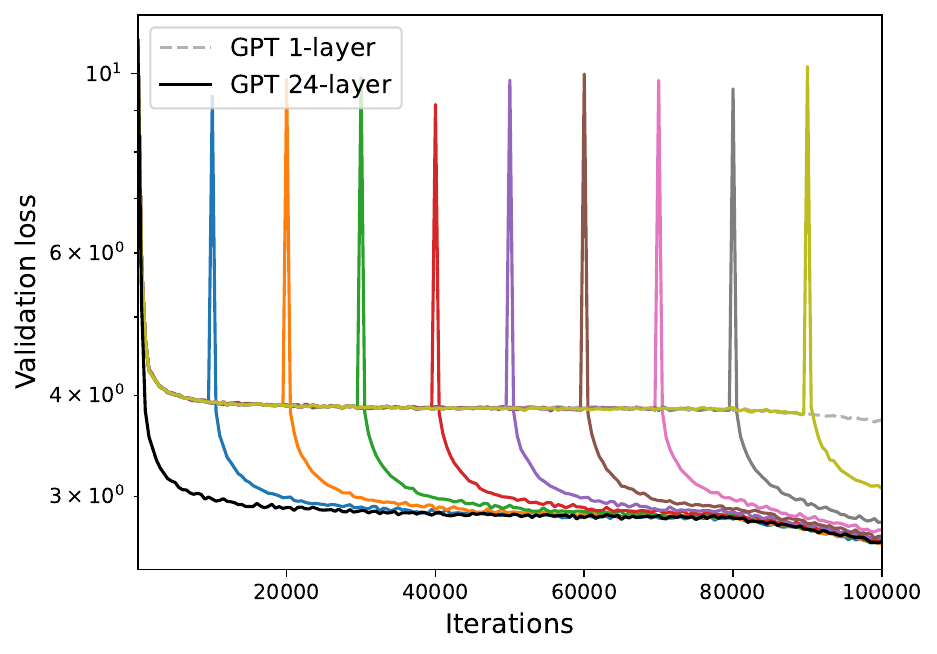}
    \includegraphics[width=0.32\linewidth]{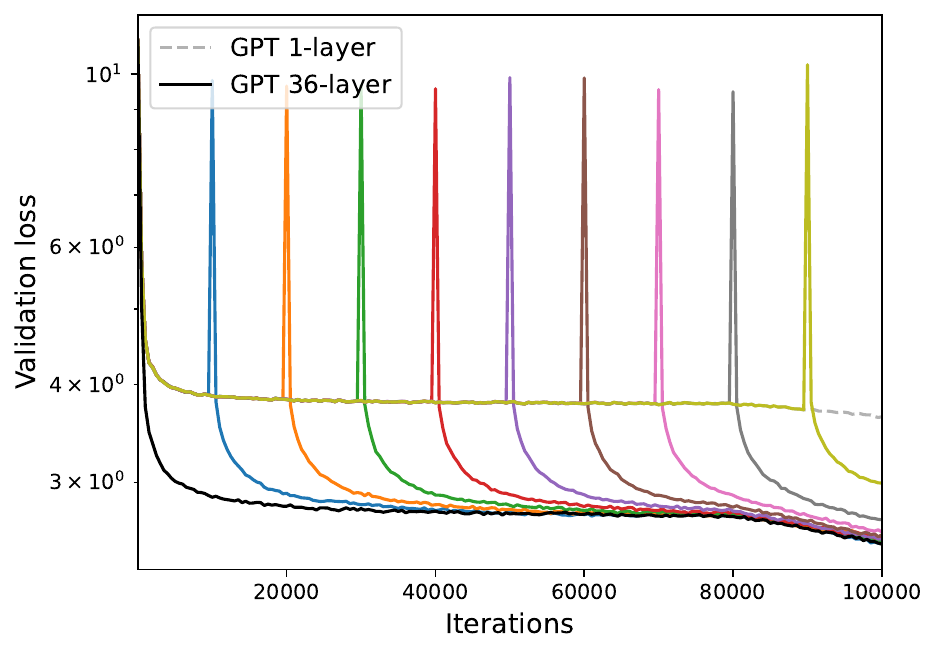}
    
    \includegraphics[width=0.32\linewidth]{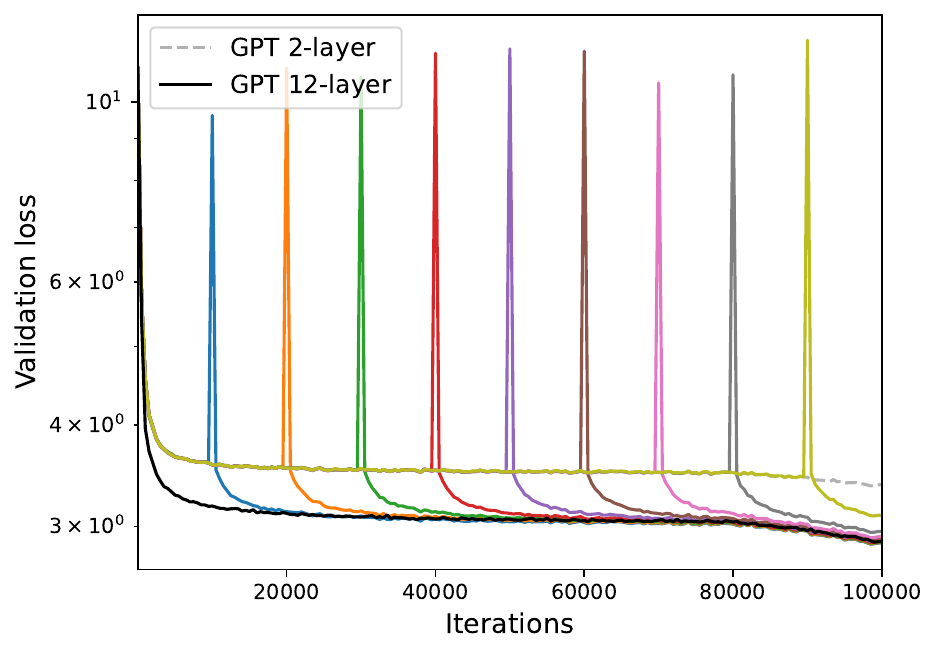}
    \includegraphics[width=0.32\linewidth]{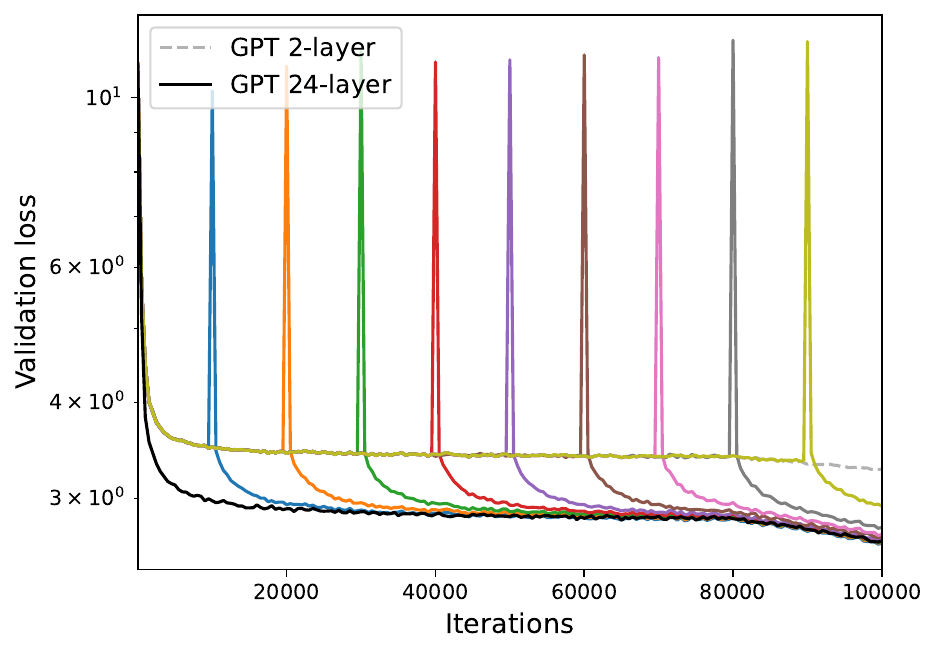}
    \includegraphics[width=0.32\linewidth]{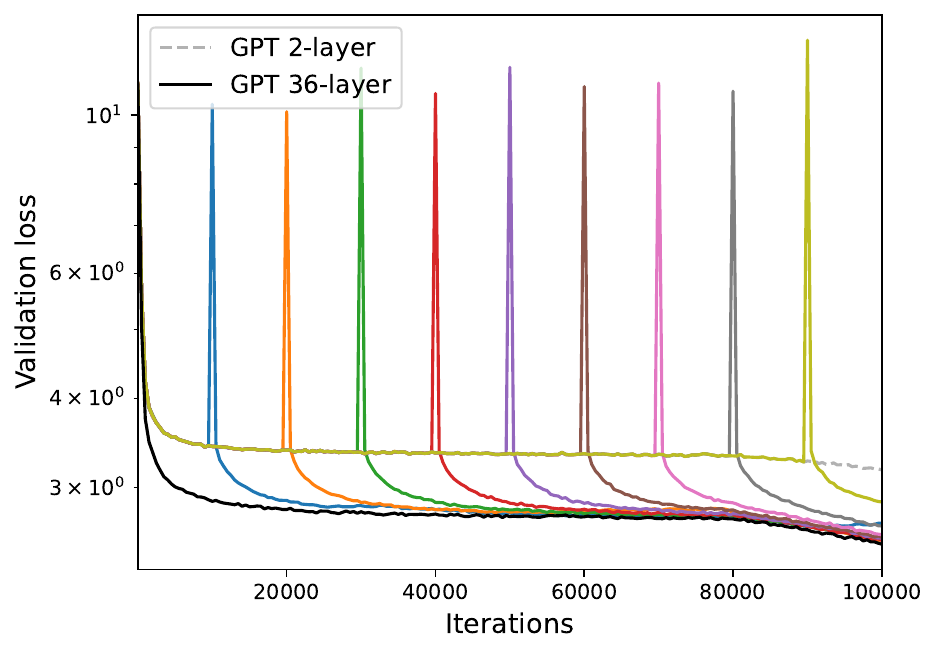}
    
    \includegraphics[width=0.32\linewidth]{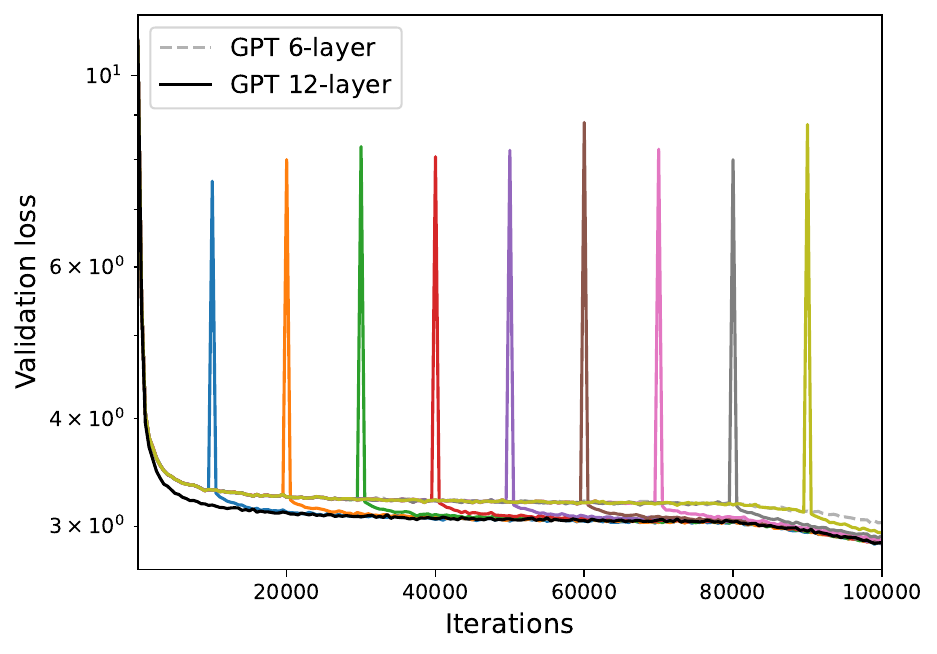}
    \includegraphics[width=0.32\linewidth]{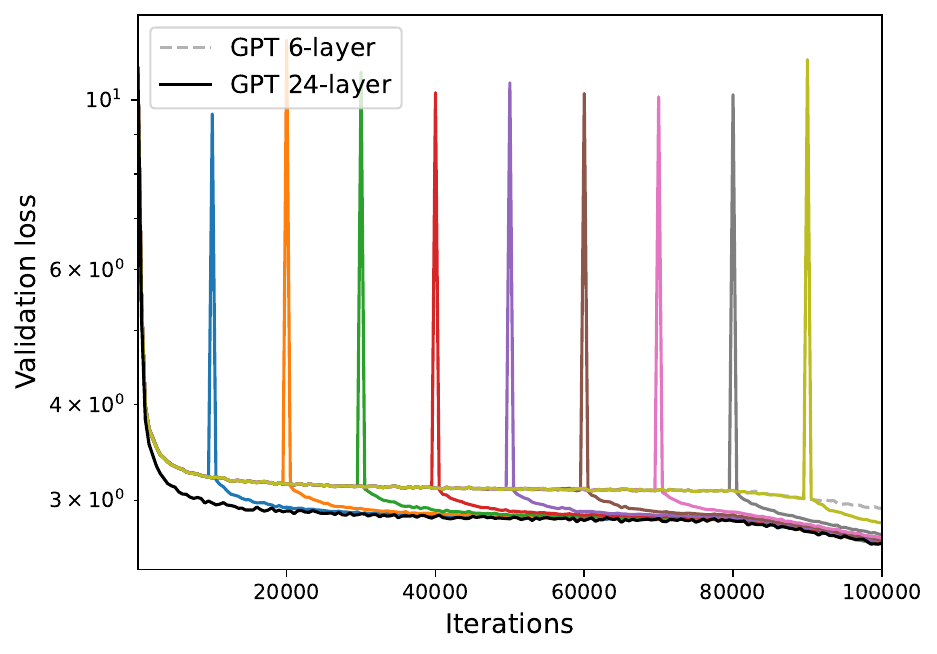}
    \includegraphics[width=0.32\linewidth]{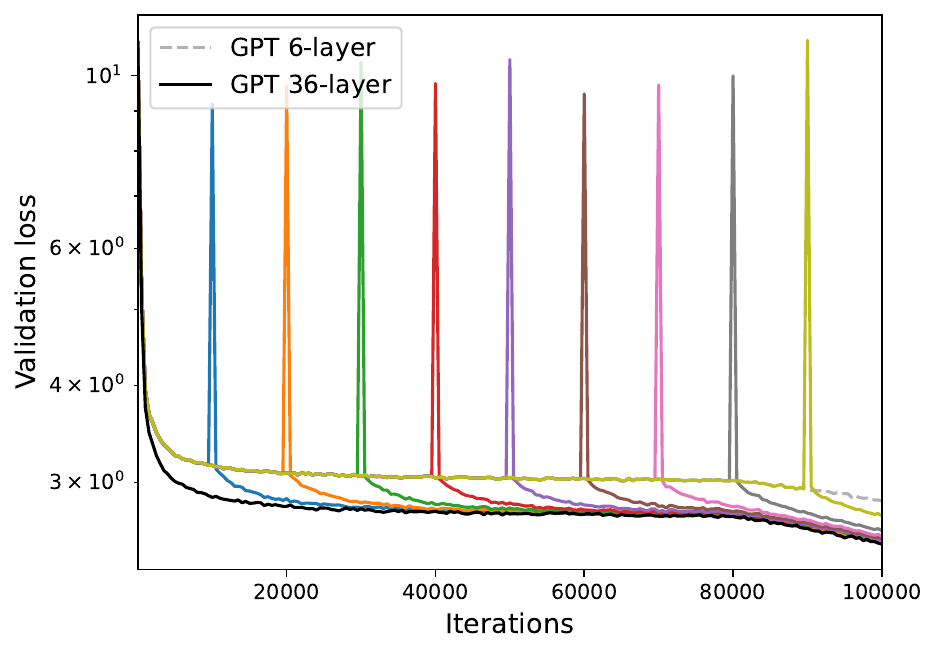}
    
    \includegraphics[width=0.32\linewidth]{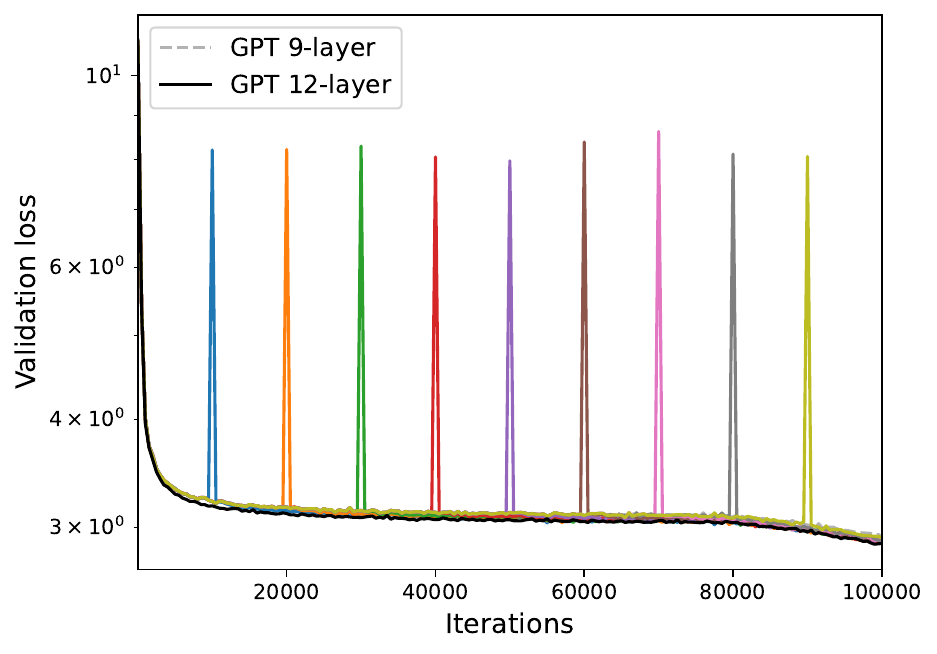}
    \includegraphics[width=0.32\linewidth]{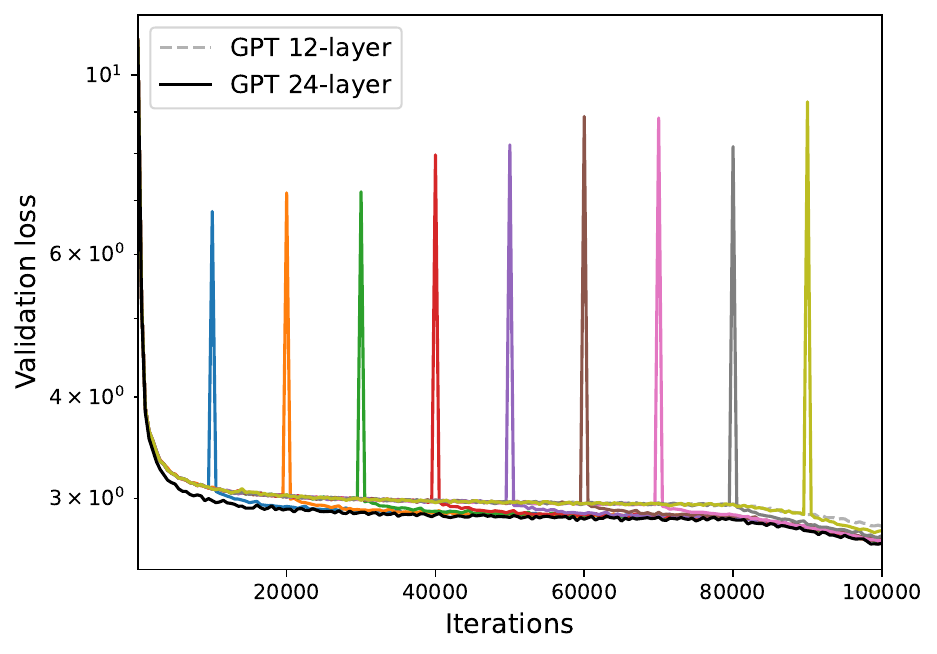}
    \includegraphics[width=0.32\linewidth]{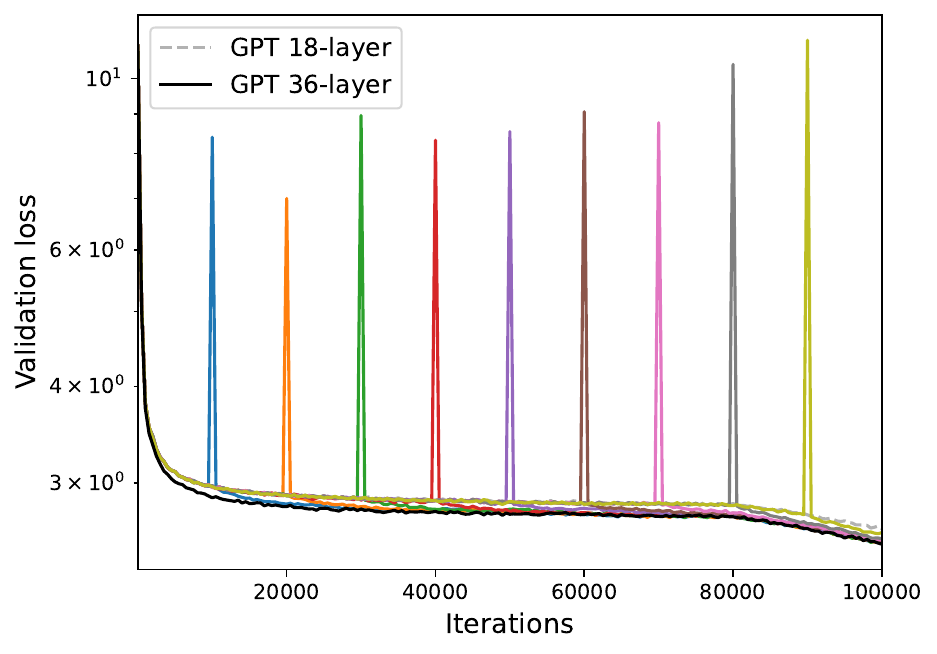}
\caption{Scaling up by depth expansion from $\{0,1,2,6,18\}$ layers to \{12,24,36\} layers GPT2  with $\{124M, 400M,1B\}$ parameters.}
\label{fig:Xto12/24}
\end{figure}

In \Cref{fig:Xto12/24}, we consistently observe the mixing behaviors on hundreds of runs, from various small models to various large models. Specifically, the mixing time is empirically robust to model sizes.

\clearpage
\begin{figure}[!htb]
    \centering
    \includegraphics[width=0.32\linewidth]{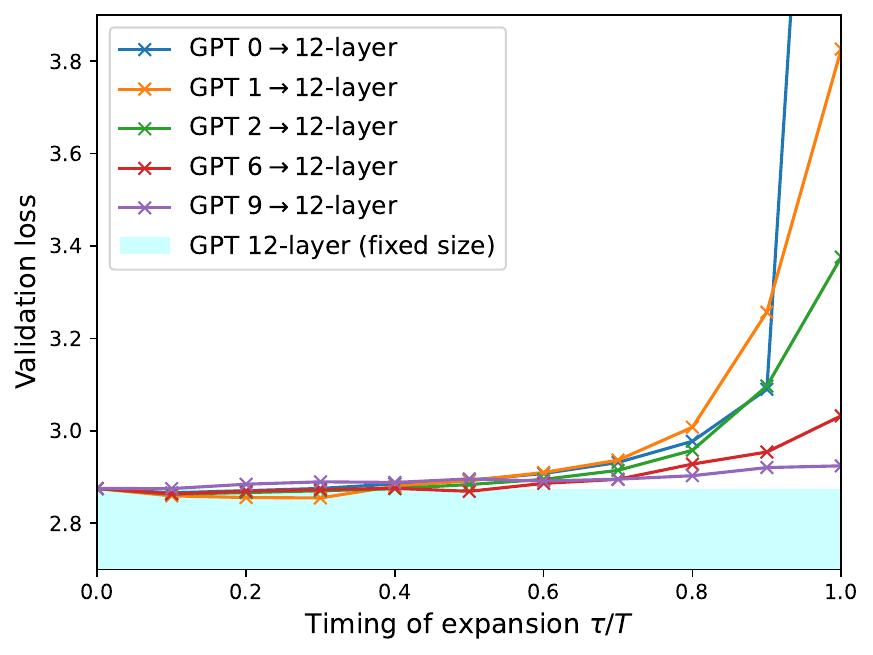}
    \includegraphics[width=0.32\linewidth]{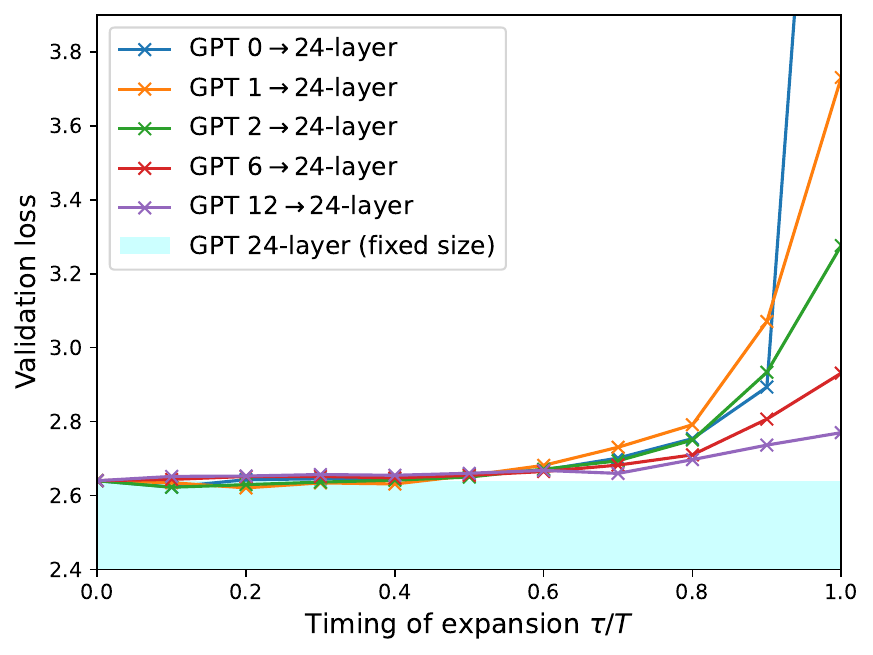}
    \includegraphics[width=0.32\linewidth]{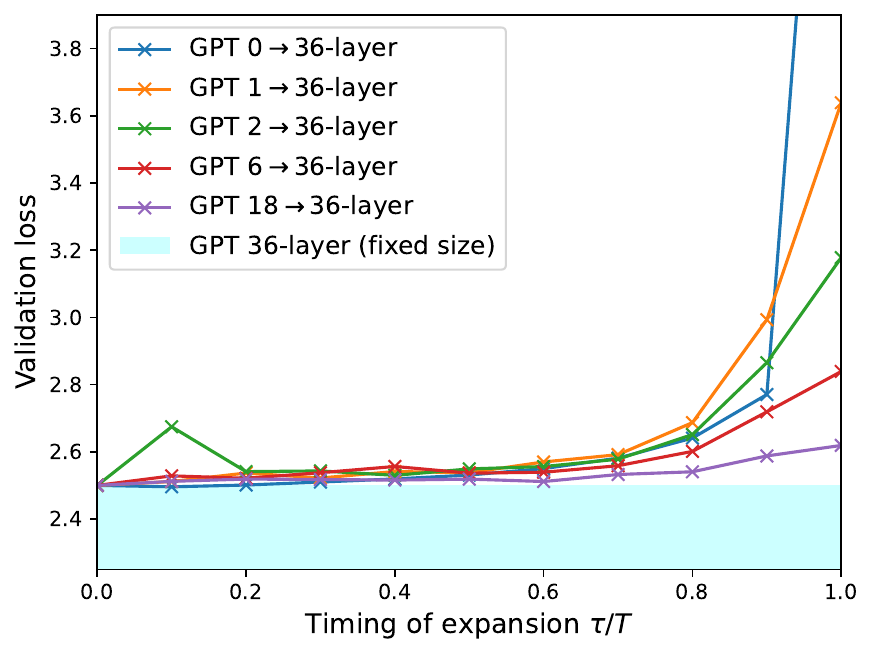}    \caption{Final loss of depth expansion at different timing, from $\{0,1,2,6,18\}$ layers to \{12,24,36\} layers GPT2  with $\{124M, 400M,1B\}$ parameters.}
\end{figure}



\subsection{Optimizer states}
We conduct an ablation study to explore how to deal with the optimizer states (e.g. momentum and variance in AdamW or Muon-NSGD) during the expansion. Previous works have mixed results: \cite{shen2022staged,fu2023triple} show that copying the old layers' optimizer states to new layers can be helpful; \cite{gong2019efficient} resets the optimizer states of all layers.

We consider the following methods for optimizer states (OS): denoting embedding as E, hidden layers as H, and last layer as L, 
\begin{itemize}
    \item (inheriting OS) inheriting existing OS: $[E,H,L]\to [E,0\times 12,L]$
    \item (copying OS) inheriting existing OS and copying hidden layers' OS: $[E,H,L]\to [E,H\times 12,L]$
    \item (no OS) not inheriting any OS: $[E,L] \text{ or } [E,H,L]\to [0,0\times 12,0]$
\end{itemize}
We observe that all methods seem to work in terms of final losses and mixing behaviors, although copying OS is less stable. 
\begin{figure}[!htb]
    \centering
    \includegraphics[width=0.4925\linewidth]{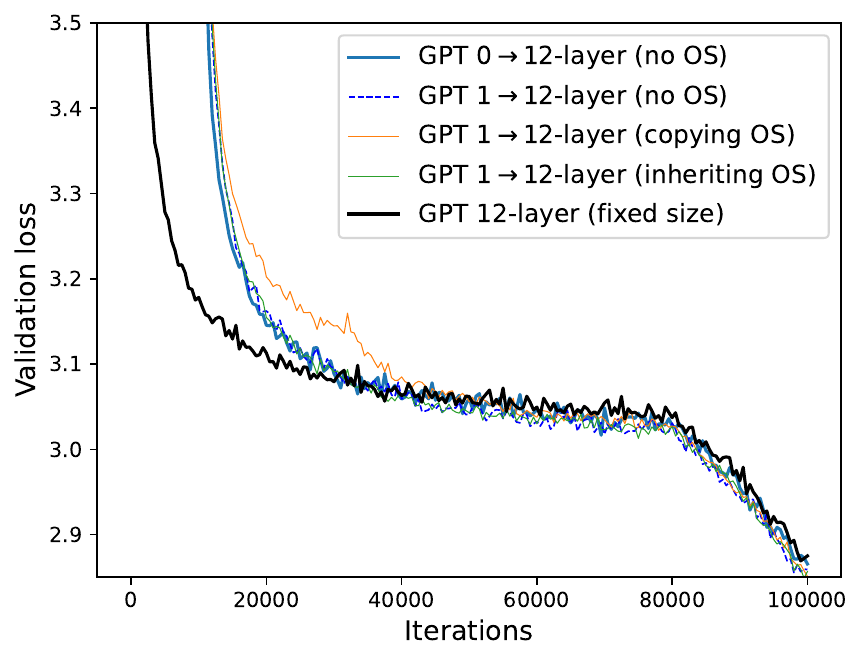}
    \vspace{-0.3cm}
\caption{Validation loss of depth expansion at $\tau=0.1T$ with different ways to set optimizer states.}
\end{figure}

\newpage
\subsection{Choosing optimizer and learning rate schedule}
In \Cref{fig:optim-lr}, we train 100k iterations with two optimizers and two learning rate schedules. The same schedule is used before and after expansion, without changing the learning rate. We observe that Muon-NSGD with WSD schedule achieves best loss at all FLOPs (also at any timing of expansion $\tau/T$). This is consistent with our theory.
\begin{figure}[H]
    \centering
    \includegraphics[width=0.4925\linewidth]{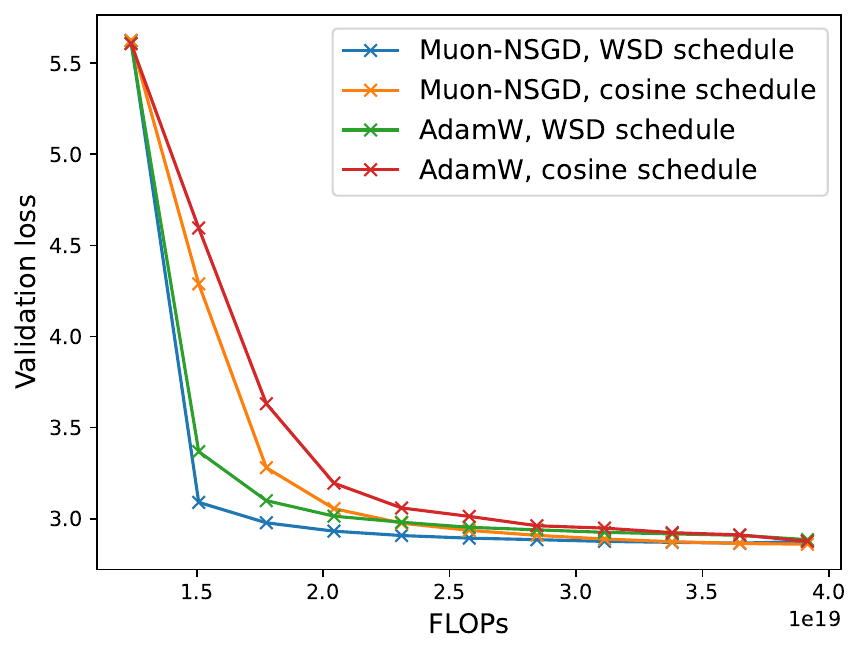}
    \vspace{-0.3cm}
    \caption{Loss-compute tradeoff (validation loss v.s. FLOPs) of zero-layer depth expansion under different optimizers and learning rate schedules. The target model is 12-layer GPT2. For WSD schedule, AdamW uses 0.0005 learning rate and Muon-NSGD uses 0.01 learning rate. For cosine schedule, the learning rates are doubled.}
    \label{fig:optim-lr}
\end{figure}

We can actually switch optimizers after depth expansion, where we switch from NSGD and AdamW to Muon-NSGD and observe mixing behaviors. We believe the gap can be further reduced with additional training. This may be particularly useful for improving the efficiency if the first optimizer is cheaper than the final optimizer.
\begin{figure}[H]
    \centering
    \includegraphics[width=\linewidth]{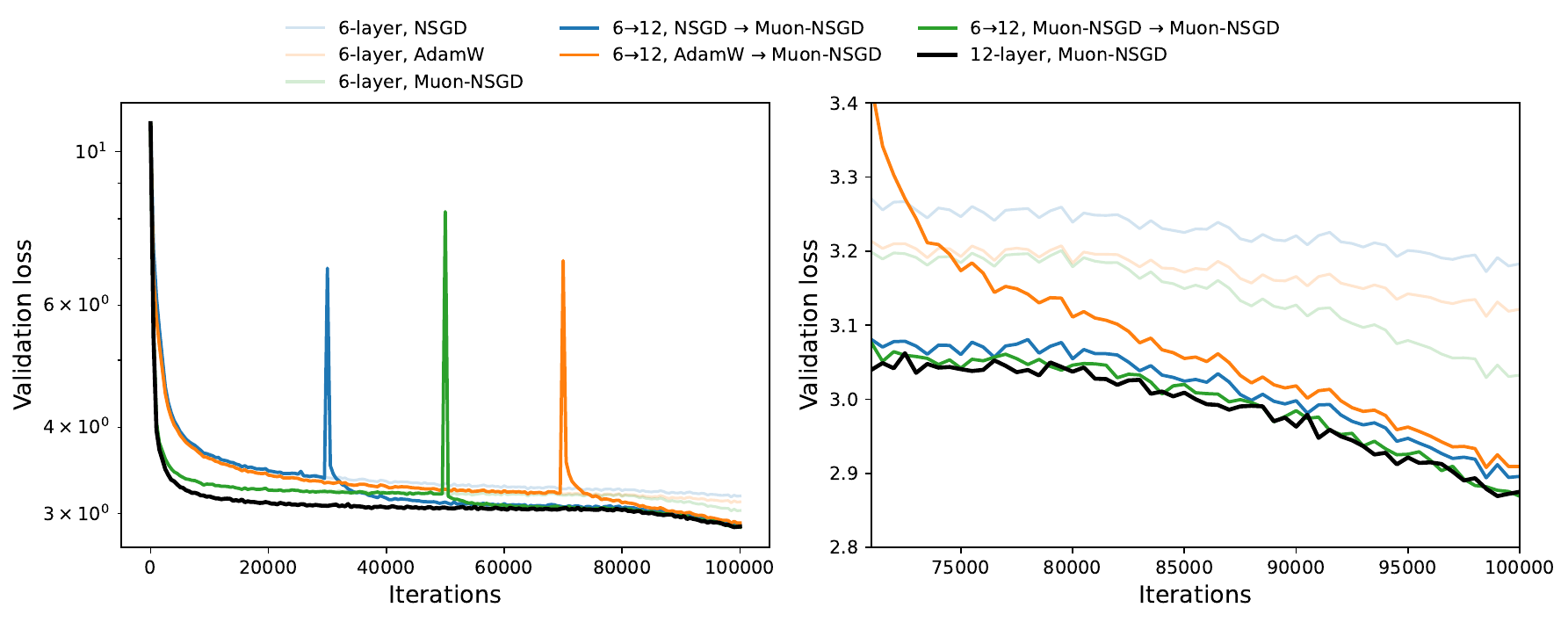}
    \vspace{-0.3cm}
    \caption{Progressive training can use different optimizers before and after the expansion. The depth expansion happens at 30\%, 50\% and 70\% of training on GPT2.}
    \label{fig:optim-lr}
\end{figure}

\newpage
\subsection{Mixing needs data, not iterations}
\begin{figure}[!htb]
\includegraphics[width=0.45\linewidth]{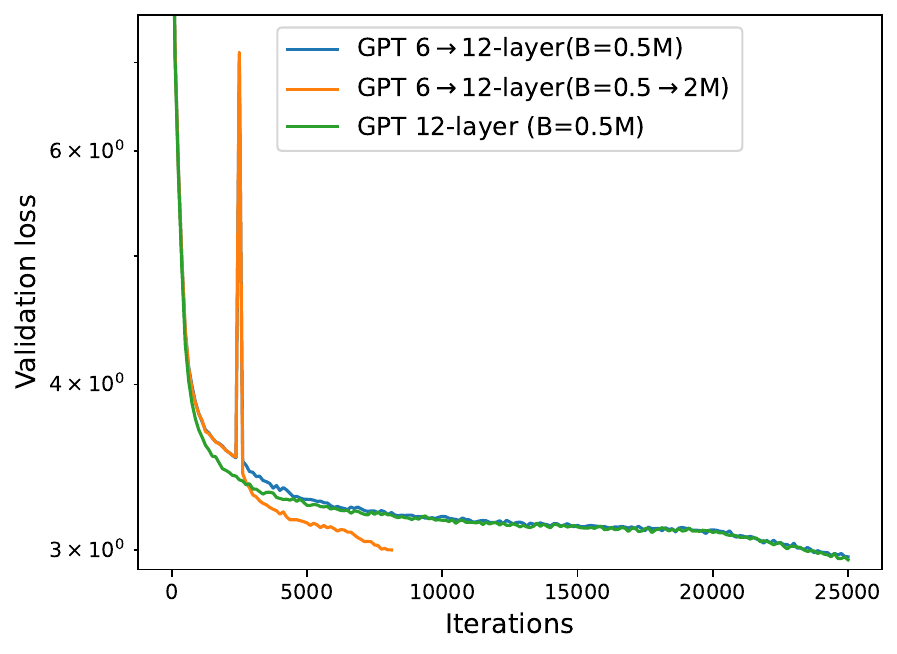}
\includegraphics[width=0.438\linewidth]{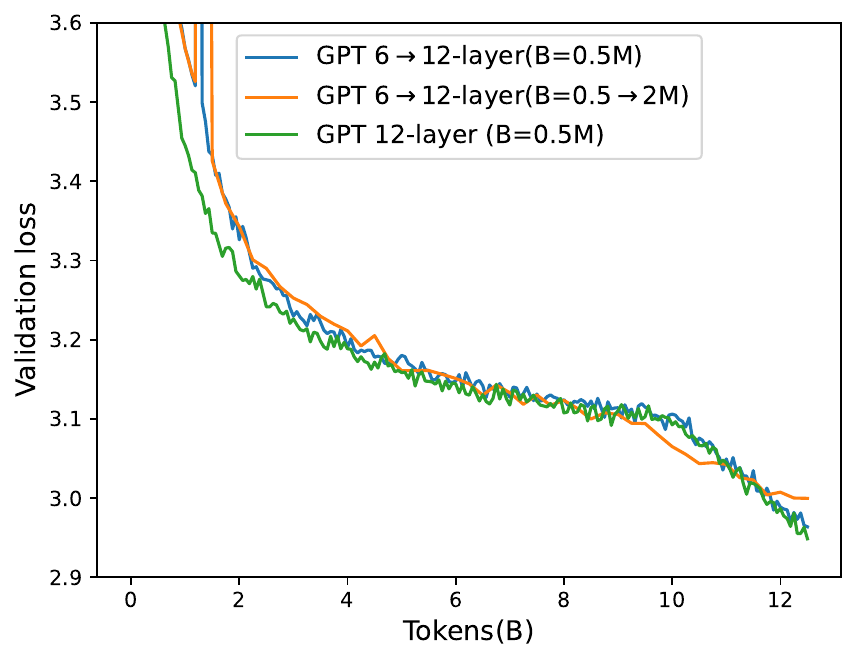}
\vspace{-0.3cm}
\caption{Progressive training needs sufficient data to mix with fixed-size large model training, largely unaffected by batch size or iterations.}
\label{fig:token matters}
\end{figure}

\vspace{-0.1cm}
Importantly, we observe that the mixing time is measured by data size, i.e. images or tokens processed, not by iterations. In \Cref{fig:token matters}, we compare a progressive training with constant batch size to another one with $4\times$ batch size after the depth expansion $(\tau=0.1T)$. The final loss is similar although large-batch training takes much fewer iterations.

\subsection{One-layer model expansion figures}
We present the one-layer model expansion results in correspondence to \Cref{fig:lr schedule} and \Cref{fig:gpt_WSD_10runs_onlygrown} here, due to space limit.

\begin{figure}[!htb]
    \vspace{-0.3cm}
    \centering
    \includegraphics[width=0.33\linewidth]{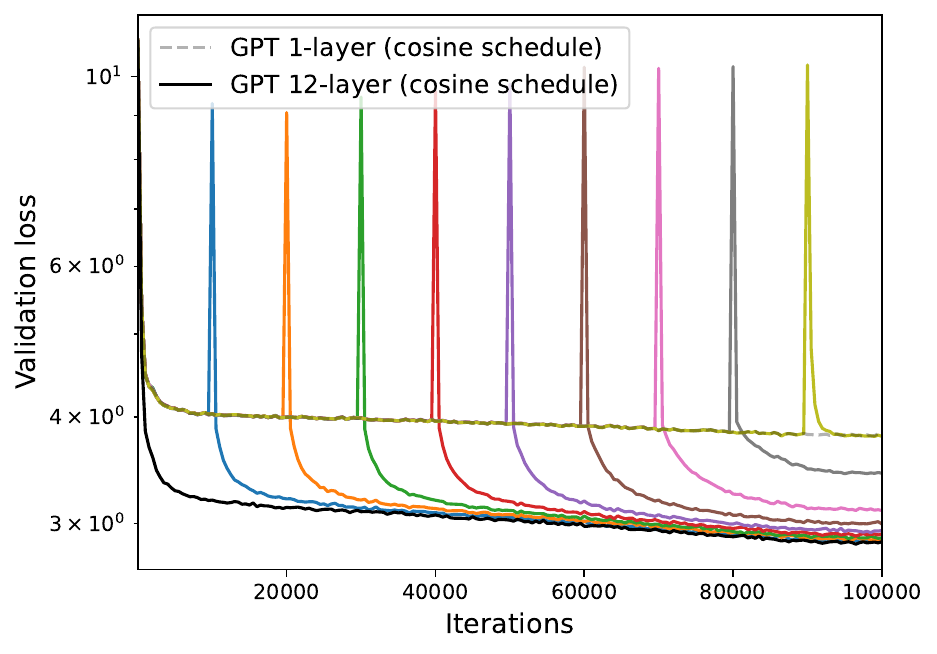}
    \includegraphics[width=0.33\linewidth]{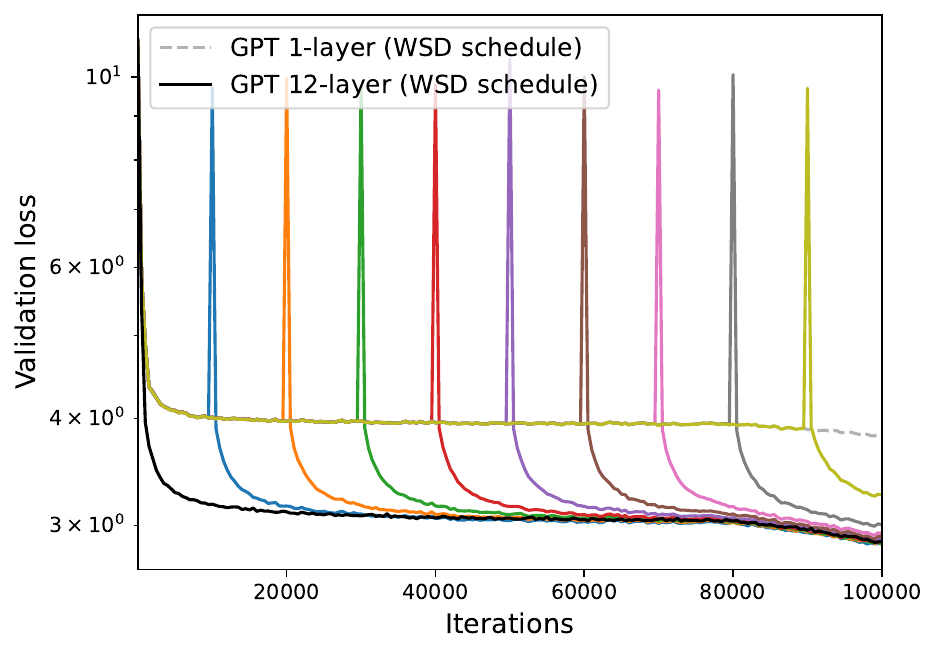}
\includegraphics[width=0.3\linewidth]{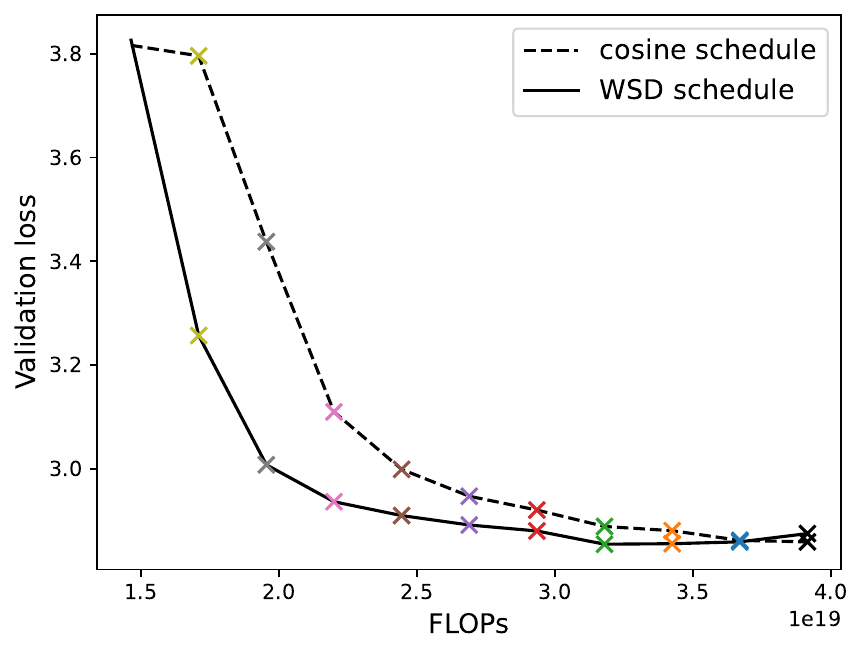}
\vspace{-0.3cm}
    \caption{Performance of one-layer progressive training and fixed-size training, where WSD schedule significantly enhances the progressive training.}
\end{figure}

\begin{figure}[!htb]
\centering
\includegraphics[width=0.43\linewidth]{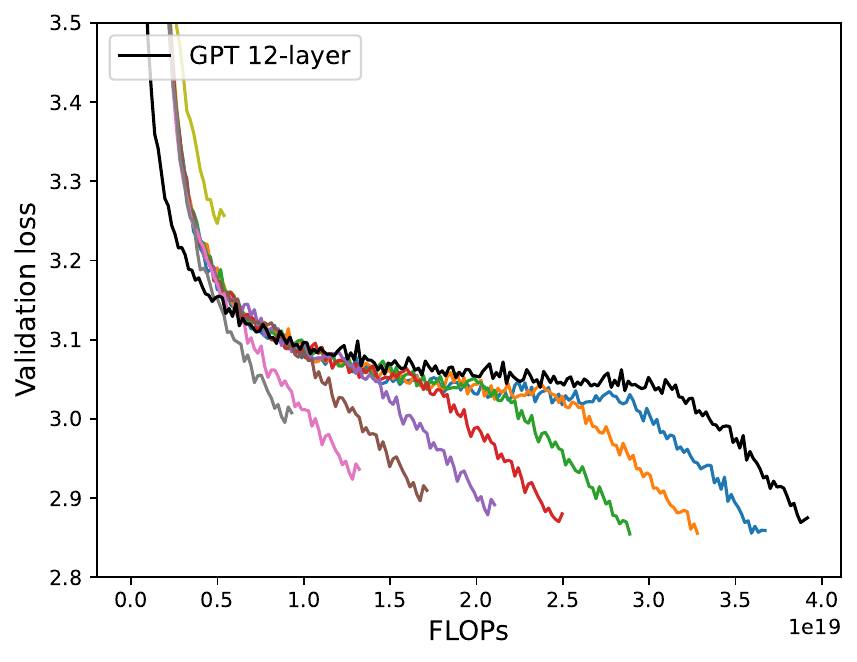}
\includegraphics[width=0.46\linewidth]{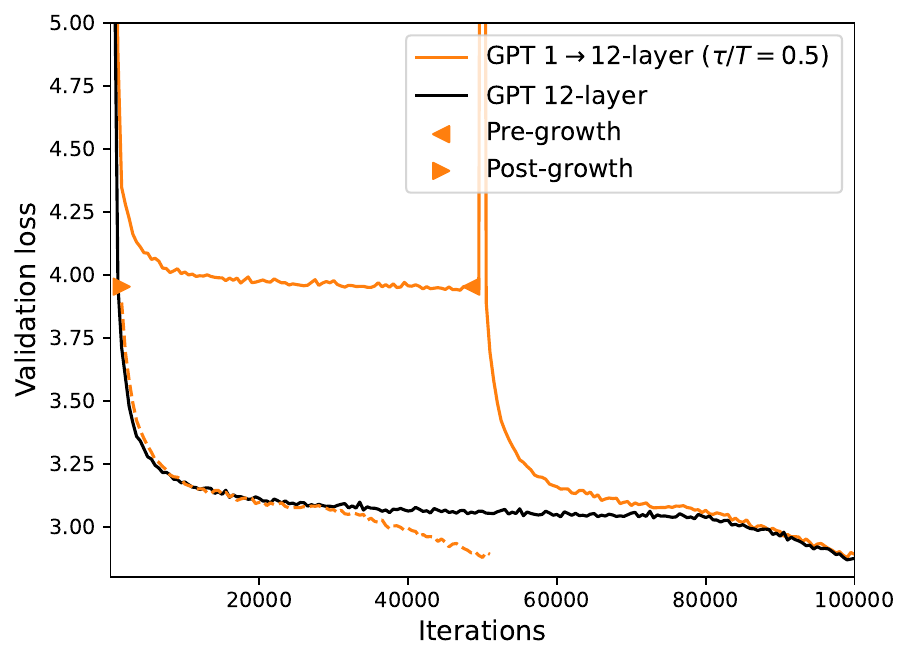}
\vspace{-0.3cm}
\captionof{figure}{Different perspectives to compare (one-layer) progressive training and fixed-size training. Left: only comparing the grown model and target model. Right: matching the pre-growth loss of source model to target model.}
\end{figure}

%% file: references.bib
@article{defazio2023optimal,
  title={Optimal linear decay learning rate schedules and further refinements},
  author={Defazio, Aaron and Cutkosky, Ashok and Mehta, Harsh and Mishchenko, Konstantin},
  journal={arXiv preprint arXiv:2310.07831},
  year={2023}
}

@article{chen2015net2net,
  title={Net2net: Accelerating learning via knowledge transfer},
  author={Chen, Tianqi and Goodfellow, Ian and Shlens, Jonathon},
  journal={arXiv preprint arXiv:1511.05641},
  year={2015}
}

@inproceedings{wanglemon,
  title={LEMON: Lossless model expansion},
  author={Wang, Yite and Su, Jiahao and Lu, Hanlin and Xie, Cong and Liu, Tianyi and Yuan, Jianbo and Lin, Haibin and Sun, Ruoyu and Yang, Hongxia},
  booktitle={The Twelfth International Conference on Learning Representations},
year={2023}
}

@article{jiang2024mixtral,
  title={Mixtral of experts},
  author={Jiang, Albert Q and Sablayrolles, Alexandre and Roux, Antoine and Mensch, Arthur and Savary, Blanche and Bamford, Chris and Chaplot, Devendra Singh and Casas, Diego de las and Hanna, Emma Bou and Bressand, Florian and others},
  journal={arXiv preprint arXiv:2401.04088},
  year={2024}
}

@article{tan2024dlo,
  title={Dlo: Dynamic layer operation for efficient vertical scaling of llms},
  author={Tan, Zhen and Dong, Daize and Zhao, Xinyu and Peng, Jie and Cheng, Yu and Chen, Tianlong},
  journal={arXiv preprint arXiv:2407.11030},
  year={2024}
}

@article{schaipp2025surprising,
  title={The Surprising Agreement Between Convex Optimization Theory and Learning-Rate Scheduling for Large Model Training},
  author={Schaipp, Fabian and H{\"a}gele, Alexander and Taylor, Adrien and Simsekli, Umut and Bach, Francis},
  journal={arXiv preprint arXiv:2501.18965},
  year={2025}
}

@inproceedings{defazio2023learning,
  title={Learning-rate-free learning by d-adaptation},
  author={Defazio, Aaron and Mishchenko, Konstantin},
  booktitle={International Conference on Machine Learning},
  pages={7449--7479},
  year={2023},
  organization={PMLR}
}

@article{kaplan2020scaling,
  title={Scaling laws for neural language models},
  author={Kaplan, Jared and McCandlish, Sam and Henighan, Tom and Brown, Tom B and Chess, Benjamin and Child, Rewon and Gray, Scott and Radford, Alec and Wu, Jeffrey and Amodei, Dario},
  journal={arXiv preprint arXiv:2001.08361},
  year={2020}
}

@article{dubey2024llama,
  title={The llama 3 herd of models},
  author={Dubey, Abhimanyu and Jauhri, Abhinav and Pandey, Abhinav and Kadian, Abhishek and Al-Dahle, Ahmad and Letman, Aiesha and Mathur, Akhil and Schelten, Alan and Yang, Amy and Fan, Angela and others},
  journal={arXiv e-prints},
  pages={arXiv--2407},
  year={2024}
}

@article{yang2025qwen3,
  title={Qwen3 technical report},
  author={Yang, An and Li, Anfeng and Yang, Baosong and Zhang, Beichen and Hui, Binyuan and Zheng, Bo and Yu, Bowen and Gao, Chang and Huang, Chengen and Lv, Chenxu and others},
  journal={arXiv preprint arXiv:2505.09388},
  year={2025}
}

@article{shazeer2020glu,
  title={Glu variants improve transformer},
  author={Shazeer, Noam},
  journal={arXiv preprint arXiv:2002.05202},
  year={2020}
}

@inproceedings{
anonymous2026convex,
title={Convex Dominance in Deep Learning: A Scaling Law of Loss and Learning Rate},
author={Bu, Zhiqi and Xu, Shiyun and Mao, Jialin},
booktitle={The Fourteenth International Conference on Learning Representations},
year={2026},
url={https://openreview.net/forum?id=dSdLqg02tx}
}

@article{zhang2019root,
  title={Root mean square layer normalization},
  author={Zhang, Biao and Sennrich, Rico},
  journal={Advances in neural information processing systems},
  volume={32},
  year={2019}
}

@article{hendrycks2016bridging,
  title={Bridging Nonlinearities and Stochastic Regularizers with Gaussian Error Linear Units},
  author={Hendrycks, Dan and Gimpel, Kevin},
  journal={arXiv preprint arXiv:1606.08415},
  year={2016}
}

@article{su2024roformer,
  title={Roformer: Enhanced transformer with rotary position embedding},
  author={Su, Jianlin and Ahmed, Murtadha and Lu, Yu and Pan, Shengfeng and Bo, Wen and Liu, Yunfeng},
  journal={Neurocomputing},
  volume={568},
  pages={127063},
  year={2024},
  publisher={Elsevier}
}

@article{liu2024deepseek2,
  title={Deepseek-v2: A strong, economical, and efficient mixture-of-experts language model},
  author={Liu, Aixin and Feng, Bei and Wang, Bin and Wang, Bingxuan and Liu, Bo and Zhao, Chenggang and Dengr, Chengqi and Ruan, Chong and Dai, Damai and Guo, Daya and others},
  journal={arXiv preprint arXiv:2405.04434},
  year={2024}
}

@inproceedings{ainslie2023gqa,
  title={GQA: Training Generalized Multi-Query Transformer Models from Multi-Head Checkpoints},
  author={Ainslie, Joshua and Lee-Thorp, James and de Jong, Michiel and Zemlyanskiy, Yury and Lebron, Federico and Sanghai, Sumit},
  booktitle={Proceedings of the 2023 Conference on Empirical Methods in Natural Language Processing},
  pages={4895--4901},
  year={2023}
}

@article{liu2024deepseek,
  title={Deepseek-v3 technical report},
  author={Liu, Aixin and Feng, Bei and Xue, Bing and Wang, Bingxuan and Wu, Bochao and Lu, Chengda and Zhao, Chenggang and Deng, Chengqi and Zhang, Chenyu and Ruan, Chong and others},
  journal={arXiv preprint arXiv:2412.19437},
  year={2024}
}

@inproceedings{he2016deep,
  title={Deep residual learning for image recognition},
  author={He, Kaiming and Zhang, Xiangyu and Ren, Shaoqing and Sun, Jian},
  booktitle={Proceedings of the IEEE conference on computer vision and pattern recognition},
  pages={770--778},
  year={2016}
}

@article{radford2019language,
  title={Language models are unsupervised multitask learners},
  author={Radford, Alec and Wu, Jeffrey and Child, Rewon and Luan, David and Amodei, Dario and Sutskever, Ilya and others},
  journal={OpenAI blog},
  volume={1},
  number={8},
  pages={9},
  year={2019}
}

@inproceedings{hoffmann2022training,
  title={Training compute-optimal large language models},
  author={Hoffmann, Jordan and Borgeaud, Sebastian and Mensch, Arthur and Buchatskaya, Elena and Cai, Trevor and Rutherford, Eliza and de Las Casas, Diego and Hendricks, Lisa Anne and Welbl, Johannes and Clark, Aidan and others},
  booktitle={Proceedings of the 36th International Conference on Neural Information Processing Systems},
  pages={30016--30030},
  year={2022}
}

@inproceedings{bugradient,
  title={Gradient descent with generalized Newton’s method},
  author={Bu, Zhiqi and Xu, Shiyun},
  booktitle={The Thirteenth International Conference on Learning Representations},
  year={2024},
}

@article{leclerc2020two,
  title={The two regimes of deep network training},
  author={Leclerc, Guillaume and Madry, Aleksander},
  journal={arXiv preprint arXiv:2002.10376},
  year={2020}
}

@inproceedings{allen2019convergence,
  title={A convergence theory for deep learning via over-parameterization},
  author={Allen-Zhu, Zeyuan and Li, Yuanzhi and Song, Zhao},
  booktitle={International conference on machine learning},
  pages={242--252},
  year={2019},
  organization={PMLR}
}

@inproceedings{bu2021dynamical,
  title={A dynamical view on optimization algorithms of overparameterized neural networks},
  author={Bu, Zhiqi and Xu, Shiyun and Chen, Kan},
  booktitle={International conference on artificial intelligence and statistics},
  pages={3187--3195},
  year={2021},
  organization={PMLR}
}

@article{jacot2018neural,
  title={Neural tangent kernel: Convergence and generalization in neural networks},
  author={Jacot, Arthur and Gabriel, Franck and Hongler, Cl{\'e}ment},
  journal={Advances in neural information processing systems},
  volume={31},
  year={2018}
}

@article{lee2019wide,
  title={Wide neural networks of any depth evolve as linear models under gradient descent},
  author={Lee, Jaehoon and Xiao, Lechao and Schoenholz, Samuel and Bahri, Yasaman and Novak, Roman and Sohl-Dickstein, Jascha and Pennington, Jeffrey},
  journal={Advances in neural information processing systems},
  volume={32},
  year={2019}
}

@inproceedings{li2020shallow,
  title={Shallow-to-Deep Training for Neural Machine Translation},
  author={Li, Bei and Wang, Ziyang and Liu, Hui and Jiang, Yufan and Du, Quan and Xiao, Tong and Wang, Huizhen and Zhu, Jingbo},
  booktitle={Proceedings of the 2020 Conference on Empirical Methods in Natural Language Processing (EMNLP)},
  pages={995--1005},
  year={2020}
}

@article{vaswani2017attention,
  title={Attention is all you need},
  author={Vaswani, Ashish and Shazeer, Noam and Parmar, Niki and Uszkoreit, Jakob and Jones, Llion and Gomez, Aidan N and Kaiser, {\L}ukasz and Polosukhin, Illia},
  journal={Advances in neural information processing systems},
  volume={30},
  year={2017}
}

@article{qin2021knowledge,
  title={Knowledge inheritance for pre-trained language models},
  author={Qin, Yujia and Lin, Yankai and Yi, Jing and Zhang, Jiajie and Han, Xu and Zhang, Zhengyan and Su, Yusheng and Liu, Zhiyuan and Li, Peng and Sun, Maosong and others},
  journal={arXiv preprint arXiv:2105.13880},
  year={2021}
}

@article{wang2023learning,
  title={Learning to grow pretrained models for efficient transformer training},
  author={Wang, Peihao and Panda, Rameswar and Hennigen, Lucas Torroba and Greengard, Philip and Karlinsky, Leonid and Feris, Rogerio and Cox, David Daniel and Wang, Zhangyang and Kim, Yoon},
  journal={arXiv preprint arXiv:2303.00980},
  year={2023}
}

@misc{Gokaslan2019OpenWeb,
    title={OpenWebText Corpus},
    author={Gokaslan, Aaron and Cohen, Vanya and Pavlick, Ellie and Tellex, Stefanie},
    howpublished={\url{http://Skylion007.github.io/OpenWebTextCorpus}},
    year={2019}
}

@inproceedings{fu2023triple,
  title={TripLe: revisiting pretrained model reuse and progressive learning for efficient vision transformer scaling and searching},
  author={Fu, Cheng and Huang, Hanxian and Jiang, Zixuan and Ni, Yun and Nai, Lifeng and Wu, Gang and Cheng, Liqun and Zhou, Yanqi and Li, Sheng and Li, Andrew and others},
  booktitle={Proceedings of the IEEE/CVF International Conference on Computer Vision},
  pages={17153--17163},
  year={2023}
}

@inproceedings{qin2022elle,
  title={ELLE: Efficient Lifelong Pre-training for Emerging Data},
  author={Qin, Yujia and Zhang, Jiajie and Lin, Yankai and Liu, Zhiyuan and Li, Peng and Sun, Maosong and Zhou, Jie},
  booktitle={Findings of the Association for Computational Linguistics: ACL 2022},
  pages={2789--2810},
  year={2022}
}

@article{xing2018walk,
  title={A walk with sgd},
  author={Xing, Chen and Arpit, Devansh and Tsirigotis, Christos and Bengio, Yoshua},
  journal={arXiv preprint arXiv:1802.08770},
  year={2018}
}

@article{hagele2024scaling,
  title={Scaling laws and compute-optimal training beyond fixed training durations},
  author={H{\"a}gele, Alex and Bakouch, Elie and Kosson, Atli and Von Werra, Leandro and Jaggi, Martin and others},
  journal={Advances in Neural Information Processing Systems},
  volume={37},
  pages={76232--76264},
  year={2024}
}

@inproceedings{reddi2023efficient,
  title={Efficient training of language models using few-shot learning},
  author={Reddi, Sashank J and Miryoosefi, Sobhan and Karp, Stefani and Krishnan, Shankar and Kale, Satyen and Kim, Seungyeon and Kumar, Sanjiv},
  booktitle={International Conference on Machine Learning},
  pages={14553--14568},
  year={2023},
  organization={PMLR}
}

@inproceedings{yaomasked,
  title={Masked Structural Growth for 2x Faster Language Model Pre-training},
  author={Yao, Yiqun and Zhang, Zheng and Li, Jing and Wang, Yequan},
  booktitle={The Twelfth International Conference on Learning Representations}
}

@article{agarwal2024stacking,
  title={Stacking as accelerated gradient descent},
  author={Agarwal, Naman and Awasthi, Pranjal and Kale, Satyen and Zhao, Eric},
  journal={arXiv preprint arXiv:2403.04978},
  year={2024}
}

@article{chen2021bert2bert,
  title={bert2bert: Towards reusable pretrained language models},
  author={Chen, Cheng and Yin, Yichun and Shang, Lifeng and Jiang, Xin and Qin, Yujia and Wang, Fengyu and Wang, Zhi and Chen, Xiao and Liu, Zhiyuan and Liu, Qun},
  journal={arXiv preprint arXiv:2110.07143},
  year={2021}
}

@inproceedings{gong2019efficient,
  title={Efficient training of bert by progressively stacking},
  author={Gong, Linyuan and He, Di and Li, Zhuohan and Qin, Tao and Wang, Liwei and Liu, Tieyan},
  booktitle={International conference on machine learning},
  pages={2337--2346},
  year={2019},
  organization={PMLR}
}

@inproceedings{dong2020towards,
  title={Towards adaptive residual network training: A neural-ode perspective},
  author={Dong, Chengyu and Liu, Liyuan and Li, Zichao and Shang, Jingbo},
  booktitle={International conference on machine learning},
  pages={2616--2626},
  year={2020},
  organization={PMLR}
}

@inproceedings{mei2019mean,
  title={Mean-field theory of two-layers neural networks: dimension-free bounds and kernel limit},
  author={Mei, Song and Misiakiewicz, Theodor and Montanari, Andrea},
  booktitle={Conference on learning theory},
  pages={2388--2464},
  year={2019},
  organization={PMLR}
}

@article{chizat2018global,
  title={On the global convergence of gradient descent for over-parameterized models using optimal transport},
  author={Chizat, Lenaic and Bach, Francis},
  journal={Advances in neural information processing systems},
  volume={31},
  year={2018}
}

@article{yang2022tensor,
  title={Tensor programs v: Tuning large neural networks via zero-shot hyperparameter transfer},
  author={Yang, Greg and Hu, Edward J and Babuschkin, Igor and Sidor, Szymon and Liu, Xiaodong and Farhi, David and Ryder, Nick and Pachocki, Jakub and Chen, Weizhu and Gao, Jianfeng},
  journal={arXiv preprint arXiv:2203.03466},
  year={2022}
}

@article{yang2020feature,
  title={Feature learning in infinite-width neural networks},
  author={Yang, Greg and Hu, Edward J},
  journal={arXiv preprint arXiv:2011.14522},
  year={2020}
}

@article{yang2023spectral,
  title={A spectral condition for feature learning},
  author={Yang, Greg and Simon, James B and Bernstein, Jeremy},
  journal={arXiv preprint arXiv:2310.17813},
  year={2023}
}

@inproceedings{wang2017growing,
  title={Growing a brain: Fine-tuning by increasing model capacity},
  author={Wang, Yu-Xiong and Ramanan, Deva and Hebert, Martial},
  booktitle={Proceedings of the IEEE conference on computer vision and pattern recognition},
  pages={2471--2480},
  year={2017}
}

@inproceedings{shen2022staged,
  title={Staged training for transformer language models},
  author={Shen, Sheng and Walsh, Pete and Keutzer, Kurt and Dodge, Jesse and Peters, Matthew and Beltagy, Iz},
  booktitle={International Conference on Machine Learning},
  pages={19893--19908},
  year={2022},
  organization={PMLR}
}

@article{yang2020progressively,
  title={Progressively stacking 2.0: A multi-stage layerwise training method for bert training speedup},
  author={Yang, Cheng and Wang, Shengnan and Yang, Chao and Li, Yuechuan and He, Ru and Zhang, Jingqiao},
  journal={arXiv preprint arXiv:2011.13635},
  year={2020}
}

@inproceedings{pan2024preparing,
  title={Preparing lessons for progressive training on language models},
  author={Pan, Yu and Yuan, Ye and Yin, Yichun and Shi, Jiaxin and Xu, Zenglin and Zhang, Ming and Shang, Lifeng and Jiang, Xin and Liu, Qun},
  booktitle={Proceedings of the AAAI Conference on Artificial Intelligence},
  volume={38},
  number={17},
  pages={18860--18868},
  year={2024}
}

@inproceedings{chang2018multi,
  title={Multi-level Residual Networks from Dynamical Systems View},
  author={Chang, Bo and Meng, Lili and Haber, Eldad and Tung, Frederick and Begert, David},
  booktitle={International Conference on Learning Representations},
  year={2018}
}

@article{du2024stacking,
  title={Stacking your transformers: A closer look at model growth for efficient llm pre-training},
  author={Du, Wenyu and Luo, Tongxu and Qiu, Zihan and Huang, Zeyu and Shen, Yikang and Cheng, Reynold and Guo, Yike and Fu, Jie},
  journal={Advances in Neural Information Processing Systems},
  volume={37},
  pages={10491--10540},
  year={2024}
}

@article{he2024upcycling,
  title={Upcycling large language models into mixture of experts},
  author={He, Ethan and Khattar, Abhinav and Prenger, Ryan and Korthikanti, Vijay and Yan, Zijie and Liu, Tong and Fan, Shiqing and Aithal, Ashwath and Shoeybi, Mohammad and Catanzaro, Bryan},
  journal={arXiv preprint arXiv:2410.07524},
  year={2024}
}

@article{wei2024skywork,
  title={Skywork-moe: A deep dive into training techniques for mixture-of-experts language models},
  author={Wei, Tianwen and Zhu, Bo and Zhao, Liang and Cheng, Cheng and Li, Biye and L{\"u}, Weiwei and Cheng, Peng and Zhang, Jianhao and Zhang, Xiaoyu and Zeng, Liang and others},
  journal={arXiv preprint arXiv:2406.06563},
  year={2024}
}

@inproceedings{liewscaling,
  title={Scaling Laws for Upcycling Mixture-of-Experts Language Models},
  author={Liew, Seng Pei and Kato, Takuya and Takase, Sho},
  booktitle={Forty-second International Conference on Machine Learning}
}

@inproceedings{nakamuradrop,
  title={Drop-Upcycling: Training Sparse Mixture of Experts with Partial Re-initialization},
  author={Nakamura, Taishi and Akiba, Takuya and Fujii, Kazuki and Oda, Yusuke and Yokota, Rio and Suzuki, Jun},
  booktitle={The Thirteenth International Conference on Learning Representations}
}

@inproceedings{komatsuzakisparse,
  title={Sparse Upcycling: Training Mixture-of-Experts from Dense Checkpoints},
  author={Komatsuzaki, Aran and Puigcerver, Joan and Lee-Thorp, James and Ruiz, Carlos Riquelme and Mustafa, Basil and Ainslie, Joshua and Tay, Yi and Dehghani, Mostafa and Houlsby, Neil},
  booktitle={The Eleventh International Conference on Learning Representations}
}

@inproceedings{
boreiko2025towards,
title={Towards understanding of orthogonalization in Muon},
author={Valentyn Boreiko and Zhiqi Bu and Sheng Zha},
booktitle={High-dimensional Learning Dynamics 2025},
year={2025},
url={https://openreview.net/forum?id=ppmyFtr9EW}
}

@inproceedings{muennighoffolmoe,
  title={OLMoE: Open Mixture-of-Experts Language Models},
  author={Muennighoff, Niklas and Soldaini, Luca and Groeneveld, Dirk and Lo, Kyle and Morrison, Jacob and Min, Sewon and Shi, Weijia and Walsh, Evan Pete and Tafjord, Oyvind and Lambert, Nathan and others},
  booktitle={The Thirteenth International Conference on Learning Representations}
}

@article{fedus2022switch,
  title={Switch transformers: Scaling to trillion parameter models with simple and efficient sparsity},
  author={Fedus, William and Zoph, Barret and Shazeer, Noam},
  journal={Journal of Machine Learning Research},
  volume={23},
  number={120},
  pages={1--39},
  year={2022}
}

@inproceedings{kim2024solar,
  title={Solar 10.7 b: Scaling large language models with simple yet effective depth up-scaling},
  author={Kim, Sanghoon and Kim, Dahyun and Park, Chanjun and Lee, Wonsung and Song, Wonho and Kim, Yunsu and Kim, Hyeonwoo and Kim, Yungi and Lee, Hyeonju and Kim, Jihoo and others},
  booktitle={Proceedings of the 2024 Conference of the North American Chapter of the Association for Computational Linguistics: Human Language Technologies (Volume 6: Industry Track)},
  pages={23--35},
  year={2024}
}

@article{wu2024llama,
  title={Llama pro: Progressive llama with block expansion},
  author={Wu, Chengyue and Gan, Yukang and Ge, Yixiao and Lu, Zeyu and Wang, Jiahao and Feng, Ye and Shan, Ying and Luo, Ping},
  journal={arXiv preprint arXiv:2401.02415},
  year={2024}
}

@article{yano2025efficient,
  title={Efficient Construction of Model Family through Progressive Training Using Model Expansion},
  author={Yano, Kazuki and Takase, Sho and Kobayashi, Sosuke and Kiyono, Shun and Suzuki, Jun},
  journal={arXiv preprint arXiv:2504.00623},
  year={2025}
}

@article{gu2020transformer,
  title={On the transformer growth for progressive bert training},
  author={Gu, Xiaotao and Liu, Liyuan and Yu, Hongkun and Li, Jing and Chen, Chen and Han, Jiawei},
  journal={arXiv preprint arXiv:2010.12562},
  year={2020}
}

@inproceedings{devlin2019bert,
  title={Bert: Pre-training of deep bidirectional transformers for language understanding},
  author={Devlin, Jacob and Chang, Ming-Wei and Lee, Kenton and Toutanova, Kristina},
  booktitle={Proceedings of the 2019 conference of the North American chapter of the association for computational linguistics: human language technologies, volume 1 (long and short papers)},
  pages={4171--4186},
  year={2019}
}

@article{dosovitskiy2020image,
  title={An image is worth 16x16 words: Transformers for image recognition at scale},
  author={Dosovitskiy, Alexey and Beyer, Lucas and Kolesnikov, Alexander and Weissenborn, Dirk and Zhai, Xiaohua and Unterthiner, Thomas and Dehghani, Mostafa and Minderer, Matthias and Heigold, Georg and Gelly, Sylvain and others},
  journal={arXiv preprint arXiv:2010.11929},
  year={2020}
}

@misc{jordan2024muon,
  author       = {Keller Jordan and Yuchen Jin and Vlado Boza and Jiacheng You and
                  Franz Cesista and Laker Newhouse and Jeremy Bernstein},
  title        = {Muon: An optimizer for hidden layers in neural networks},
  year         = {2024},
  url          = {https://kellerjordan.github.io/posts/muon/}
}
